\documentclass[twocolumn]{autart}    

\usepackage{graphicx}          

\usepackage{subfigure}
\usepackage{epsfig}
\usepackage{amsmath}
\usepackage{cite}                               
\usepackage{xcolor}

\begin{document}

\begin{frontmatter}

\title{Linear SLAM: Linearising the SLAM Problems\\ using Submap Joining\thanksref{footnoteinfo}} 

\thanks[footnoteinfo]{A preliminary version of this paper has been published in the 2013 IEEE/RSJ International Conference on Intelligent Robots and Systems \cite{Liang:IROS-13}. Corresponding author L.~Zhao. Tel. +61 2 9514 1223.}

\author{Liang Zhao}\ead{Liang.Zhao@uts.edu.au},    
\author{Shoudong Huang}\ead{Shoudong.Huang@uts.edu.au},               
\author{Gamini Dissanayake}\ead{Gamini.Dissanayake@uts.edu.au}  

\address{Centre for Autonomous Systems,
Faculty of Engineering and Information Technology\\
University of Technology Sydney, Australia}  


\begin{abstract}                          
The main contribution of this paper is a new submap joining based approach for solving large-scale Simultaneous Localization and Mapping (SLAM) problems. Each local submap is independently built using the local information through solving a small-scale SLAM; the joining of submaps mainly involves solving linear least squares and performing nonlinear coordinate transformations. Through approximating the local submap information as the state estimate and its corresponding information matrix, judiciously selecting the submap coordinate frames, and approximating the joining of a large number of submaps by joining only two maps at a time, either sequentially or in a more efficient Divide and Conquer manner, the nonlinear optimization process involved in most of the existing submap joining approaches is avoided. Thus the proposed submap joining algorithm does not require initial guess or iterations since linear least squares problems have closed-form solutions. The proposed Linear SLAM technique is applicable to feature-based SLAM, pose graph SLAM and D-SLAM, in both two and three dimensions, and does not require any assumption on the character of the covariance matrices. Simulations and experiments are performed to evaluate the proposed Linear SLAM algorithm. Results using publicly available datasets in 2D and 3D show that Linear SLAM produces results that are very close to the best solutions that can be obtained using full nonlinear optimization algorithm started from an accurate initial guess. The C/C++ and MATLAB source codes of Linear SLAM are available on OpenSLAM.
\end{abstract}

\end{frontmatter}

\section{Introduction}

Simultaneous Localization and Mapping (SLAM) is the problem of using a mobile robot to build a map of an unknown environment and at the same time locating the robot within the map \cite{dissa2001,Cadena2016}. Recently, nonlinear optimization techniques have become popular for solving
SLAM problems. Following the seminal work by Lu and Milios \cite{Lu}, many efficient SLAM solutions have been developed by exploiting the sparseness of the information
matrix and applying different approaches for solving the sparse linear
equations \cite{SAM,kurt,G2O,iSAM,Kaess12ijrr}.
However, since SLAM is formulated as a high dimensional
nonlinear optimization problem, finding the global minimum is nontrivial.  Although
many SLAM algorithms appear to
provide good results for most of the practical datasets (e.g. \cite{Olson,TORO-journal,Shoudong:IROS-10}), there is no guarantee that the algorithm can
converge to the global optimum and having an accurate initial
value is very critical, especially for large-scale SLAM problems. Moreover, for very large-scale SLAM problems (e.g. life long SLAM), it is in general intractable to keep all the original data and perform full nonlinear optimization to obtain the SLAM solution. Some sort of information fusion is necessary to reduce the computational complexity.

Local submap joining has shown to be an efficient way to build
large-scale maps \cite{Divide-and-Conquer:-EKF-SLAM,Tardos-map-joining,Williams-PhD-thesis,Shoudong:SLSJF-TRO,Atlas,Multi-pose-graph-ICRA10,TSAMICRA07}. The idea of many map joining algorithms
such as Sparse Local Submap Joining Filter (SLSJF) \cite{Shoudong:SLSJF-TRO} is to treat the estimated state of each local map as an integrated
observation (the uncertainty is expressed by the local map covariance matrix) in the map joining step. By summarising the original data within the local map in this way, the SLAM problem can be solved more efficiently. It should be noted that most of the existing map joining problems are formulated as nonlinear estimation or
nonlinear optimization problems which are solved by Extended Kalman Filter (EKF) \cite{Divide-and-Conquer:-EKF-SLAM,Tardos-map-joining,Williams-PhD-thesis}, Extended Information Filter (EIF) \cite{Shoudong:SLSJF-TRO} or nonlinear
least squares \cite{Atlas,Multi-pose-graph-ICRA10,TSAMICRA07,Shoudong:I-SLSJF-ACRA} techniques.

This paper provides a new map joining framework which only requires solving linear least squares problems and performing nonlinear coordinate transformations. The method can be applied to join pose-feature maps (maps that contain both robot poses and features), pose-only maps (maps that only contain robot poses), and feature-only maps (maps that only contain features), for both 2D and 3D scenarios. There is no assumption on the structure of the covariance matrices of the local maps (no need to be spherical as in \cite{huangICRA2012,WangRSS2012,Rosen2016}). Since linear least squares problems have closed-form solutions, there is no need of an initial guess and no need of iterations to solve the reformulated map joining problems.
It is demonstrated using publicly available datasets that the proposed Linear SLAM algorithm can provide accurate SLAM results. For practical applications, the proposed algorithm can be easily combined with other robust local map building algorithms  in complex unknown environments such as \cite{Robotica_1_2017,Robotica_2_2018} to solve practical large-scale SLAM problems  in a more robust manner.

The paper is an extended version of our preliminary work  \cite{Liang:IROS-13}. The major improvements of this paper over \cite{Liang:IROS-13} are:
(1) It has been shown that Linear SLAM is also applicable to the joining of  feature-only maps; thus it provides a unified framework for solving different versions of SLAM problems (pose-feature, pose-only, and feature-only); (2) The consistency analysis of Linear SLAM is performed to demonstrate its estimation consistency; (3) The theoretical analysis on the computational complexity of Linear SLAM is also performed to demonstrate its efficiency; (4) More quantitive comparisons and evaluations of Linear SLAM are performed; (5) More insights and discussions are provided; and (6) More efficient C/C++ implementation of the algorithm is done and the source code is available on OpenSLAM.

The paper is organized as follows. Section \ref{sec:related-work} discusses the related work. Section \ref{sec:local-map-in-SLAM} points out that different local maps can be built using the same SLAM data by selecting different coordinate frames. Section \ref{sec:why-linear} explains the process of using linear least squares to solve the problem of joining two pose-feature maps. Section \ref{sec:sequence} explains how to use
the linear method in the sequential and Divide and Conquer local submap joining. Section \ref{sec:linear-pose-graph-SLAM} and Section \ref{sec:linear-DSLAM} describe how to apply the linear algorithm to the joining of two pose-only maps and two feature-only maps, respectively. In Section
\ref{sec:results}, simulation and experimental results using publicly available datasets are given to demonstrate the accuracy of Linear SLAM, and compare with some other submap joining based SLAM algorithms.
Section \ref{sec:discussions} discusses the pros and cons of the proposed algorithm. Finally, Section \ref{sec:conclusion} concludes the paper. To improve the readability of the paper, some technical details are provided in the appendices.

\section{Related Work}\label{sec:related-work}

Local submap joining has been a strategy applied by many researchers \cite{Atlas,Multi-pose-graph-ICRA10,Divide-and-Conquer:-EKF-SLAM,TSAMICRA07,Tardos-map-joining,Williams-PhD-thesis,Shoudong:SLSJF-TRO,Shoudong:I-SLSJF-ACRA}.

The earlier work on map joining are proposed by Tardos \textit{et al.} (Map joining 1.0) \cite{Tardos-map-joining} and Williams \cite{Williams-PhD-thesis}. Both of them apply sequential map joining where a global map and a local map are joined at a time. In Map joining 1.0 \cite{Tardos-map-joining}, the idea used in joining two maps is to first transform one of the map using an estimate of relative pose from another map to have both submaps in a common reference, and then apply the coincidence constraint to the common elements. Williams \cite{Williams-PhD-thesis} uses an EKF to fuse the local maps sequentially where each local map is treated as an integrated observation. Since the global map becomes larger and larger after sequentially fusing the local maps, the dense covariance matrix makes the EKF algorithm less efficient. To improve the efficiency of the map joining process, SLSJF is proposed in \cite{Shoudong:SLSJF-TRO} where EIF is used and exactly sparse information matrix is achieved by keeping the robot starting poses and end poses of the local maps in the global state vector. CF-SLAM proposed in \cite{Combined-filter} further demonstrated that by using the Divide and Conquer strategy on top of EIF, the efficiency (and in most cases accuracy) could be improved further.

To overcome the potential estimate inconsistency of EKF/EIF in map joining, I-SLSJF is proposed in \cite{Shoudong:I-SLSJF-ACRA}, which is an optimization based map joining treating each local map as an integrated observation, and using the result of SLSJF as the initial value for the nonlinear least squares optimization. More optimization based map joining are proposed in the past few years, for both feature-based SLAM \cite{TSAMICRA07,Grisetti-IROS2012} and pose graph SLAM \cite{Multi-pose-graph-ICRA10,HOG-Man,COP-SLAM}. In \cite{TSAMICRA07}, the local maps without any common nodes are first optimized independently, and then the base node of each local map is optimized by using the inter-measurements between different local maps, where the linearization of the local maps can be cached and reused. While in \cite{Grisetti-IROS2012}, after building the local maps, the condensed factors are generated based on the solution of the local maps, where the solution to the new graph is a global configuration of the origins and the shared variables. Then a good initial guess can be determined from the solutions of the origins and the shared variables, and the full nonlinear least squares is performed to get the exact results. In \cite{HOG-Man}, a hierarchical pose graph optimization on manifolds is proposed, where only the coarse structure of the scene is corrected, resulting in an efficient SLAM algorithm. In \cite{COP-SLAM}, a closed-form online 3D pose-chain SLAM algorithm is proposed. Due to the spherical assumption on the covariance matrices, it can efficiently optimize pose-chains by exploiting their Lie group structure. The spherical covariance assumption significantly reduces the complexity of the SLAM problems, as shown in recent works \cite{WangRSS2012,Rosen2016,WangAutomatica2015}.
In \cite{Multi-pose-graph-ICRA10}, a relative formulation of the relationship between multiple pose graphs is proposed to avoid the initialization problem and thus lead to an efficient solution; the anchor nodes are closely related to the base nodes in \cite{TSAMICRA07}.

Map joining has shown to be able to improve the efficiency for large-scale SLAM as well as reduce the linearisation errors. Similar to recursive nonlinear estimation (e.g. \cite{iSAM}), map joining also solves a recursive version of the problem step by step, thus rarely gets stuck in local minima. However, when the local map information is summarized as local map estimate and its covariance matrix, the map joining problem is slightly different from the original full nonlinear optimization SLAM problem using the original SLAM data. In general, the more accurate the local maps are, the closer the map joining result to the best solution to the full nonlinear least squares SLAM. Recently, good progress has been made in using nonlinear optimization techniques to solve small-scale SLAM with mathematical guarantees, such as one-step or two-step SLAM \cite{WangAutomatica2015} or through verifications techniques \cite{CarloneICRA15}. These make the application of map joining techniques more promising.

Existing map joining algorithms are based on nonlinear estimation or nonlinear optimization, with no convergence guarantee. This paper proposes a new map joining algorithm, Linear SLAM, where only solving linear least squares and performing nonlinear coordinate transformations are needed, thus guarantees convergence. Although in Map joining 1.0 \cite{Tardos-map-joining}, the idea of first transforming one of the map using an estimate of relative pose from another map to have both submaps in a common reference is similar to Linear SLAM in the case of point features, significant inaccuracy is introduced in Map joining 1.0 because external relative pose information is used to perform the map transformation. This can be clearly seen from its poor performance as compared to Map joining 2.0 \cite{Divide-and-Conquer:-EKF-SLAM}. While, Linear SLAM has shown to perform better than most of the existing nonlinear map joining algorithms (see Section \ref{sec:results}).

The following sections will explain the details of Linear SLAM.

\begin{figure*}
\centering
\centering \subfigure[The odometry and observation information for building a local map] {\label{fig:observations}
\includegraphics[height=45mm]{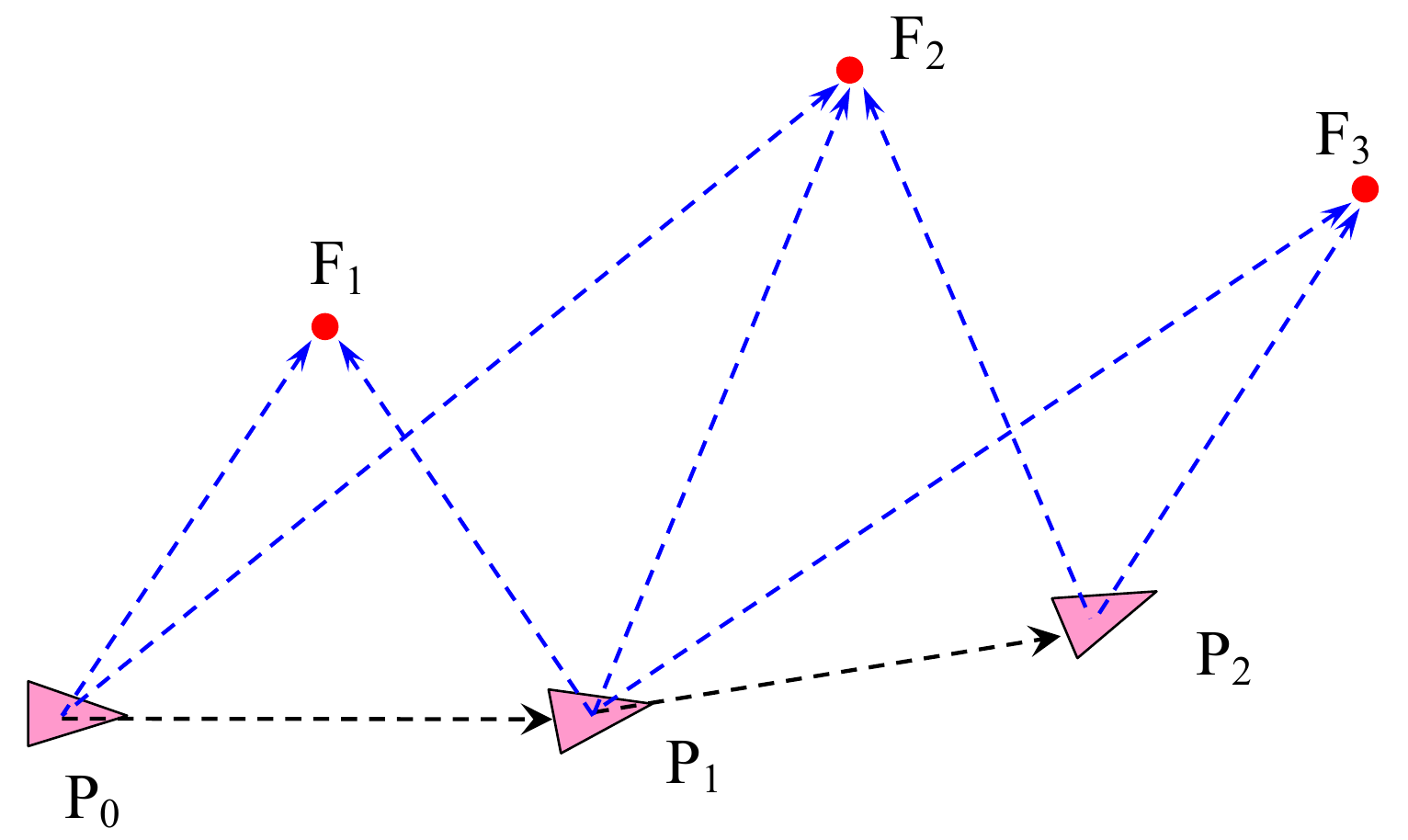}} \hspace{5mm}
\centering \subfigure[Local map with coordinate frame defined by $\textbf{P}_0$]{\label{fig:localmap1}
\includegraphics[height=45mm]{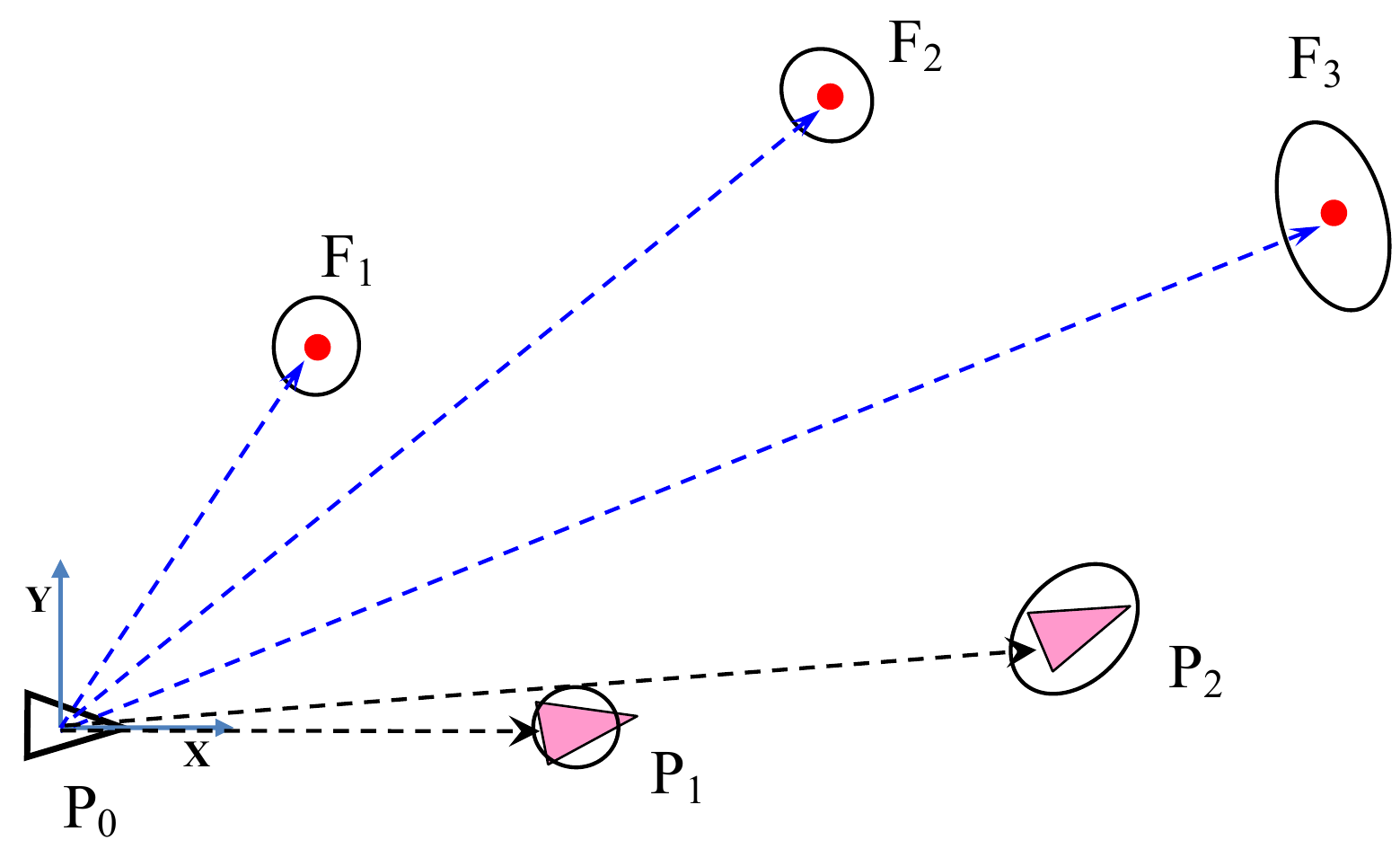}}
\centering \subfigure[Local map with coordinate frame defined by $\textbf{P}_2$]{\label{fig:localmap2}
\includegraphics[height=45mm]{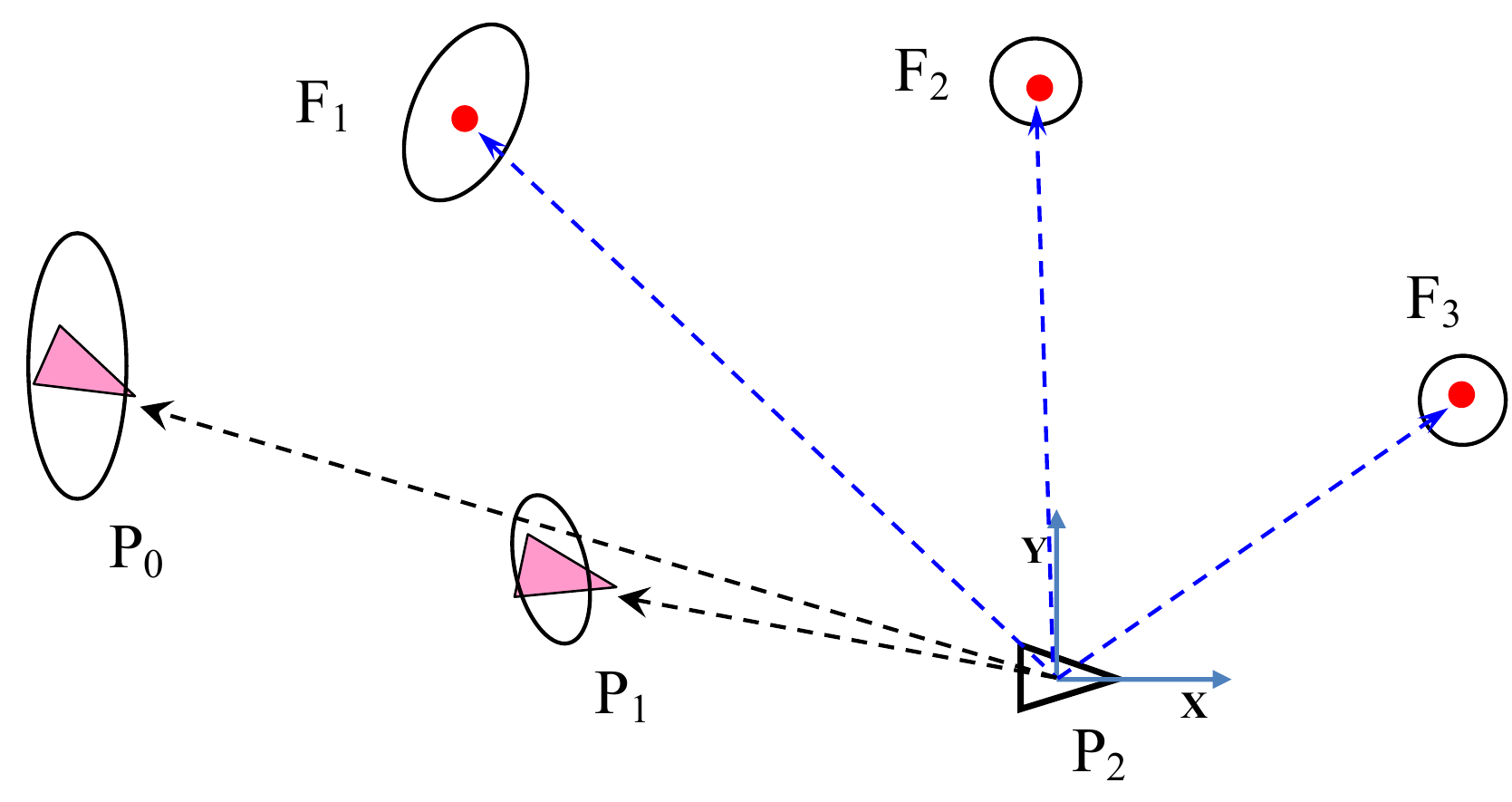}}\hspace{5mm}
\centering \subfigure[Local map with coordinate frame defined by $\textbf{F}_1$ and $\textbf{F}_2$]{\label{fig:localmap3}
\includegraphics[height=45mm]{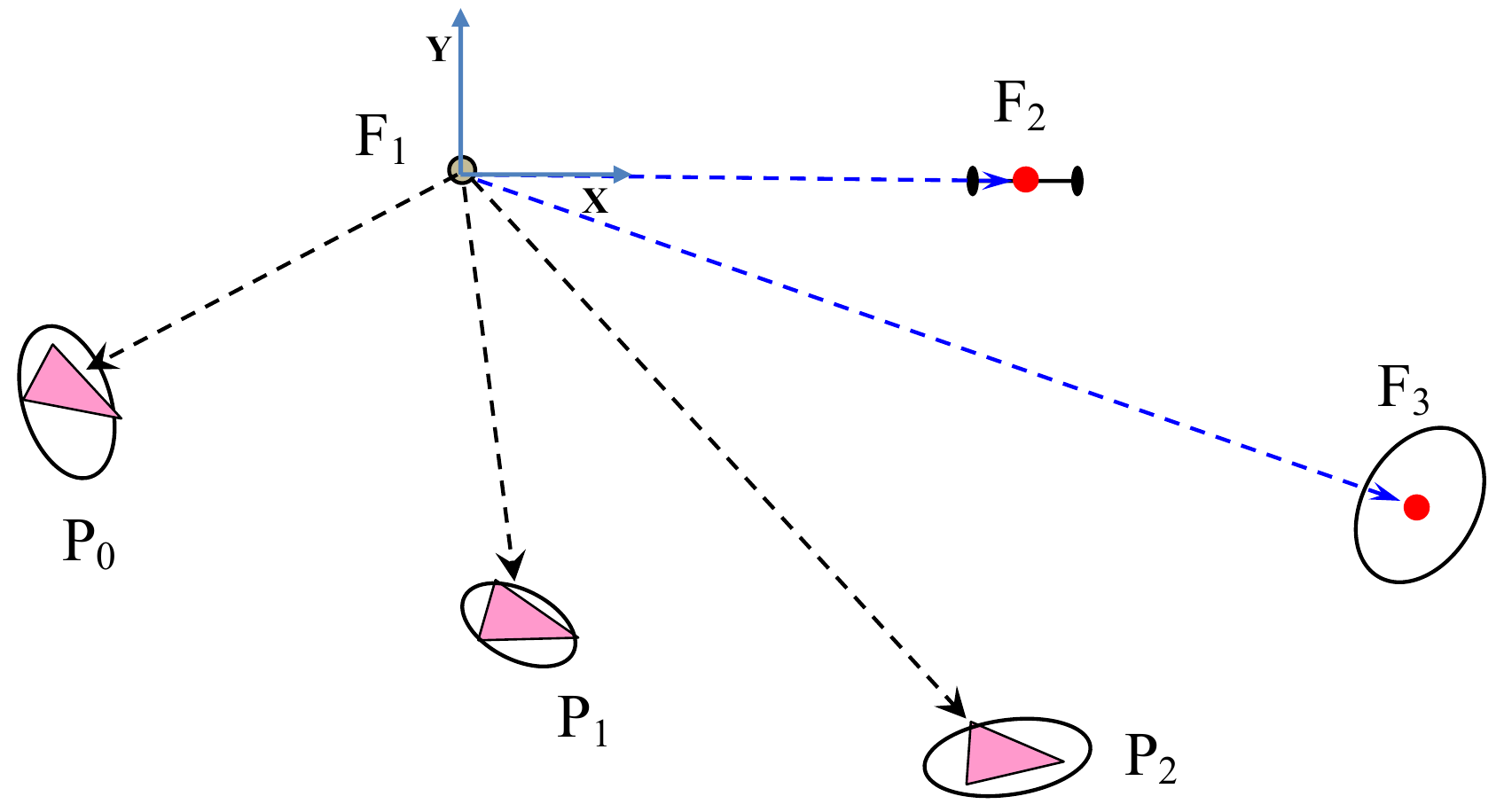}}
\caption{Different local maps built from the same odometry and observation information. Here $\textbf{P}_i$ represents the robot pose and $\textbf{F}_j$ represents the feature position.} \label{Fig:LocalMaps}
\end{figure*}

\section{{Different Local Submaps in SLAM}}\label{sec:local-map-in-SLAM}

The aim of this section is to show that we have the freedom to choose the coordinate frame of the local and global maps, which is the fundamental idea of the proposed Linear SLAM and the reason why the traditional map joining methods are nonlinear without considering this. We also show that the marginalization of local maps makes the proposed Linear SLAM a unified linear approach for solving different SLAM problems.

A local submap in SLAM refers to a map built using only a small portion of the SLAM data. In a feature-based SLAM, the data used for building a local submap contains odometry and observation information related to a small set of robot poses and features. Fig. \ref{fig:observations} shows an example of odometry and observation information related to three poses $\textbf{P}_0,\textbf{P}_1,\textbf{P}_2$ and three features $\textbf{F}_1,\textbf{F}_2,\textbf{F}_3$.

\subsection{The Freedom of Choosing the Coordinate Frame}\label{sec:coordinate-frame}

Consider the problem of building a local map by performing nonlinear least squares optimization based on the odometry and observation information given in the pose-feature graph shown in Fig. \ref{fig:observations}. Since the odometry and observation are all relative information, we must specify the coordinate frame when building a map.

One common choice in SLAM is to fix the first pose $\textbf{P}_0$ as $(0,0,0)$ to define the coordinate frame, then we can obtain a local map shown in Fig. \ref{fig:localmap1} after the nonlinear least squares optimization. The state vector of the local map contains two poses $\textbf{P}_1$ and $\textbf{P}_2$ and three features, all in the coordinate frame of $\textbf{P}_0$.

Alternatively, we can fix the last pose $\textbf{P}_2$ as $(0,0,0)$ to define the coordinate frame, then we can obtain a local map shown in Fig. \ref{fig:localmap2}. The state vector of the local map contains two poses $\textbf{P}_0$ and $\textbf{P}_1$ and three features, all in the coordinate frame of $\textbf{P}_2$.

The choice of the coordinate frame can also be that defined by two features $\textbf{F}_1$ and $\textbf{F}_2$, that is, the feature position $\textbf{F}_1$ is fixed as (0,\;0) and the x-axis is along the direction from $\textbf{F}_1$ to $\textbf{F}_2$. In this case, we can obtain a local map as shown in Fig. \ref{fig:localmap3}. The state vector of this local map contains three poses $\textbf{P}_0$, $\textbf{P}_1$ and $\textbf{P}_2$ and feature $\textbf{F}_3$ as well as the x-coordinate of feature $\textbf{F}_2$, all in the coordinate frame defined by the two features $\textbf{F}_1$ and $\textbf{F}_2$.

The uncertainty of the local map is described by the associated covariance matrix (or information matrix).
The above three local maps (including their covariance matrices) can be transferred from one to another by performing a coordinate transformation.

\subsection{The Option of Marginalization}\label{sec:marginalization}

If one wants to reduce the size of the local map, marginalization can be performed after building the local map containing all the poses and features. For example, one can marginalize out all the poses apart from the last pose, then a state vector as that used in EKF SLAM is obtained \cite{dissa2001}. One can also marginalize out all the features, then a state vector as that used in pose graph SLAM is obtained \cite{Olson}. Alternatively, one can also marginalize out all the poses, then a state vector only containing features can be obtained, as that used in D-SLAM mapping \cite{Wang:DSLAM-IJRR}.

After marginalization, the new covariance matrix (or information matrix) corresponding to the state vector with reduced dimension can be obtained.

In this paper, a map that contains both poses and features is called a pose-feature map; a map that only contains poses is called a pose-only map; while a map that only contains features is called a feature-only map.

\subsection{The Idea of Map Joining}\label{sec:localmaps_for_mapjoining}

Assuming each local map estimate is consistent, map joining \cite{Williams-PhD-thesis,Shoudong:SLSJF-TRO} uses each local map as an integrated observation to replace the original odometry and observation information. This significantly reduces the computational cost for large-scale SLAM. The covariance matrices of the local maps play an important role since they describe the uncertainties of the integrated observations. Note that when computing the local map covariance matrix (information matrix), the covariance matrices of the original odometry and observation information are used, thus the local map can be regarded as a summary of the odometry and observation information involved.

The local maps need to have common poses or features such that they can be joined together. At least one common pose or two common features for 2D (three common features for 3D) are required to join two local maps together.

The key point of this section is that we have the freedom to choose the coordinate frame of a local map. Moreover, we also have the freedom to choose the coordinate frame of the global map.
In this paper, we will demonstrate that by judiciously selecting the coordinate frames of the local maps at different map joining steps, the map joining result can be obtained by solving linear least squares problems and performing nonlinear coordinate transformations. This is the major difference to the traditional map joining. The proposed approach can be applied to the joining of different versions of local maps including pose-feature maps, pose-only maps and feature-only maps. Thus, a unified linear approach for solving SLAM problems is obtained. The following sections provide some details of the approach.

\section{Joining Two Pose-feature Maps using Linear Least Squares}\label{sec:why-linear}

A pose-feature map consists of a number of features and either one robot pose (such as the map built by conventional EKF SLAM) or a number of robot poses (such as the map built by nonlinear least squares optimization without marginalizing out poses).

\begin{figure}
\centering
{\includegraphics[height=85mm]{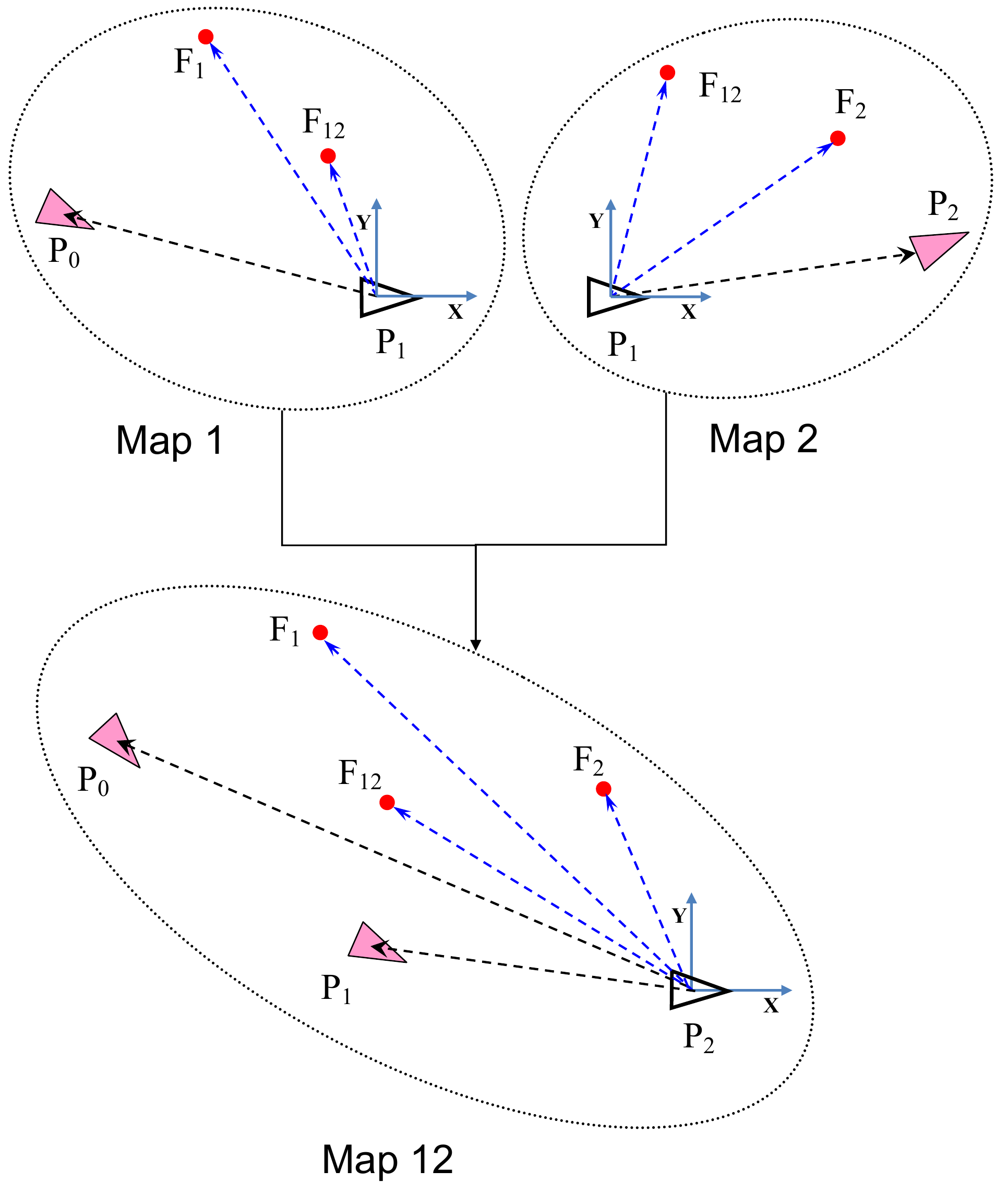}} \caption{The
proposed new way of building and joining two maps.}
\label{fig:joining-two-localmaps-new}
\end{figure}

\subsection{The Coordinate Frames of the Maps}

In this paper, we propose to choose the coordinate frames of the two pose-feature maps and the combined map as follows (see Fig. \ref{fig:joining-two-localmaps-new}).

Map 1 is built in the
coordinate frame defined by its end pose $\textbf{P}_1$. Map 2 is built in the
coordinate frame defined by its start pose $\textbf{P}_1$ (which is the same as the end pose of Map 1) \footnote{Map 1 and
Map 2 can either be small local maps built using the local odometry and observation information, or be larger maps as the result of combining
a number of small local maps. There can be some other robot poses
in Map 1 and Map 2, they are not shown in Fig. \ref{fig:joining-two-localmaps-new} for
simplification.}.
The coordinate frame of the combined map, Map 12, is defined by $\textbf{P}_2$, the
robot end pose of Map 2.

In the following, we will show that although the joining of Map 1
and Map 2 to get Map 12 in Fig. \ref{fig:joining-two-localmaps-new} is still a nonlinear optimization problem,
it is equivalent to a linear least squares optimization problem plus a
(nonlinear) coordinate transformation.

\subsection{Nonlinear Least Squares Formulation of Joining Two Maps}\label{Subsec::N12}

Suppose Map 1 and Map 2 in Fig.
\ref{fig:joining-two-localmaps-new} are denoted by $M^{L_1}$ and $M^{L_2}$ and are given by
\begin{equation}
M^{L_1} = (\hat{\textbf{X}}^{L_1}, \; I^{L_1}), \quad M^{L_2} =
(\hat{\textbf{X}}^{L_2}, \; I^{L_2})
\label{M1M2localmaps}
\end{equation}
where $\hat{\textbf{X}}^{L_1}$ and $\hat{\textbf{X}}^{L_2}$ are
the estimates of the state vectors $\textbf{X}^{L_1}$ and
$\textbf{X}^{L_2}$ of each map, and $I^{L_1}$ and $I^{L_2}$ are the associated information matrices.

The state vectors $\textbf{X}^{L_1}$ and
$\textbf{X}^{L_2}$ are defined as \footnote{To simplify the
notations, the `transpose's in the state vectors are sometimes omitted in this paper. For example,
$\textbf{X}^{L_1},\textbf{P}^{L_1}_{0},\textbf{X}^{L_1}_{F_1},\textbf{X}^{L_1}_{F_{12}}$ are all column vectors and the rigorous
notation should be $\textbf{X}^{L_1} = [(\textbf{P}^{L_1}_{0})^T, \; (\textbf{X}^{L_1}_{F_1})^T, \; (\textbf{X}^{L_1}_{F_{12}})^T]^T$.}
\begin{equation}\label{Eq::state vector MJ}
\begin{aligned}
\textbf{X}^{L_1} &= [\textbf{P}^{L_1}_{0}, \; \textbf{X}^{L_1}_{F_1}, \; \textbf{X}^{L_1}_{F_{12}}]\\
\textbf{X}^{L_2} &= [\textbf{P}^{L_2}_{2}, \; \textbf{X}^{L_2}_{F_2}, \; \textbf{X}^{L_2}_{F_{12}}].\\
\end{aligned}
\end{equation}
Note that $M^{L_1}$ is
in the coordinate frame of $\textbf{P}_{1}$, and $M^{L_2}$ is also
in the coordinate frame of $\textbf{P}_{1}$. Here $\textbf{P}_{1}$ is the end pose of $M^{L_1}$ (also the start
pose of $M^{L_2}$). It defines the coordinate frame of both
$M^{L_1}$ and $M^{L_2}$ and thus is not included in the state vectors of the two maps.

In the state vectors $\textbf{X}^{L_1}$ and $\textbf{X}^{L_2}$ in (\ref{Eq::state vector
MJ}), $\textbf{X}^{L_1}_{F_1}$ and $\textbf{X}^{L_2}_{F_2}$ represent the
features only appear in $M^{L_1}$ or $M^{L_2}$, respectively, while
$\textbf{X}^{L_1}_{F_{12}}$ and $\textbf{X}^{L_2}_{F_{12}}$ represent the
common features appear in both the two maps.

In the state vector $\textbf{X}^{L_1}$ in (\ref{Eq::state vector
MJ}), the start pose $\textbf{P}_{0}$ of $M^{L_1}$ is
presented by
\begin{equation}\label{Eq::Pose P0 MJ}
\textbf{P}^{L_1}_{0} = [\textbf{t}^{L_1}_0, \; \textbf{r}^{L_1}_0]
\end{equation}
where $\textbf{t}^{L_1}_0$ is the translation vector, and
$\textbf{r}^{L_1}_0$ is/are the rotation angle/angles of pose
$\textbf{P}^{L_1}_{0}$. Similarly, in the state vector $\textbf{X}^{L_2}$ in (\ref{Eq::state vector
MJ}), the end pose $\textbf{P}_{2}$ of $M^{L_2}$ is
presented by
\begin{equation}\label{Eq::Pose P2 MJ}
\textbf{P}^{L_2}_{2} = [\textbf{t}^{L_2}_2, \; \textbf{r}^{L_2}_2].
\end{equation}

Now the state estimates $\hat{\textbf{X}}^{L_1}$ and $\hat{\textbf{X}}^{L_2}$ can be written clearly as
\begin{equation}\label{Eq::state vector estimate MJ}
\begin{aligned}
\hat{\textbf{X}}^{L_1} &= [\hat{\textbf{t}}^{L_1}_0, \; \hat{\textbf{r}}^{L_1}_0, \; \hat{\textbf{X}}^{L_1}_{F_1}, \; \hat{\textbf{X}}^{L_1}_{F_{12}}]\\
\hat{\textbf{X}}^{L_2} &= [\hat{\textbf{t}}^{L_2}_2, \; \hat{\textbf{r}}^{L_2}_2, \; \hat{\textbf{X}}^{L_2}_{F_2}, \; \hat{\textbf{X}}^{L_2}_{F_{12}}].
\end{aligned}
\end{equation}

Our goal is to join $M^{L_1}$ and $M^{L_2}$ to get the map
$M^{G_{12}}$, where $M^{G_{12}}$ is in the coordinate frame of
$\textbf{P}_{2}$.

The state vector of $M^{G_{12}}$ containing pose $\textbf{P}_0$, pose $\textbf{P}_1$ and all the features is defined as
\begin{equation}
\begin{aligned}
\textbf{X}^{G_{12}} &= [\textbf{P}^{G_{12}}_{0}, \;\textbf{P}^{G_{12}}_{1}, \; \textbf{X}^{G_{12}}_{F_1}, \; \textbf{X}^{G_{12}}_{F_2}, \; \textbf{X}^{G_{12}}_{F_{12}}]\\
& =
[\textbf{t}^{G_{12}}_{0}, \; \textbf{r}^{G_{12}}_{0}, \; \textbf{t}^{G_{12}}_{1}, \; \textbf{r}^{G_{12}}_{1}, \; \textbf{X}^{G_{12}}_{F_1}, \; \textbf{X}^{G_{12}}_{F_2}, \; \textbf{X}^{G_{12}}_{F_{12}}]
\end{aligned}
\end{equation}
where all the variables in the state vector $\textbf{X}^{G_{12}}$
are in the coordinate frame of $\textbf{P}_{2}$.

Similar to \cite{Shoudong:I-SLSJF-ACRA}, in the map joining problem, the state estimates $\hat{\textbf{X}}^{L_1}$ and $\hat{\textbf{X}}^{L_2}$ are treated as two integrated ``observations" where the associated information matrices $I^{L_1}$ and $I^{L_2}$ describe the uncertainties of the observations. The optimization problem of joining the two maps is to minimize the following objective function
\begin{equation}\label{Eq::ObjFunc_N12}
\begin{aligned}
&f(\textbf{X}^{G_{12}})=\|\textbf{e}_1\|^2_{I^{L_{1}}}+\|\textbf{e}_2\|^2_{I^{L_{2}}}\\
=&\left \| \left[
\begin{aligned}
R_{1} \, (\textbf{t}^{G_{12}}_{0}-\textbf{t}^{G_{12}}_{1}) &- \hat{\textbf{t}}^{L_1}_0\\
r^{-1}(R_{0} R_{1}^T) &- \hat{\textbf{r}}^{L_1}_0\\
R_{1} \, (\textbf{X}^{G_{12}}_{F_1}-\textbf{t}^{G_{12}}_{1}) &- \hat{\textbf{X}}^{L_1}_{F_1}\\
R_{1} \, (\textbf{X}^{G_{12}}_{F_{12}}-\textbf{t}^{G_{12}}_{1}) &- \hat{\textbf{X}}^{L_1}_{F_{12}}\\
\end{aligned} \right]\right \|^2_{I^{L_{1}}}\\
+&\left \| \left[
\begin{aligned}
-R_{1} \, \textbf{t}^{G_{12}}_{1} &- \hat{\textbf{t}}^{L_2}_2\\
r^{-1}(R_{1}^T) &- \hat{\textbf{r}}^{L_2}_2\\
R_{1} \, (\textbf{X}^{G_{12}}_{F_2}-\textbf{t}^{G_{12}}_{1}) &- \hat{\textbf{X}}^{L_2}_{F_2}\\
R_{1} \, (\textbf{X}^{G_{12}}_{F_{12}}-\textbf{t}^{G_{12}}_{1}) &- \hat{\textbf{X}}^{L_2}_{F_{12}}\\
\end{aligned} \right]\right \|^2_{I^{L_2}}
\end{aligned}
\end{equation}
where $\textbf{e}_i, (i=1,2)$ is the residual vector of the integrated ``observation", and $\|\textbf{e}_i\|^2_{I^{L_{i}}}=\textbf{e}_i^{T}I^{L_{i}}\textbf{e}_i$ ($i=1,2$) denotes the weighted norm of vector $\textbf{e}_i$ with the given information matrix $I^{L_{i}}$.

\begin{figure}[tb]
\centering
\includegraphics[width=0.52\textwidth]{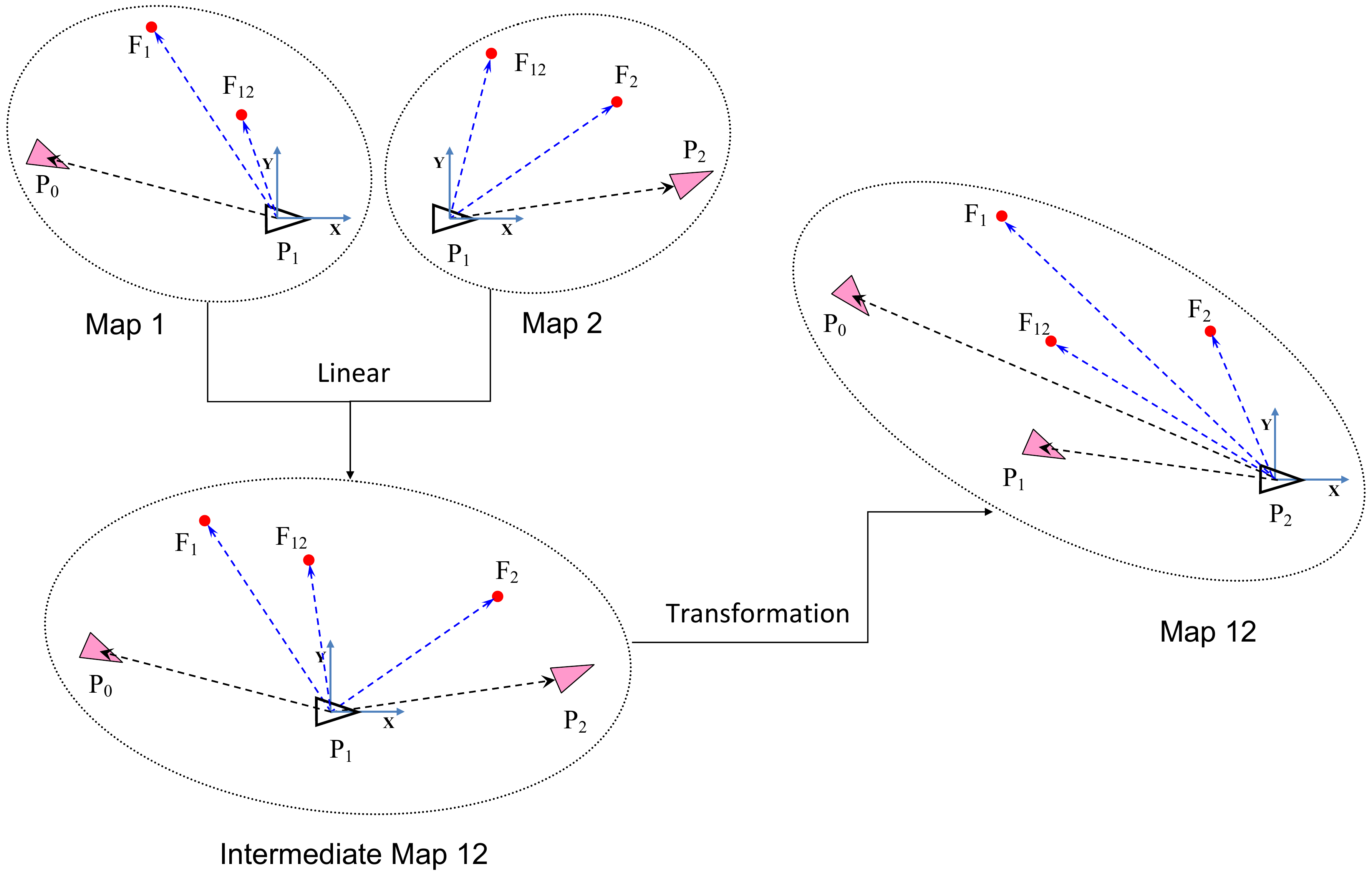}
\caption{Linear way to build Map 12.}
\label{fig:joining-two-maps-transform}
\end{figure}

In (\ref{Eq::ObjFunc_N12}), $R_{0} = r(\textbf{r}^{G_{12}}_{0})$ and $R_{1} = r(\textbf{r}^{G_{12}}_{1})$ are the rotation matrices of pose $\textbf{P}^{G_{12}}_{0}$ and $\textbf{P}^{G_{12}}_{1}$ in the global state vector $\textbf{X}^{G_{12}}$, respectively. And $r(\cdot)$ and $r^{-1}(\cdot)$ are the
angle-to-matrix and matrix-to-angle functions. For 2D scenarios,
$r^{-1}(R_{0} R_{1}^T) =
\textbf{r}^{G_{12}}_{0}-\textbf{r}^{G_{12}}_{1}$ and
$r^{-1}(R_{1}^T) = -\textbf{r}^{G_{12}}_{1}$.

The problem of minimizing (\ref{Eq::ObjFunc_N12}) is a nonlinear least squares problem. Solving it we can obtain the
estimate of Map 12 together with its information matrix as
\begin{equation}\label{Eq::EstN12}
M^{G_{12}} = (\hat{\textbf{X}}^{G_{12}},\; I^{G_{12}}).
\end{equation}

\subsection{Linear Method to Obtain $M^{G_{12}}$}\label{testShoudong}

If we define new variables as follows (they are actually the seven distinct (nonlinear) functions in (\ref{Eq::ObjFunc_N12}))
\begin{equation}\label{Eq::NewSV}
\begin{aligned}
\begin{aligned}
\bar{\textbf{t}}^{G_{12}}_{0} &= R_{1} \, (\textbf{t}^{G_{12}}_{0}-\textbf{t}^{G_{12}}_{1}) \\
\bar{\textbf{r}}^{G_{12}}_{0} &= r^{-1}(R_{0} R_{1}^T) \\
\bar{\textbf{t}}^{G_{12}}_{2} &= -R_{1} \, \textbf{t}^{G_{12}}_{1} \\
\bar{\textbf{r}}^{G_{12}}_{2} &= r^{-1}(R_{1}^T) \\
\bar{\textbf{X}}^{G_{12}}_{F_1} &= R_{1} \, (\textbf{X}^{G_{12}}_{F_1}-\textbf{t}^{G_{12}}_{1}) \\
\bar{\textbf{X}}^{G_{12}}_{F_2} &= R_{1} \, (\textbf{X}^{G_{12}}_{F_2}-\textbf{t}^{G_{12}}_{1}) \\
\bar{\textbf{X}}^{G_{12}}_{F_{12}} &= R_{1} \, (\textbf{X}^{G_{12}}_{F_{12}}-\textbf{t}^{G_{12}}_{1}) \\
\end{aligned}
\end{aligned}
\end{equation}
and define a new state vector as
\begin{equation}\label{Eq::SVL12}
\begin{aligned}
\bar{\textbf{X}}^{G_{12}} &= [\bar{\textbf{t}}^{G_{12}}_{0}, \; \bar{\textbf{r}}^{G_{12}}_{0},
\; \bar{\textbf{t}}^{G_{12}}_{2}, \; \bar{\textbf{r}}^{G_{12}}_{2},
\; \bar{\textbf{X}}^{G_{12}}_{F_1}, \;
\bar{\textbf{X}}^{G_{12}}_{F_2}, \;
\bar{\textbf{X}}^{G_{12}}_{F_{12}}]\\
&= g(\textbf{X}^{G_{12}})
\end{aligned}
\end{equation}
then the nonlinear least squares problem to minimize the objective function
(\ref{Eq::ObjFunc_N12}) becomes a linear least squares problem to minimize the following objective function
\begin{equation}\label{Eq::ObjFunc_L12}
\bar{f}(\bar{\textbf{X}}^{G_{12}})= \left \| \left[
\begin{aligned}
\bar{\textbf{t}}^{G_{12}}_{0} &- \hat{\textbf{t}}^{L_1}_0\\
\bar{\textbf{r}}^{G_{12}}_{0} &- \hat{\textbf{r}}^{L_1}_0\\
\bar{\textbf{X}}^{G_{12}}_{F_1} &- \hat{\textbf{X}}^{L_1}_{F_1}\\
\bar{\textbf{X}}^{G_{12}}_{F_{12}} &- \hat{\textbf{X}}^{L_1}_{F_{12}}\\
\end{aligned} \right]\right \|^2_{I^{L_{1}}}\\
+\left \| \left[
\begin{aligned}
\bar{\textbf{t}}^{G_{12}}_{2} &- \hat{\textbf{t}}^{L_2}_2\\
\bar{\textbf{r}}^{G_{12}}_{2} &- \hat{\textbf{r}}^{L_2}_2\\
\bar{\textbf{X}}^{G_{12}}_{F_2} &- \hat{\textbf{X}}^{L_2}_{F_2}\\
\bar{\textbf{X}}^{G_{12}}_{F_{12}} &- \hat{\textbf{X}}^{L_2}_{F_{12}}\\
\end{aligned} \right]\right \|^2_{I^{L_2}}.
\end{equation}

This linear least squares problem can be written in a compact form as
\begin{equation}\label{Eq::LLSMJ}
\mbox{minimize~~} \bar{f}(\bar{\textbf{X}}^{G_{12}})=\| A \, \bar{\textbf{X}}^{G_{12}} - \textbf{Z} \|^2_{I_Z}
\end{equation}
where $\textbf{Z}$ is the constant vector combining the state estimates of the two maps
\begin{equation}\label{Eq::ObsFuncMJ}
\textbf{Z}= [\hat{\textbf{X}}^{L_1}, \; \hat{\textbf{X}}^{L_{2}}],
\end{equation}
$I_Z$ is the corresponding information matrix given by
\begin{equation}\label{Eq::MJ Information}
I_Z =
\mbox{diag}(I^{L_1}, I^{L_{2}}),
\end{equation}
$\bar{\textbf{X}}^{G_{12}}$ is the state vector represents
the global map defined in (\ref{Eq::SVL12}). The coefficient matrix $A$ is sparse and is given by
\begin{equation}\label{Eq::A_Pose_Feature}
A  =
\begin{bmatrix}
E\\
 &E\\
 & & & &E\\
 & & & & & &E\\
 & &E\\
 & & &E\\
 & & & & &E\\
 & & & & & &E\\
\end{bmatrix}
\end{equation}
where $E$ denotes the identity matrix with the size corresponding to the different
variables in the state vector $\bar{\textbf{X}}^{G_{12}}$.

The optimal solution $\hat{\bar{\textbf{X}}}^{G_{12}}$ to the linear least squares problem (\ref{Eq::LLSMJ}) can be obtained by solving the sparse linear equation
\begin{equation}\label{Eq::LLSMJSolveSt}
A^T I_Z A \; \hat{\bar{\textbf{X}}}^{G_{12}} = A^T I_Z \textbf{Z}.
\end{equation}
The corresponding information matrix can be computed by
\begin{equation}\label{Eq::LLSMJSolveI}
\bar{I}^{G_{12}} = A^T I_Z A.
\end{equation}

From (\ref{Eq::SVL12}), we have $\textbf{X}^{G_{12}} = g^{-1}(\bar{\textbf{X}}^{G_{12}})$. Note that the function $g^{-1}(\cdot)$ has a closed-form as follows:
\begin{equation}\label{Eq:transform-g}
\begin{aligned}
&\textbf{X}^{G_{12}} = g^{-1}(\bar{\textbf{X}}^{G_{12}}) \\
&\Rightarrow \left \{
\begin{aligned}
\textbf{t}^{G_{12}}_{0} &= \bar{R}_{2} \, (\bar{\textbf{t}}^{G_{12}}_{0}-\bar{\textbf{t}}^{G_{12}}_{2})\\
\textbf{r}^{G_{12}}_{0} &= r^{-1}(\bar{R}_{0} \bar{R}_{2}^T)\\
\textbf{t}^{G_{12}}_{1} &= -\bar{R}_{2} \, \bar{\textbf{t}}^{G_{12}}_{2}\\
\textbf{r}^{G_{12}}_{1} &= r^{-1}(\bar{R}_{2}^T)\\
\textbf{X}^{G_{12}}_{F_1} &= \bar{R}_{2} \, (\bar{\textbf{X}}^{G_{12}}_{F_1}-\bar{\textbf{t}}^{G_{12}}_{2})\\
\textbf{X}^{G_{12}}_{F_2} &= \bar{R}_{2} \, (\bar{\textbf{X}}^{G_{12}}_{F_2}-\bar{\textbf{t}}^{G_{12}}_{2})\\
\textbf{X}^{G_{12}}_{F_{12}} &= \bar{R}_{2} \, (\bar{\textbf{X}}^{G_{12}}_{F_{12}}-\bar{\textbf{t}}^{G_{12}}_{2})
\end{aligned}
\right.
\end{aligned}
\end{equation}
where $\bar{R}_{2} = r(\bar{\textbf{r}}^{G_{12}}_{2})$, $\bar{R}_{0} = r(\bar{\textbf{r}}^{G_{12}}_{0})$ are the rotation matrices of
pose $\textbf{P}_{2}$ and pose $\textbf{P}_{0}$.

Now the solution to the nonlinear least squares problem (\ref{Eq::ObjFunc_N12}) can be obtained by
\begin{equation}\label{Eq::Transg}
\hat{\textbf{X}}^{G_{12}} = g^{-1}(\hat{\bar{\textbf{X}}}^{G_{12}}).
\end{equation}

The corresponding information matrix $I^{G_{12}}$ can be obtained by
\begin{equation}\label{Eq::InfoM}
I^{G_{12}} = \nabla^T \,\bar{I}^{G_{12}} \, \nabla
\end{equation}
where $\nabla$ is the Jacobian of $\bar{\textbf{X}}^{G_{12}}$ with respect to
$\textbf{X}^{G_{12}}$, evaluated at $\hat{\textbf{X}}^{G_{12}}$
\begin{equation}\label{Eq::nabla}
\nabla = \frac{\partial g(\textbf{X}^{G_{12}})}{\partial
\textbf{X}^{G_{12}}}|_{\hat{\textbf{X}}^{G_{12}}}.
\end{equation}

\noindent\textbf{Remark 1.}  An intuitive explanation of the above process is shown in Fig.
\ref{fig:joining-two-maps-transform}. In fact, $\bar{\textbf{X}}^{G_{12}} = g(\textbf{X}^{G_{12}})$ in (\ref{Eq::NewSV}) and (\ref{Eq::SVL12}) is the coordinate transformation function,
transforming pose $\textbf{P}_{0}$, $\textbf{P}_{1}$ and feature
$\textbf{F}$ in the coordinate frame of $\textbf{P}_{2}$, into
$\textbf{P}_{0}$, $\textbf{P}_{2}$ and $\textbf{F}$ in the
coordinate frame of $\textbf{P}_{1}$. Thus $\textbf{X}^{G_{12}} = g^{-1}(\bar{\textbf{X}}^{G_{12}})$ in (\ref{Eq:transform-g}) has the closed-form formula which is another coordinate transformation, from $\textbf{P}_{0}$, $\textbf{P}_{2}$ and $\textbf{F}$ in the
coordinate frame of $\textbf{P}_{1}$, back to $\textbf{P}_{0}$, $\textbf{P}_{1}$ and
$\textbf{F}$ in the coordinate frame of $\textbf{P}_{2}$.
The linear way to solve the map joining problem is
equivalent to solving a linear least squares problem first and then performing a
coordinate transformation, as illustrated in Fig.
\ref{fig:joining-two-maps-transform}. The fact that the two maps are built in the same coordinate frame is the key to make this linear map joining approach possible.

The following lemma states rigorously that the map $M^{G_{12}}$ obtained using the linear method is equivalent to that obtained using the nonlinear method.

\noindent\textbf{Lemma 1:} Joining $M^{L_1}$ and $M^{L_2}$
to build the global map $M^{G_{12}}$ in the coordinate frame of the end pose of $M^{L_2}$, is equivalent to first building the global map $\bar{M}^{G_{12}}$ in the coordinate frame of the start pose of $M^{L_2}$, and then transforming the global map $\bar{M}^{G_{12}}$ into $M^{G_{12}}$ by coordinate transformation.

\begin{figure}
\centering
\includegraphics[height=85mm]{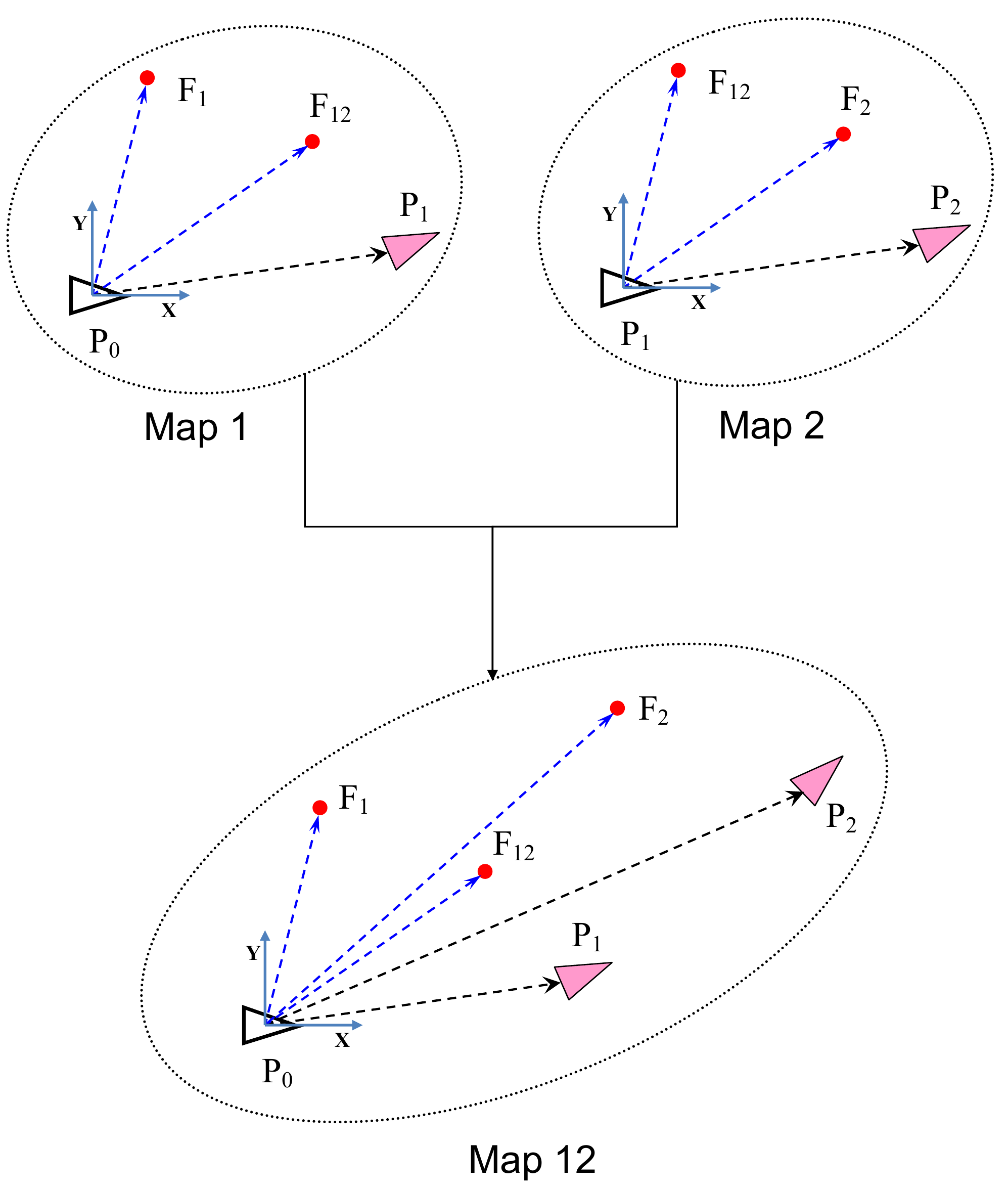}
\caption{Traditional way of joining two maps (nonlinear, e.g. in SLSJF \cite{Shoudong:SLSJF-TRO}).}
\label{fig:joining-two-localmaps-traditional}
\end{figure}

{\bf Proof:} See Appendix \ref{Appendix::A1}.

\subsection{Traditional Way of Building and Joining Two Maps}

As comparison, the basic step in traditional map joining (such as sequential \cite{Williams-PhD-thesis,Shoudong:SLSJF-TRO} or Divide and Conquer \cite{Divide-and-Conquer:-EKF-SLAM}) is also joining two maps, the process of which is illustrated in Fig. \ref{fig:joining-two-localmaps-traditional}.


In traditional map joining, Map 1 is built in the coordinate frame defined by its start pose $\textbf{P}_0$. Map 2 is built in the coordinate frame defined by its start pose $\textbf{P}_1$ (the end pose of Map 1). Comparing to (\ref{M1M2localmaps}) and (\ref{Eq::state vector estimate MJ}), Map 1 in traditional map joining is
\begin{equation}\label{TraditionalM1M2localmaps}
\begin{aligned}
M^{L_1^0} &= (\hat{\textbf{X}}^{L_1^0}, \; I^{L_1^0})\\
\hat{\textbf{X}}^{L_1^0} &= [\hat{\textbf{t}}^{L_1^0}_1, \; \hat{\textbf{r}}^{L_1^0}_1, \; \hat{\textbf{X}}^{L_1^0}_{F_1}, \; \hat{\textbf{X}}^{L_1^0}_{F_{12}}]\\
\end{aligned}
\end{equation}
where $\textbf{P}_{0}$ defines the coordinate frame of $M^{L_1^0}$, and thus is not included in the state vector of the Map 1. Map 2 $M^{L_2}$ is the same as used in the Linear SLAM in Section \ref{Subsec::N12}.

The coordinate frame of the combined map is defined by $\textbf{P}_0$, which is the same as Map 1 $M^{L_1^0}$. The state vector of the combined map in traditional map joining is
\begin{equation}
\textbf{X}^{G_{12}^0}  =
[\textbf{t}^{G_{12}^0}_{1}, \; \textbf{r}^{G_{12}^0}_{1}, \; \textbf{t}^{G_{12}^0}_{2}, \; \textbf{r}^{G_{12}^0}_{2}, \; \textbf{X}^{G_{12}^0}_{F_1}, \; \textbf{X}^{G_{12}^0}_{F_2}, \; \textbf{X}^{G_{12}^0}_{F_{12}}].
\end{equation}

Joining these two maps optimally is clearly a nonlinear optimization problem \cite{Shoudong:I-SLSJF-ACRA} because the coordinate frames of the two maps are different
\begin{equation}\label{Eq::ObjFunc_N12_Traditional}
\begin{aligned}
&\tilde{f}(\textbf{X}^{G_{12}^0}) = \\
&\left \| \left[
\begin{aligned}
\textbf{t}^{G_{12}^0}_{1} &- \hat{\textbf{t}}^{L_1^0}_1\\
\textbf{r}^{G_{12}^0}_{1} &- \hat{\textbf{r}}^{L_1^0}_1\\
\textbf{X}^{G_{12}^0}_{F_1} &- \hat{\textbf{X}}^{L_1^0}_{F_1}\\
\textbf{X}^{G_{12}^0}_{F_{12}} &- \hat{\textbf{X}}^{L_1^0}_{F_{12}}\\
\end{aligned} \right]\right \|^2_{I^{L_{1}^0}}
+\left \| \left[
\begin{aligned}
R_{1}^0 \, (\textbf{t}^{G_{12}^0}_{2}-\textbf{t}^{G_{12}^0}_{1}) &- \hat{\textbf{t}}^{L_2}_2\\
r^{-1}(R_{2}^0 \, {R_{1}^0}^T) &- \hat{\textbf{r}}^{L_2}_2\\
R_{1}^0 \, (\textbf{X}^{G_{12}^0}_{F_2}-\textbf{t}^{G_{12}^0}_{1}) &- \hat{\textbf{X}}^{L_2}_{F_2}\\
R_{1}^0 \, (\textbf{X}^{G_{12}^0}_{F_{12}}-\textbf{t}^{G_{12}^0}_{1}) &- \hat{\textbf{X}}^{L_2}_{F_{12}}\\
\end{aligned} \right]\right \|^2_{I^{L_2}}
\end{aligned}
\end{equation}
where $R_{1}^0 = r(\textbf{r}^{G_{12}^0}_{1})$ and $R_{2}^0 = r(\textbf{r}^{G_{12}^0}_{2})$ are the rotation matrices of poses $\textbf{P}_{1}$ and $\textbf{P}_{2}$ in the global state vector $\textbf{X}^{G_{12}^0}$, respectively.

Thus, in the traditional map joining algorithms, the vector in the first term of objective function (\ref{Eq::ObjFunc_N12_Traditional}) is linear, while the vector in the second term in (\ref{Eq::ObjFunc_N12_Traditional}) is nonlinear. And it is impossible to choose an intermediate state vector to make (\ref{Eq::ObjFunc_N12_Traditional}) linear.

\section{Joining A Sequence of Local Maps} \label{sec:sequence}

Based on the linear solution to joining two maps as shown in Fig.
\ref{fig:joining-two-localmaps-new}, Fig.
\ref{fig:joining-two-maps-transform} and Section \ref{testShoudong}, we can use either sequential
map joining or Divide and Conquer map joining to solve large-scale SLAM problems.

\begin{figure*}
\centering \subfigure[Proposed sequential map joining.]
{\label{fig:mapjoining-new}
\includegraphics[height=94mm]{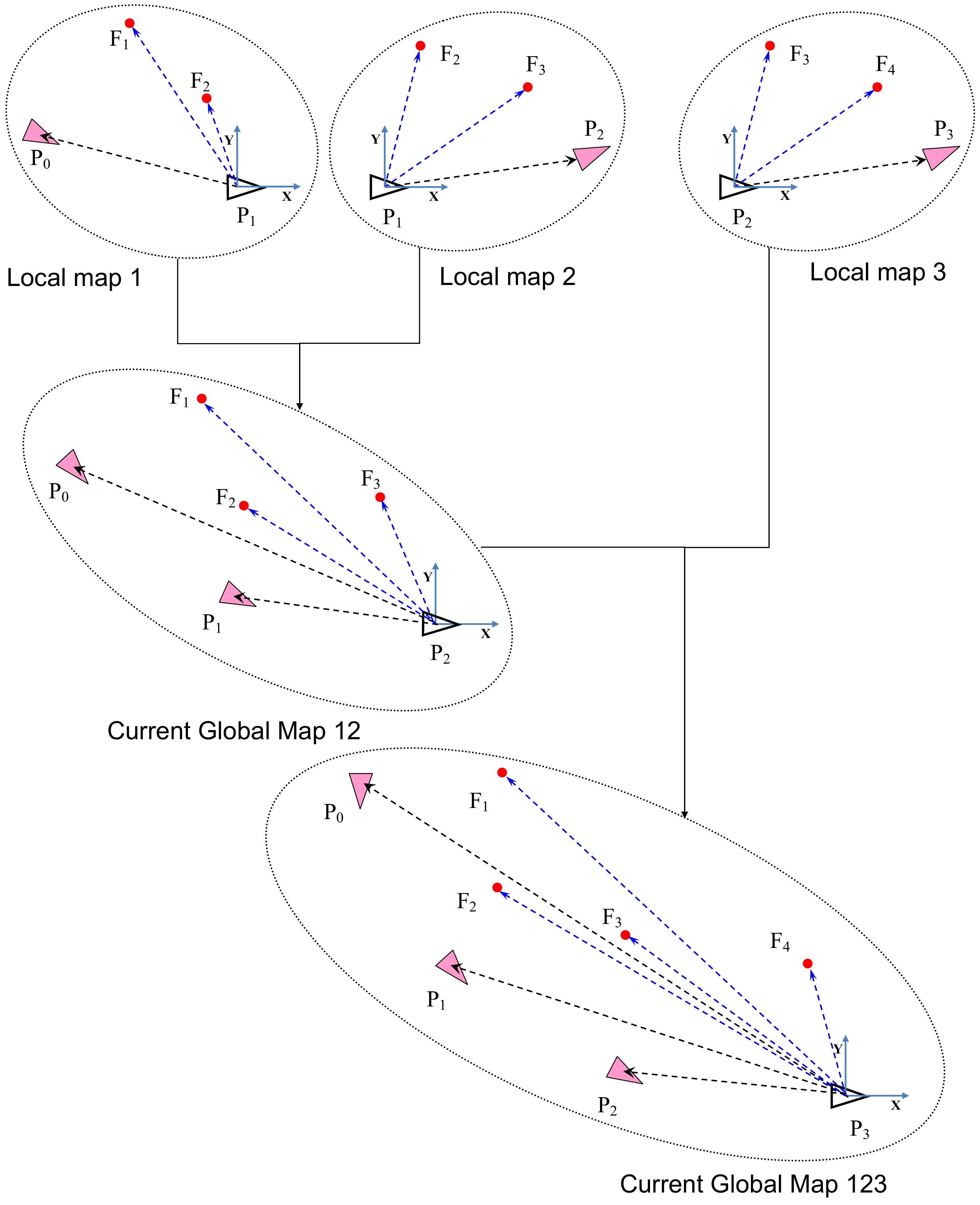}}
\hspace{1mm} \centering \subfigure[Proposed Divide and Conquer map
joining.] {\label{fig:mapjoining-new-DC}
\includegraphics[height=94mm]{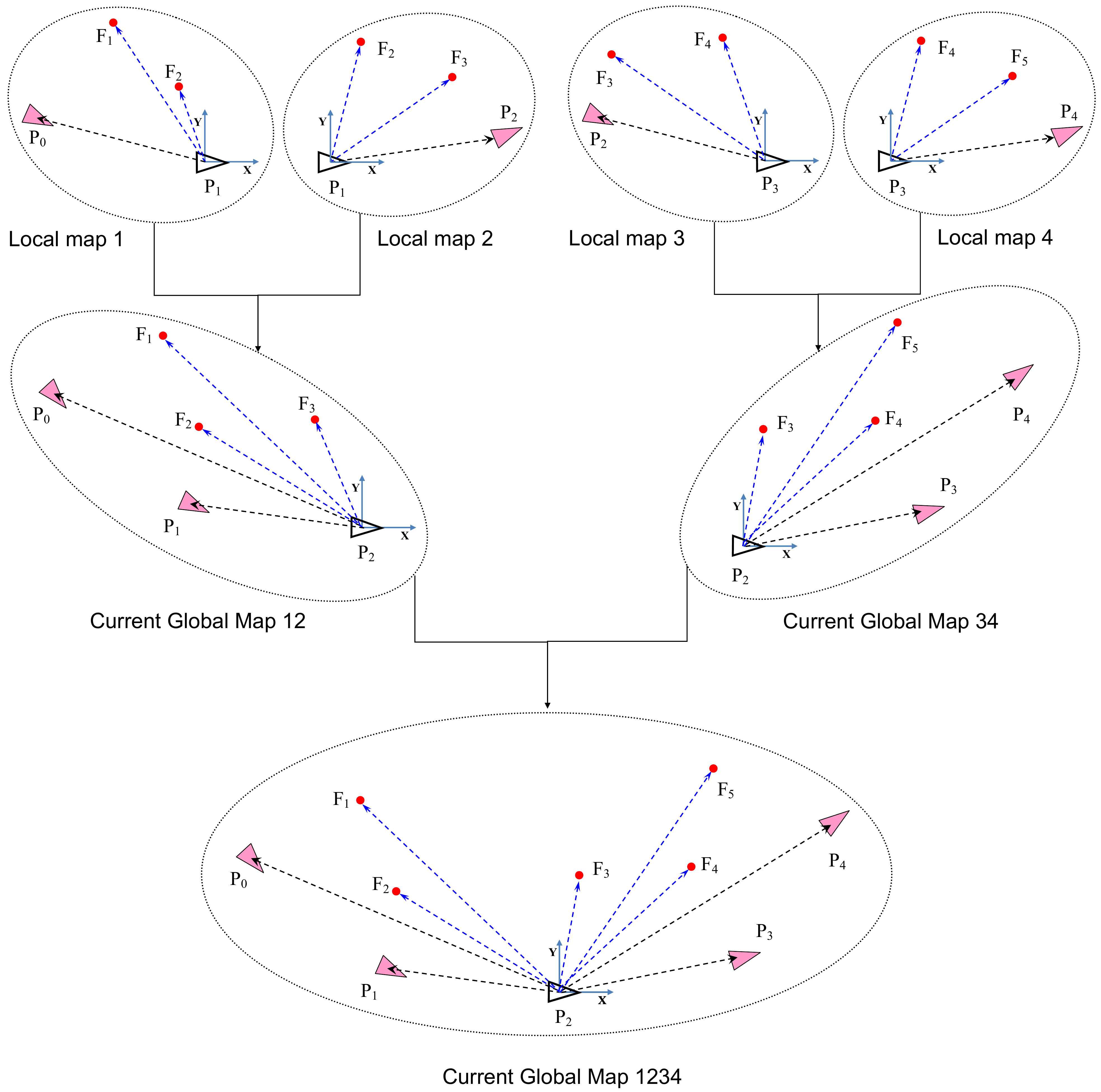}}
 \caption{Sequential map joining and Divide and Conquer map joining.} \label{fig:seqential-and-DC}
\end{figure*}

\subsection{Proposed Sequential Map Joining}

The new sequential map joining process we proposed can be illustrated in Fig. \ref{fig:mapjoining-new}. Please note: (1) Local map 1 is built in the coordinate frame of its end pose instead of its start pose; Thus it can be fused with Local map 2 using the linear method in Section \ref{testShoudong}. (2) Global map 12, the result of joining Local map 1 and Local map 2, is in the coordinate frame of the end pose of Local map 2. Thus it can be fused with Local map 3 using the linear method in Section \ref{testShoudong}.

As comparison, in traditional map joining algorithms (such as \cite{Williams-PhD-thesis,Shoudong:SLSJF-TRO}), all the initial local map 1, 2 and 3 are built in the coordinate frames of their start poses. And at each step, the global map (map 12 or map 123) is always built in the coordinate frame of its start pose which is the coordinate frame of map 1.

\subsection{Proposed Divide and Conquer Map Joining}

The new Divide and Conquer map joining process we proposed is illustrated in Fig. \ref{fig:mapjoining-new-DC}. Please note: (1) Local map 1 and Local map 2 are built in the same coordinate frame; Local map 3 and Local map 4 are built in the same coordinate frame; (2) Global map 12 is built in the
same coordinate frame as Global map 34. Thus Global map 12 and Global map 34 can be joined together using the linear method in Section \ref{testShoudong}.

While in traditional map joining (such as \cite{Divide-and-Conquer:-EKF-SLAM}), local map 1, 2, 3 and 4 are all in the coordinate frames of their start poses. Global map 12 is built in the same coordinate frame as local map 1. Global map 34 is built in the same coordinate frame as local map 3. And the final global map 1234 is built in the same coordinate frame as local map 1.

\noindent\textbf{Remark 2.} Local map 1 in Fig. \ref{fig:mapjoining-new} or Fig. \ref{fig:mapjoining-new-DC} can be simply built by performing a nonlinear least squares using all the observation and odometry data with state vector defined as the robot poses and feature positions in the coordinate frame of the robot end pose in the local map (see Fig. \ref{fig:localmap2}). Alternatively, we can first build the local map in the coordinate frame of the robot start pose $\textbf{P}_0$, and then apply a coordinate transformation.

\subsection{Computational Complexity}\label{Section Computational Complexity}

Suppose in the feature-based SLAM problem, the total number of feature/odometry observations is $O_G$ and the number of robot poses/features in the state vector is $S_G$. In each iteration, the nonlinear system is first linearized and the linear least squares to be solved in each iteration is similar to the problem in (\ref{Eq::LLSMJ}). Then, the computational complexity of one iteration for the full nonlinear least squares SLAM is $\Theta(O_G+S_G^3)$, where the computational complexity of computing the information matrix $J^T I_Z J$ in (\ref{Eq::LLSMJSolveSt}) (here $A$ is replaced by $J$, where $J$ is the Jacobian of the observations w.r.t the state vector) is proportional to the number of the observations $\Theta(O_G)$ \cite{Mouragnon_IVC2009}, and $\Theta(S_G^3)$ is the computational complexity of solving the traditional resolution of the linear system (\ref{Eq::LLSMJSolveSt}) \cite{Mouragnon_IVC2009}\footnote{The sparsity and ordering of the information matrix also have a significant impact on the computational cost of solving linear system. This is not discussed here.}. Thus, by considering the iteration number $m$, the computational complexity of solving nonlinear least squares SLAM is $\Theta(mO_G+mS_G^3)$.

For the proposed Linear SLAM algorithm, the computational cost consists of two parts: building the local maps and joining the local maps.

\begin{table*}
\tabcolsep 0pt \caption{Computational Complexity$^\dag$ of Linear SLAM and Nonlinear Map Joining with Different Local Map Numbers ($n$) in Comparison with Nonlinear Least Squares SLAM}
\label{Tab:computational complexity}
\footnotesize{Victoria Park dataset, $O_G=52288$, $S_G=7197$ and suppose $m=10$.}
\begin{center}
\begin{tabular*}{1.0\textwidth}{@{\extracolsep{\fill}} lccccccccc}
\hline\hline
\multicolumn{1}{c}{} &\multicolumn{1}{c}{} &\multicolumn{4}{c}{Linear SLAM} &\multicolumn{4}{c}{Nonlinear Map Joining} \;\\
\multicolumn{1}{c}{} &\multicolumn{1}{c}{} &\multicolumn{2}{c}{Sequential} &\multicolumn{2}{c}{Divide and Conquer} &\multicolumn{2}{c}{Sequential} &\multicolumn{2}{c}{Divide and Conquer} \;\\
\; $n$ & Local Map & Joining & Overall & Joining & Overall & Joining & Overall & Joining & Overall \;\\
\hline
\; 1* &1 &N/A  &N/A  &N/A  &N/A &N/A  &N/A  &N/A  &N/A  \;\\
\; 5 &0.0400 &0.1792  &0.2192  &0.1640  &0.2040 &1.7920  &1.8320  &1.6400  &1.6800  \;\\
\; 10 &0.0100 &0.3024  &0.3124  &0.1680  &0.1780 &3.0240  &3.0340  &1.6800 &1.6900  \;\\
\; 20 &0.0025 &0.5512  &0.5537  &0.1690  &0.1715 &5.5124  &5.5149  &1.6900  &1.6925  \;\\
\; 50 &4.0e-4 &1.3005  &1.3009  &0.1393  &0.1397 &13.005  &13.005  &1.3928  &1.3932  \;\\
\; 100 &1.0e-4 &2.5502  &2.5503  &0.1393  &0.1394 &25.502  &25.503  &1.3932  &1.3933  \;\\
\hline\hline
\end{tabular*}
\end{center}
\dag The computational complexity is relative to the nonlinear least squares SLAM.\\
*One local map means it is the nonlinear least squares SLAM to obtain the global map directly. The computational complexity is set to 1 as the benchmark for comparison.
\end{table*}

Suppose the observations and the robot poses/features in the state vector are equally divided to build $n$ local maps. Thus in each local map, there are $\frac{1}{n}O_G$ observations and $\frac{1}{n}S_G$ poses and features. Similar to building the global map described above, if each local map is built by using nonlinear least squares SLAM with $m$ iterations, the computational complexity of building $n$ local maps is $\Theta(\frac{1}{n}O_G+{(\frac{1}{n}S_G)}^3)\times m\times n = \Theta(mO_G+\frac{m}{n^2}S_G^3)$. Thus, the computational complexity of building $n$ local maps is much less than that of directly building the global map by nonlinear least squares SLAM.

For the linear joining of two local maps in Section \ref{sec:why-linear}, the computational complexity of computing the information matrix in (\ref{Eq::LLSMJ}) depends on the number of observations in the two local maps, which is $\Theta(\frac{2}{n}S_G)$. While, the computational complexities of solving the linear system (\ref{Eq::LLSMJSolveSt}) and transforming the joint map in (\ref{Eq::Transg}) and (\ref{Eq::InfoM}) both depends on the number of poses/features in the state vector of the joint map, which are $\Theta({(\frac{2}{n}S_G)}^3)$ and $\Theta(\frac{2}{n}S_G)$, respectively. Thus, the computational complexity of joining of two local maps using the proposed linear method is $\Theta(\frac{4}{n}S_G+\frac{8}{n^3}S_G^3)$. The iteration number $m$ is not applied here, since it is linear least squares optimization.

Thus, when joining a sequence of local maps in the sequential way, the overall computational complexity of the linear SLAM algorithm is
\begin{equation}\label{Eq::complexitysequential}
\Theta(mO_G+\frac{m}{n^2}S_G^3)+\sum_{i=2}^{n}\Theta(\frac{2i}{n}S_G+(\frac{i}{n}S_G)^3),
\end{equation}
where $i$ is the ID of the current local map to be joined.

While, in the Divide and Conquer manner, the overall computational complexity of the linear SLAM algorithm is
\begin{equation}\label{Eq::complexityDC}
\begin{aligned}
&\Theta(mO_G+\frac{m}{n^2}S_G^3) + \Theta(2S_G+S_G^3)\\ &+\sum_{k=1}^{c(\log_2(n))-1}\Theta(\frac{2^{k+1}}{n}S_G+(\frac{2^k}{n}S_G)^3) \times z(\frac{n}{2^k})
\end{aligned}
\end{equation}
where $k$ is the level in the Divide and Conquer process, $c(\cdot)$ is the function which round toward positive infinity and $z(\cdot)$ is the function which round toward zero. Thus, $c(\log_2(n))$ is the number of levels and $z(\frac{n}{2^k})$ is the number of times of performing joining two local maps at the $k^{th}$ level in the Divide and Conquer process.

In comparison, for the traditional nonlinear map joining algorithm using nonlinear least squares optimization, suppose the iteration number is also $m$ when joining each of two local maps, the computation cost in the sequential manner is
\begin{equation}\label{Eq::nonlinearcomplexitysequential}
\Theta(mO_G+\frac{m}{n^2}S_G^3)+\sum_{i=2}^{n}\Theta(\frac{mi}{n}S_G+m(\frac{i}{n}S_G)^3).
\end{equation}
And the computation cost of the traditional nonlinear map joining algorithms in the Divide and Conquer manner is
\begin{equation}\label{Eq::complexityDC}
\begin{aligned}
&\Theta(mO_G+\frac{m}{n^2}S_G^3) + \Theta(mS_G+mS_G^3)\\ &+\sum_{k=1}^{c(\log_2(n))-1}\Theta(\frac{2^{k}m}{n}S_G+m(\frac{2^k}{n}S_G)^3) \times z(\frac{n}{2^k}).
\end{aligned}
\end{equation}

The computational complexity of the proposed Linear SLAM algorithm and the traditional map joining algorithm using nonlinear least squares optimization with respect to the global nonlinear least squares SLAM are shown in Table \ref{Tab:computational complexity}. As an example of using Victoria Park dataset, performing Linear SLAM in the Divide and Conquer manner the computational complexity is always much less than that of global nonlinear least squares SLAM. While, in the sequential manner the computational complexity can be more if the number of local maps is large. The number and size of local maps also affect the computational complexity of the map joining algorithm. Ideally as shown in Table \ref{Tab:computational complexity}, the larger the number of local maps are built from the original dataset, the more the computational complexity is in the sequential manner, while the computational complexity will remain similar in the Divide and Conquer manner. While, as shown in Fig. \ref{fig:Impact}, the real computational cost also depends on the implementation of the algorithm, e.g. the memory operation in the implementation of Divide and Conquer.



\section{Joining Two Pose-only Maps}\label{sec:linear-pose-graph-SLAM}

Recently pose graph SLAM becomes popular where relative poses among the robot poses are computed using the sensor data and then an optimization is performed to obtain an estimate of the state vector containing only robot poses. The linear approach proposed in Section \ref{sec:why-linear} and Section \ref{sec:sequence} can also be applied to the joining of pose-only maps. A pose-only local map can be the result of solving a small pose graph SLAM or the marginalization (of all the features) from a pose-feature local map. In this section, we explain the process of joining two pose-only maps using linear least squares, which can be illustrated using Fig. \ref{fig:joining-two-localmaps-new} (by replacing all the features by poses).

\subsection{Local Pose-only Maps}\label{subsec:pose-graph-Maps}

Suppose there are two pose-only maps $M^{L_1}$
and $M^{L_2}$ given in the form of (\ref{M1M2localmaps})
where $\hat{\textbf{X}}^{L_1}$ and $\hat{\textbf{X}}^{L_2}$ are
the estimates of the state vectors $\textbf{X}^{L_1}$ and
$\textbf{X}^{L_2}$ defined as
\begin{equation}\label{Eq::state vector PG}
\begin{aligned}
\textbf{X}^{L_1} &= (\textbf{X}^{L_1}_{P_{L1}}, \; \textbf{X}^{L_1}_{P_{L12}})\\
\textbf{X}^{L_2} &= (\textbf{X}^{L_2}_{P_{L2}}, \; \textbf{X}^{L_2}_{P_{L12}})\\
\end{aligned}
\end{equation}
and $I^{L_1}$ and $I^{L_2}$ are the associated information matrices. Both
$M^{L_1}$ and $M^{L_2}$ are in the coordinate frame of
$\textbf{P}_{1}$.

In the state vector $\textbf{X}^{L_1}$ and $\textbf{X}^{L_2}$ in
(\ref{Eq::state vector PG}), $\textbf{X}^{L_1}_{P_{L12}}$ and
$\textbf{X}^{L_2}_{P_{L12}}$ are the common poses between $M^{L_1}$
and $M^{L_2}$, while $\textbf{X}^{L_1}_{P_{L1}}$ and
$\textbf{X}^{L_2}_{P_{L2}}$ are the poses that appear in only one of the maps. All the poses can be
defined similarly as (\ref{Eq::Pose P0 MJ}) in Section
\ref{Subsec::N12}.

\subsection{Joining Two Pose-only Maps by Linear Least Squares}

Joining these two pose-only maps $M^{L_1}$ and $M^{L_2}$ to build the intermediate
map $\bar{M}^{G_{12}} = (\hat{\bar{\textbf{X}}}^{G_{12}},
\; \bar{I}^{G_{12}})$ can also be solved by linear least squares. Here the global state vector $\bar{\textbf{X}}^{G_{12}}$ is defined as
\begin{equation}\label{Eq:GlobalStateVectorPG}
\bar{\textbf{X}}^{G_{12}} = (\bar{\textbf{X}}^{G_{12}}_{P_{L1}}, \; \bar{\textbf{X}}^{G_{12}}_{P_{L2}}, \; \bar{\textbf{X}}^{G_{12}}_{P_{L12}})
\end{equation}
where $\bar{\textbf{X}}^{G_{12}}_{P_{L12}}$ are the common poses,
$\bar{\textbf{X}}^{G_{12}}_{P_{L1}}$ and
$\bar{\textbf{X}}^{G_{12}}_{P_{L2}}$ are the poses that only appear in $M^{L_1}$ and $M^{L_2}$, respectively. All the variables in the global state vector $\bar{\textbf{X}}^{G_{12}}$ are in the coordinate frame of $\textbf{P}_1$. This intermediate map $\bar{M}^{G_{12}}$ can then be transformed into $M^{G_{12}}$ by the nonlinear coordinate transformation, where $M^{G_{12}}$ is defined in the coordinate frame of $\textbf{P}_2$.

\noindent\textbf{Remark 3.} It should be mentioned that, for the pose-feature map joining, the only common pose between the two local maps defines the coordinate frames of the two local maps as shown in Fig. \ref{fig:joining-two-localmaps-new}; this pose is not in the state vectors of the two local maps. So there is no common pose in the observations $\textbf{Z}$ in (\ref{Eq::ObsFuncMJ}). While, for the pose-only map joining problem, there can be many common poses between two local maps, and the wraparound of the rotation angles of the common poses must be carefully considered when performing map joining. As pointed out in \cite{Carlone2011,Carlone_TRO2014}, wraparound of the orientation angles is a critical issue in the linear formulation. However, as only two maps are joined each time in the proposed Linear SLAM algorithm, it is sufficient to wrap the angle in one of the two observations of a common pose in $\textbf{Z}$ to make the difference between the two angles in the observations fall into $(-\pi,\pi]$.

\begin{figure}
\centering {\includegraphics[height=85mm]{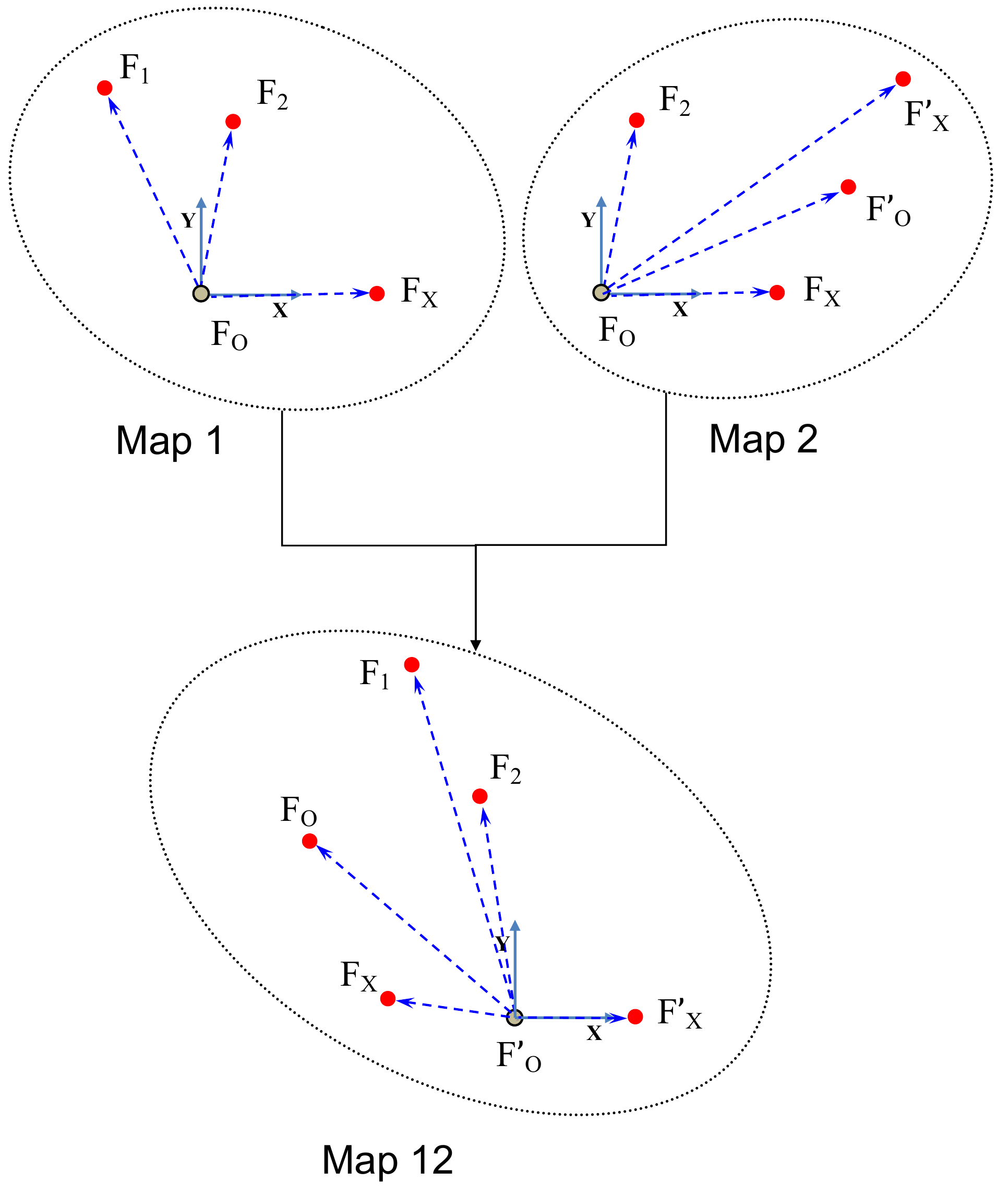}}
\caption{The proposed way of building and joining two feature-only maps. Map 1 and Map 2 are both in the coordinate frame defined by Feature $\textbf{F}_{O}$ and $\textbf{F}_{X}$, while Map 12 is in the coordinate frame defined by Feature $\textbf{F}'_{O}$ and $\textbf{F}'_{X}$.} \label{fig:joining-two-DSLAM}
\end{figure}

\section{Joining Two Feature-only Maps}\label{sec:linear-DSLAM}

Apart from traditional feature-based SLAM and pose graph SLAM, an alternative approach for SLAM is decoupling the localization and mapping (D-SLAM \cite{Wang:DSLAM-IJRR}). In the D-SLAM mapping part, the map only consists of a number of features. In this paper, we call this kind of maps ``feature-only maps". A feature-only map can also be obtained by marginalizing out all the poses from a pose-feature map (marginalizing out $\textbf{P}_0$, $\textbf{P}_1$ and $\textbf{P}_2$ in the map in Fig. \ref{fig:localmap3}). In this paper, we propose to use Linear SLAM to join 2D or 3D feature-only maps by first extending the idea in \cite{Wang:DSLAM-IJRR} and \cite{Huang:DSLAM-AR} into 3D case.

\subsection{Local Feature-only Maps}

Suppose there are two feature-only maps $M^{L_1}$
and $M^{L_2}$ given in the form of (\ref{M1M2localmaps}),
where the state vectors $\textbf{X}^{L_1}$ and
$\textbf{X}^{L_2}$ of each map are defined similarly to those in Section \ref{subsec:pose-graph-Maps}, by replacing the poses by features.

In feature-only maps, there is no robot pose in the map. Instead, two features for 2D \cite{Wang:DSLAM-IJRR} and three features for 3D are used to define the coordinate frame of the map. The details of the coordinates definition are described in Appendix \ref{Appendix::A2}.

When joining two feature-only maps, at least two common features for 2D (three common features for 3D) are required to make sure there are enough information to combine the two maps together \cite{Huang:DSLAM-AR}.

\begin{figure*}
\centering \subfigure[VicPark200] {\label{fig:VicPark200}
\includegraphics[width=0.3\textwidth,height=0.2\textheight]{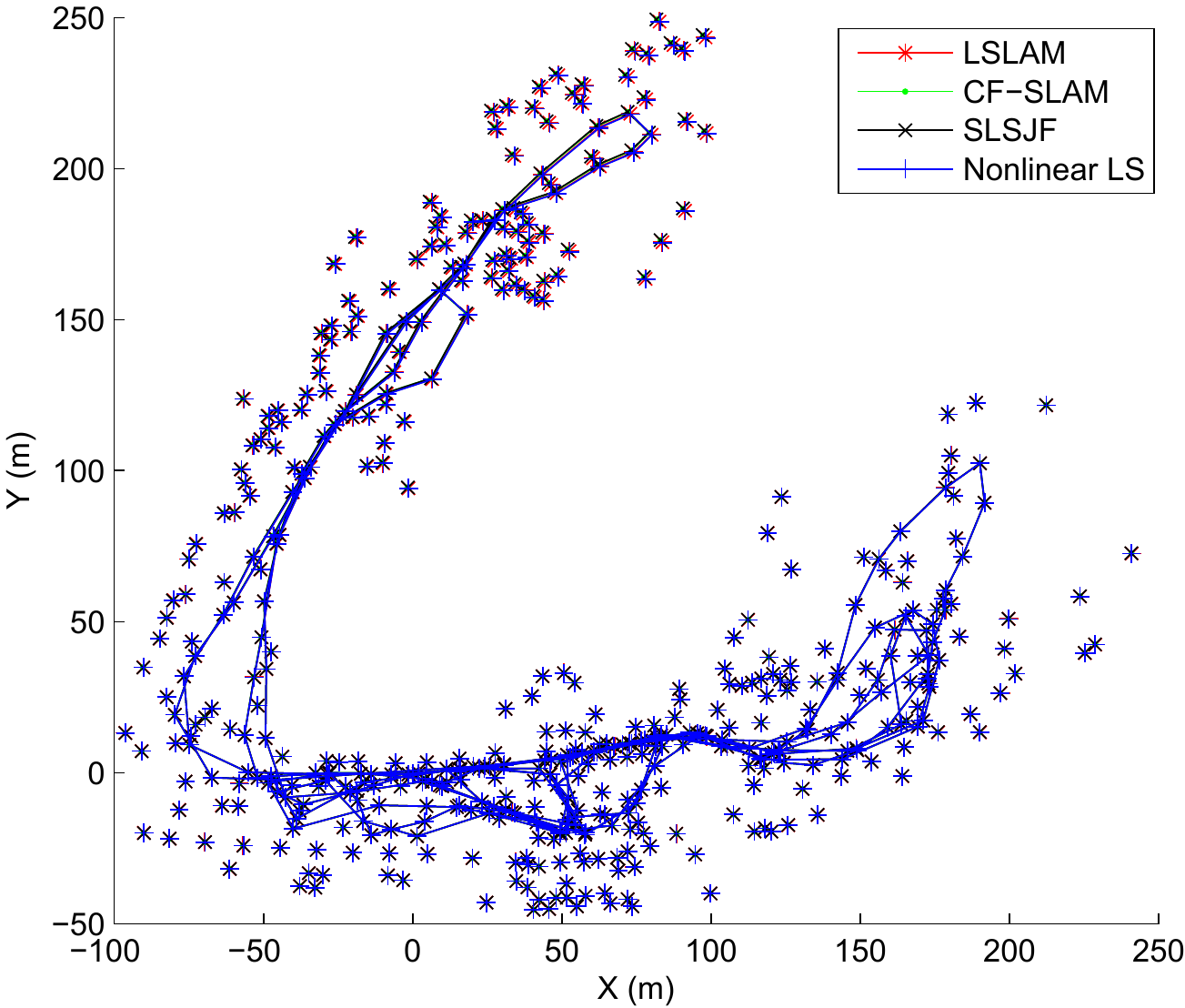}}
\centering \subfigure[DLR200] {\label{fig:DLR200}
\includegraphics[width=0.3\textwidth,height=0.2\textheight]{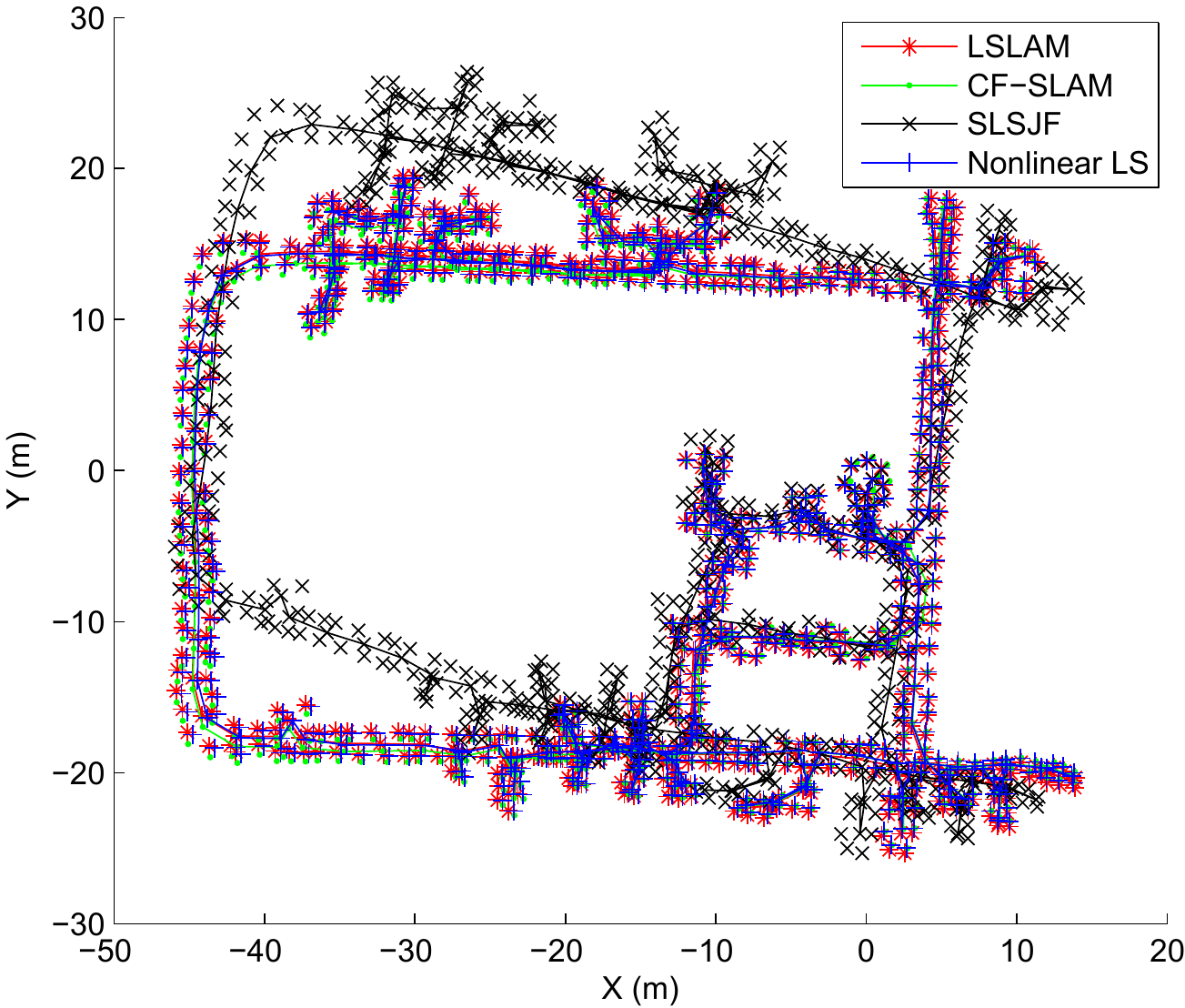}}
\centering \subfigure[VicPark6898] {\label{fig:VicPark6898}
\includegraphics[width=0.3\textwidth,height=0.2\textheight]{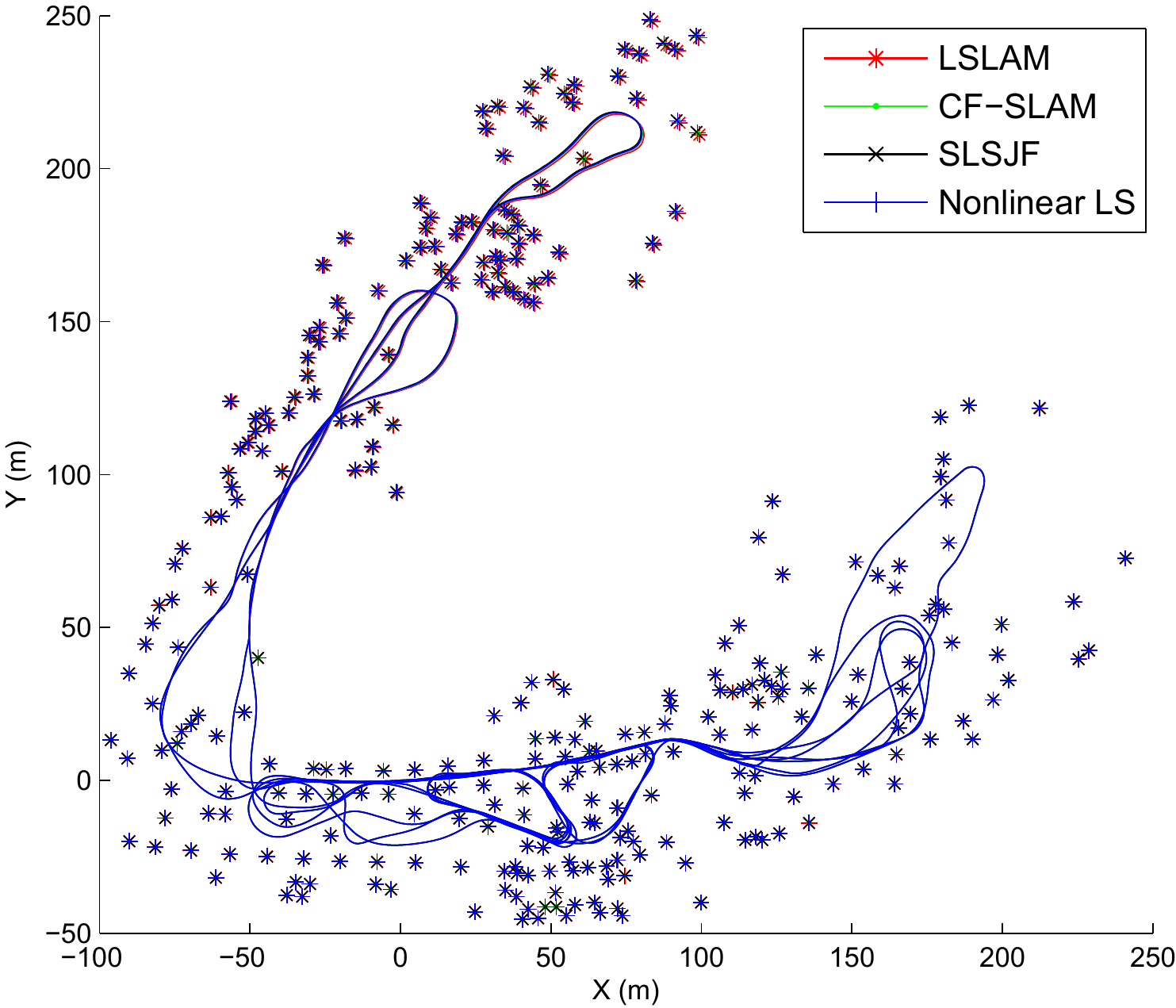}}
\centering \subfigure[DLR3298] {\label{fig:DLR3298}
\includegraphics[width=0.3\textwidth,height=0.2\textheight]{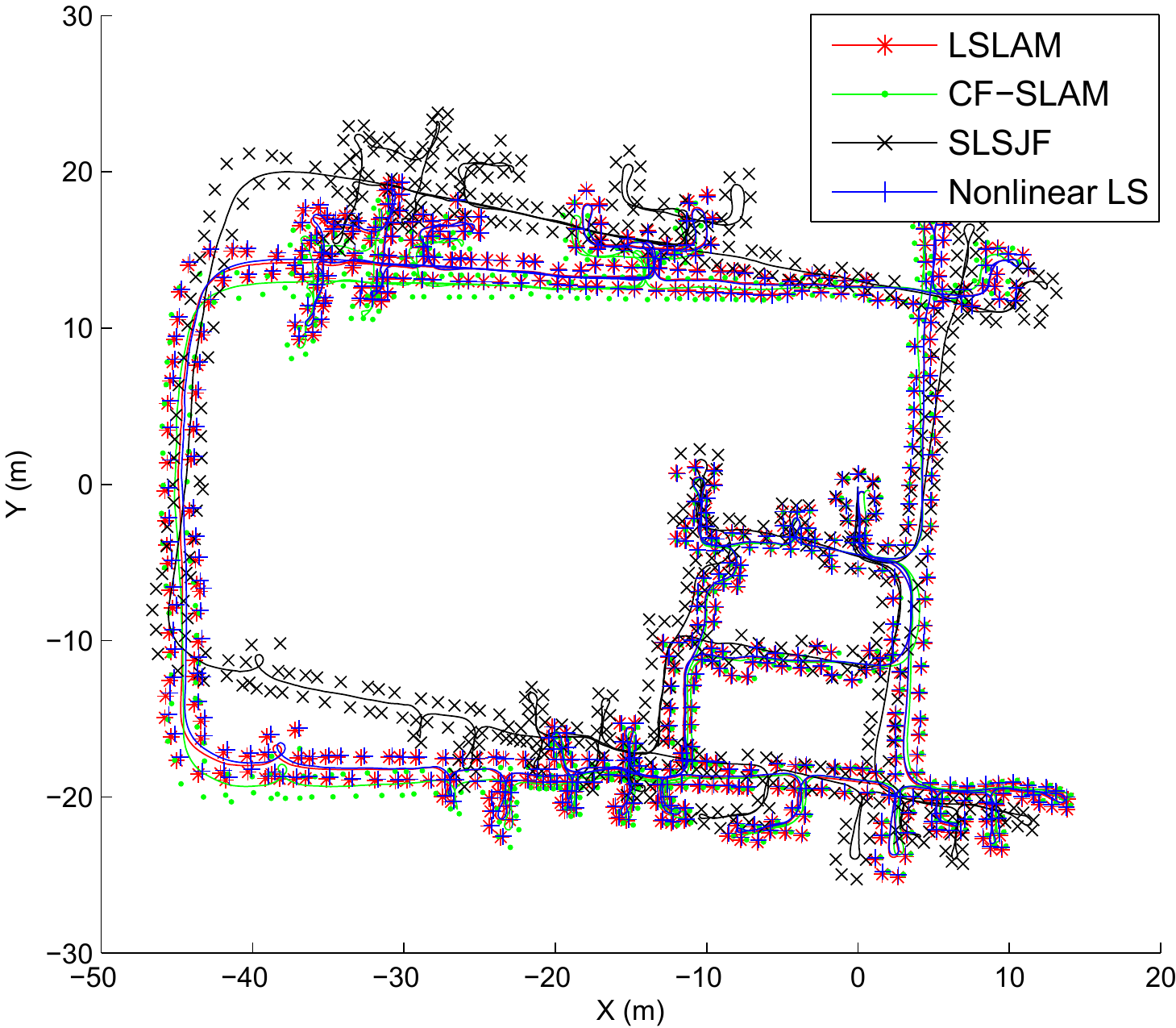}}
\centering \subfigure[3D Simu871] {\label{fig:3D Simu870}
\includegraphics[width=0.3\textwidth,height=0.2\textheight]{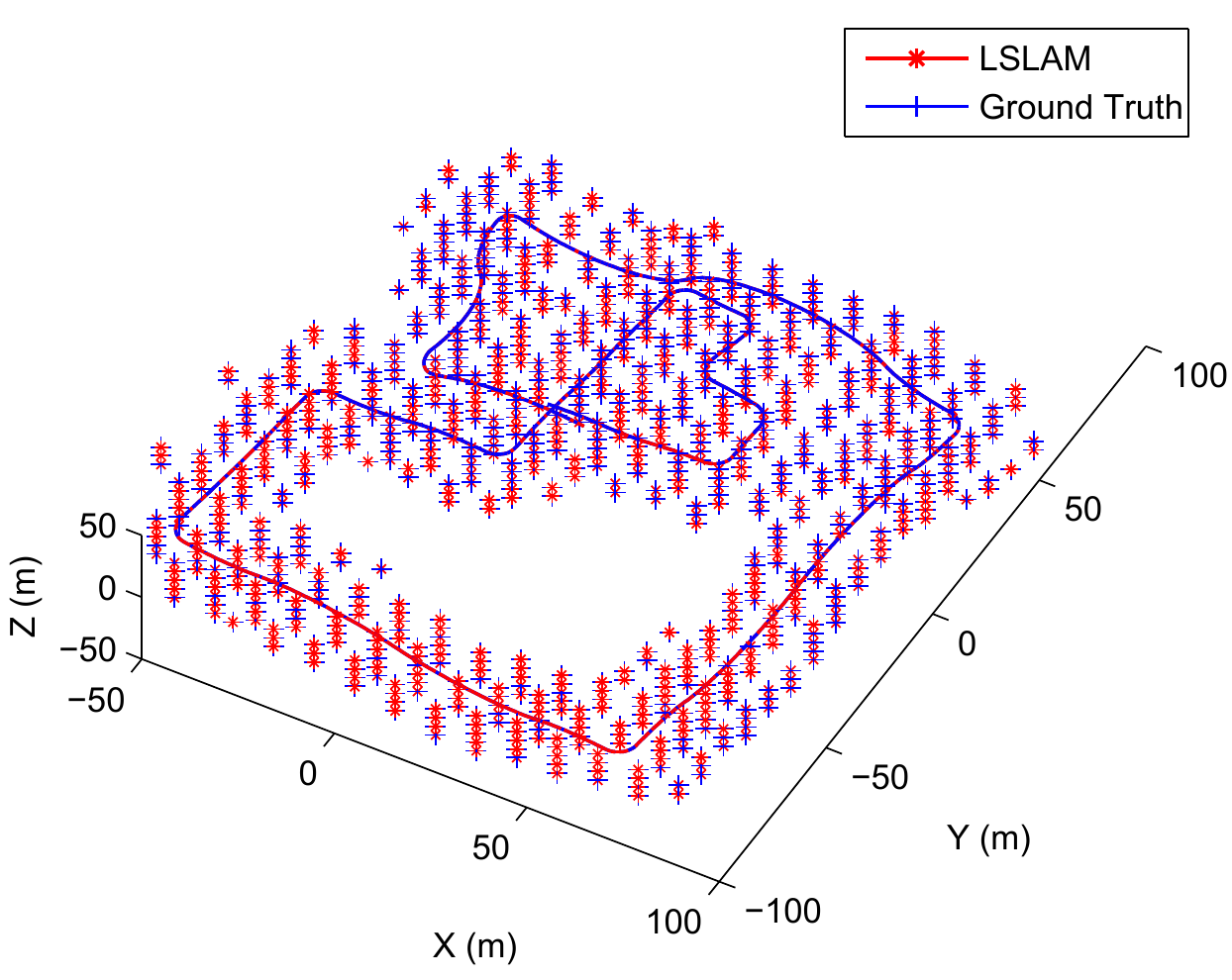}}
 \caption{2D and 3D pose-feature map joining results of Linear SLAM (LSLAM). Fig. \ref{fig:VicPark200} and Fig. \ref{fig:DLR200} are the results of joining the 200 local maps for 2D Victoria Park and DLR datasets. Fig. \ref{fig:VicPark6898} and Fig. \ref{fig:DLR3298} are the results of Victoria Park and DLR datasets by joining 6898/3298 local maps. The 2D results are compared with CF-SLAM, T-SAM, SLSJF and nonlinear least squares optimization. Fig. \ref{fig:3D Simu870} is the result of a 3D simulation with 870 local maps. It is only compared to the ground truth as the algorithms compared above are only applicable for 2D case. The quantitive comparison is given in Table \ref{Tab:RMSEFeatureBased}.} \label{fig:FeatureBased}
\end{figure*}

As shown in Fig. \ref{fig:joining-two-DSLAM}, for the proposed map joining approach, both $M^{L_1}$ and $M^{L_2}$ are in the same coordinate frame defined by the common features between them.

\subsection{Joining Two Feature-only Maps by Linear Least Squares}

The goal is to join the two feature-only maps $M^{L_1}$ and $M^{L_2}$ to build the feature-only map $M^{G_{12}} = (\hat{\textbf{X}}^{G_{12}}, \; I^{G_{12}})$ as shown in Fig. \ref{fig:joining-two-DSLAM}. Similar to the joining of pose-feature maps, the intermediate map $\bar{M}^{G_{12}} = (\hat{\bar{\textbf{X}}}^{G_{12}}, \; \bar{I}^{G_{12}})$ which is in the same coordinates as $M^{L_1}$ and $M^{L_2}$, is first built by linear least squares, and then it is transformed into $M^{G_{12}}$.

\noindent\textbf{Remark 4.} The process of building $\bar{M}^{G_{12}}$ and the transformation from $\bar{M}^{G_{12}}$ to $M^{G_{12}}$ is similar to those described in Section \ref{testShoudong}. The transformation of the state vector of feature-only maps is different from that of pose-feature maps or pose-only maps since the coordinate frame is not defined by a pose. The details of the definition and the transformation of the coordinate frames in feature-only maps in both 2D and 3D cases are described in Appendix \ref{Appendix::A2}.

\section{Simulation and Experimental Results Using Benchmark Datasets}\label{sec:results}

In this section, 2D and 3D simulation and real datasets of feature-based SLAM and pose graph SLAM have been used to evaluate the proposed Linear SLAM algorithm. All the Linear SLAM results presented below are from the Divide and Conquer map joining which is computationally less expensive (Section \ref{Section Computational Complexity}) than the sequential map joining. All the experiments are run on an Intel i5-3230M@2.6GHz CPU on a laptop.

\subsection{Pose-feature Map Joining}\label{Sec:PoseFeatureResult}

\subsubsection{2D Pose-feature Map Joining}

For 2D pose-feature map joining, the Victoria Park dataset \cite{Guivant01} and the DLR dataset \cite{DLR-data} are used.  Here, VicPark200(6898) / DLR200(3298) means the 200(6898) / 200(3298) local maps built from the Victoria Park / DLR dataset. These local map datasets are all available on OpenSLAM under project 2D-I-SLSJF.

\begin{table*}
\tabcolsep 0pt \caption{RMSE* and $\chi^2$ of Different Feature-Based SLAM Algorithms}
\label{Tab:RMSEFeatureBased}
\begin{center}
\begin{tabular*}{1.00\textwidth}{@{\extracolsep{\fill}} lccccccccccccc}
\hline\hline
\multicolumn{1}{c}{} &\multicolumn{3}{c}{CF-SLAM} &\multicolumn{3}{c}{T-SAM} &\multicolumn{3}{c}{SLSJF} &\multicolumn{3}{c}{Linear SLAM} & \multicolumn{1}{c}{Nonlinear LS}\;\\
\; Dataset & Pose & Feature & $\chi^2$ & Pose & Feature & $\chi^2$ & Pose & Feature & $\chi^2$ & Pose & Feature & $\chi^2$  & $\chi^2$ \;\\
\hline
\; VicPark200 &0.1949 &0.3191 & 3654 & & & & 0.2154 & 0.3599 & \textbf{3644} & \textbf{0.0391} & \textbf{0.0643} & 3645 & 3643 \;\\
\; DLR200 & 0.2391 & 0.2400 & 34480 & & & & 3.6574	& 3.6773 & 14603 & \textbf{0.1537} & \textbf{0.1545} & \textbf{12577} & 11164 \;\\
\; VicPark6898 & \textbf{0.0330} & \textbf{0.0509} & 9098 & 0.9329 & 1.4080 &117229 & 0.0718 & 0.1169 & \textbf{9013} & 0.1401 & 0.2348 & 9021 & 9013 \;\\
\; DLR3298 & 0.6308 & 0.6339 & 85827 & 0.6064 & 0.5976 & 112474 & 2.5066 & 2.5449 & \textbf{28316} & \textbf{0.0976} & \textbf{0.1042} & 28373 & 27689 \;\\
\hline\hline
\end{tabular*}
\end{center}
* RMSE (in meters) of the position of poses and features is w.r.t. the results of the full nonlinear least squares optimization (Nonlinear LS). For VicPark200 and DLR200 datasets, the Nonlinear LS means the optimization with all the local maps together as the observations. For VicPark6898 and DLR3298 datasets, the Nonlinear LS is the optimization with all the original observations. Because the building and joining submaps in T-SAM is different from the ones in the others, the RMSE and $\chi^2$ of the results from T-SAM are w.r.t. the Nonlinear LS with the original observations.
\end{table*}

\begin{figure}
\centering \subfigure[Computational Cost] {\label{fig:Time}
\includegraphics[width=0.45\textwidth]{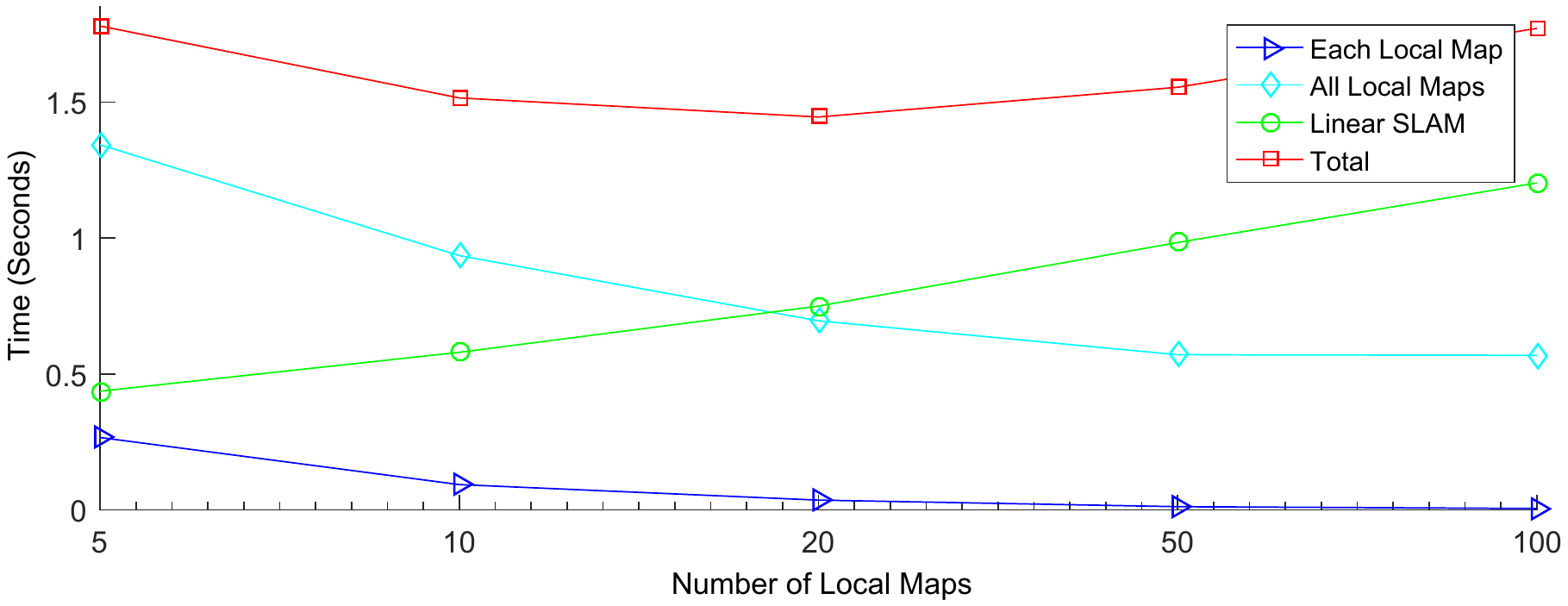}}
\centering \subfigure[Accuracy] {\label{fig:Accuracy}
\includegraphics[width=0.45\textwidth]{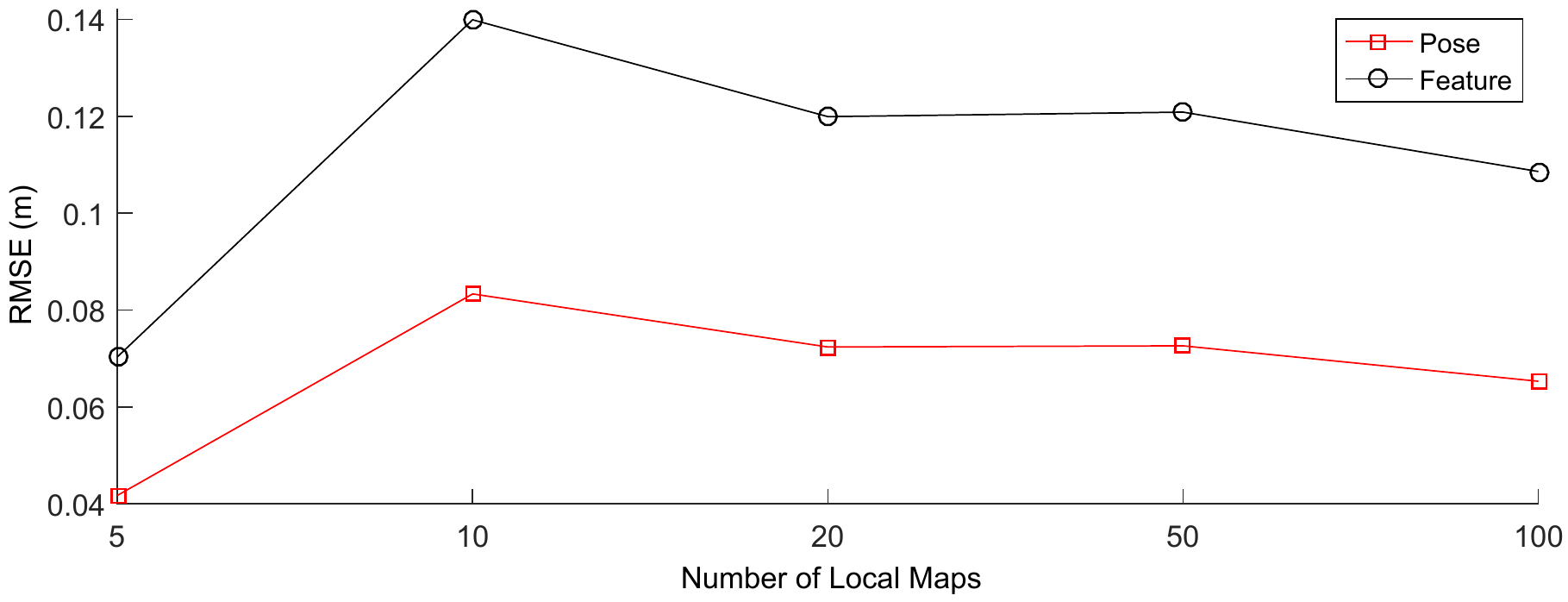}}
 \caption{The impact on the accuracy and computational cost by the number of local maps using the Victoria Park dataset.} \label{fig:Impact}
\end{figure}

The results of Linear SLAM using the four local map datasets are shown in Fig. \ref{fig:VicPark200} to Fig. \ref{fig:DLR3298}. The computational time are 1.42s, 0.47s, 2.83s and 0.85s, respectively. For comparison, the results from CF-SLAM \cite{Combined-filter}, T-SAM \cite{TSAMICRA07}, SLSJF \cite{Shoudong:SLSJF-TRO} and nonlinear least squares optimization (Nonlinear LS) algorithms are also shown in the figures. The root-mean-square error (RMSE) of Linear SLAM and the other three algorithms as compared with the optimal nonlinear LS results are shown in Table \ref{Tab:RMSEFeatureBased}. The final $\chi^2$ errors of the different algorithms are also compared in Table \ref{Tab:RMSEFeatureBased}.

For VicPark200 and DLR200 datasets, since the original inputs are the local maps (in the EKF SLAM format), we use I-SLSJF \cite{Shoudong:I-SLSJF-ACRA} (full nonlinear least squares with all the local maps as observations) as implementation for the Nonlinear LS. For VicPark6898 and DLR3298 datasets, since each local map was directly obtained using the observations of features from one pose together with the odometry to the next pose, the I-SLSJF results using these two datasets are equivalent to those using full least squares optimization with all the original observations and odometry information. Also for these two datasets, the whole Linear SLAM algorithm only requires linear least squares (plus nonlinear coordinate transformations) because there is no nonlinear optimization process for local map building involved.

For the comparison, in CF-SLAM, the inputs are also the local maps datasets from 2D-I-SLSJF on OpenSLAM, which are the same as the ones used in Linear SLAM and SLSJF. For T-SAM, because the building and joining of local maps are different from SLSJF, CF-SLAM and Linear SLAM, in our implementation, 200 local maps are built from the original observations and odometries for Victoria Park and DLR datasets in the way proposed in \cite{TSAMICRA07} (the way of cutting original dataset into local maps for T-SAM may not be optimal, and a better cut may further improve the results of T-SAM. However, it is not discussed in this paper). And as all the robot poses are kept in T-SAM, the accuracy and $\chi^2$ of the results are w.r.t. the Nonlinear LS with the original observations.

As shown in Fig. \ref{fig:VicPark200} to Fig. \ref{fig:DLR3298} and Table \ref{Tab:RMSEFeatureBased}, for the pose-feature map joining, in terms of RMSE, three out of four results of Linear SLAM are better than those of CF-SLAM and SLSJF, and all the four results are better than those of T-SAM, while nearly the same as the optimal Nonlinear LS results. The $\chi^2$ of Linear SLAM results are smaller than those of CF-SLAM, similar to those from the SLSJF results, and very close to that of the optimal nonlinear least squares solutions.

\subsubsection{Impact of the Size of the Local Maps}

The impact of the size of local maps on the accuracy and computational cost by the proposed Linear SLAM algorithm is illustrated using the Victoria Park dataset. The result is shown in Fig. \ref{fig:Impact}. The time used for building all the local maps decreases when the number of local maps increases from 5 to 100. For the map joining, the computational time slightly increases as the number of local maps increases due to the property of the hierarchical Divide and Conquer process. The total time of local map building plus map joining remains similar, in which the smallest is when the number of local maps is 20. In terms of the accuracy, the result does not vary much when the number of local maps is changed from 5 to 100. The RMSEs of both the poses and features are the smallest when there are only 5 local maps. It should be noted that the impact of the size of local maps can be different for different datasets due to the different number of observations and the different number of poses/features.

\begin{figure}
\centering \subfigure[2D Simu8240 dataset with 50 local maps] {\label{SimuNEES_a}
\includegraphics[height=70mm]{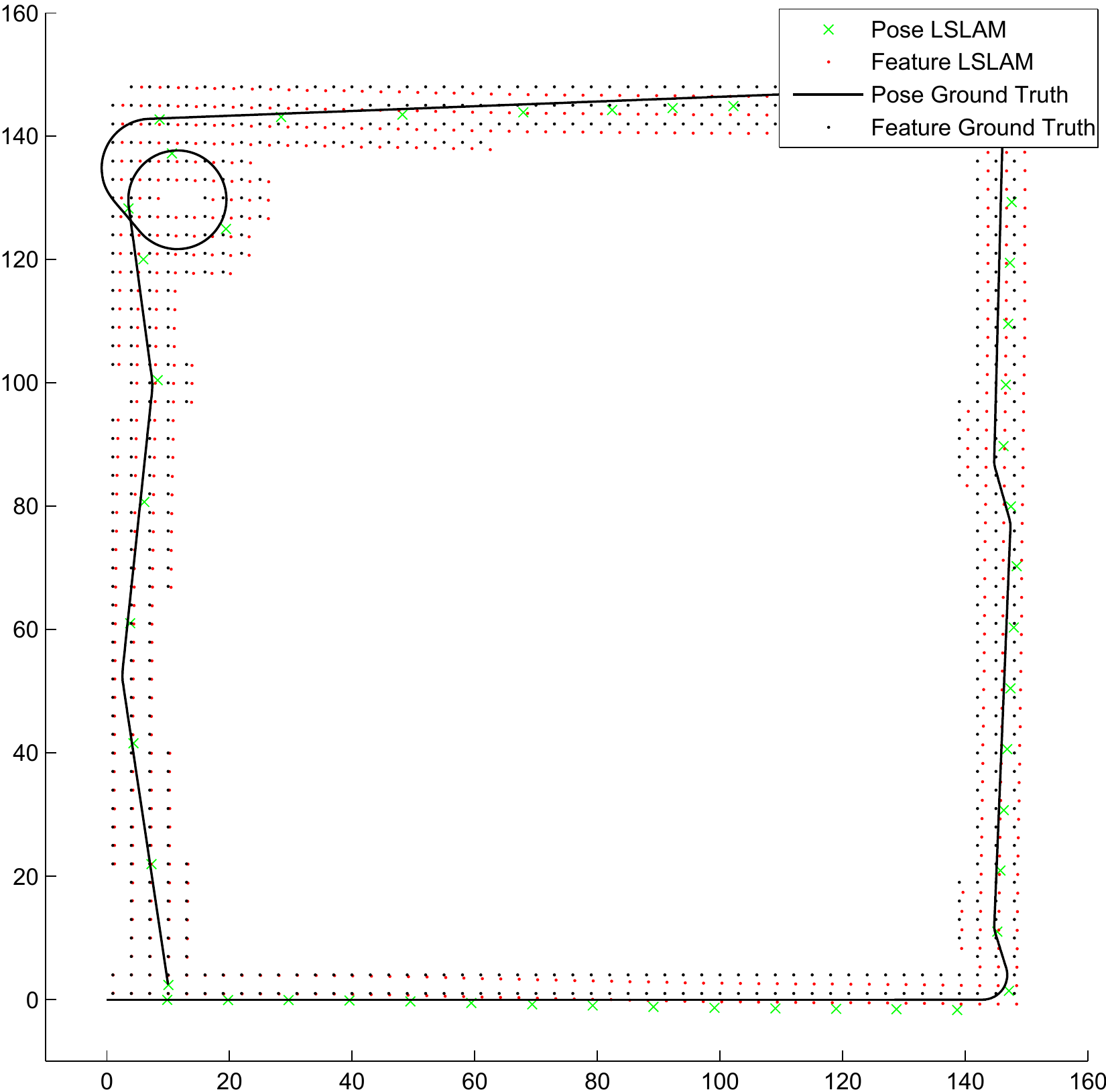}}
\centering \subfigure[2D Simu35188 dataset with 700 local maps]{\label{SimuNEES_b}
\includegraphics[height=70mm]{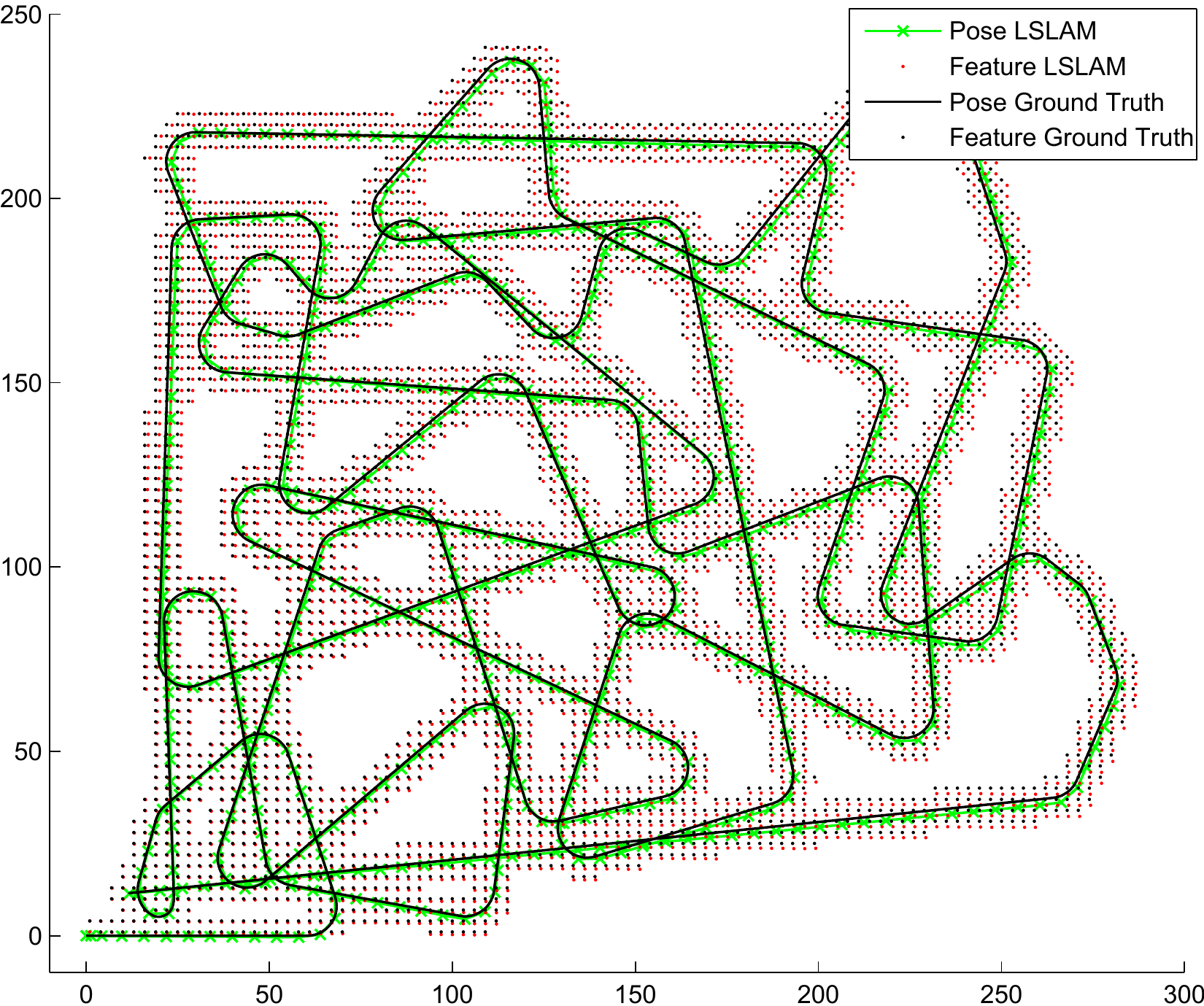}}
\caption{2D feature-based SLAM simulations and Linear SLAM
results for consistency check.} \label{Fig:SimuNEES}
\end{figure}

\begin{table}
\tabcolsep 0pt \caption{Consistency of Linear SLAM Results by
NEES Check (95\%)} \label{Tab:NEES Check}
\begin{center}
\def\temptablewidth{0.50\textwidth}
\begin{tabular*}{\temptablewidth}{@{\extracolsep{\fill}}lccc}
\hline\hline
\quad Datasets & Dimension & 95\% bound & NEES\\
\hline
\quad Simu8240 & 1224 & 1.3065e+03 & 1.2392e+03 \qquad \\
\quad Simu35188 & 8404 & 8.6184e+03 & 7.9163e+03 \qquad \\
\hline\hline
\end{tabular*}
\end{center}
\end{table}

\subsubsection{Consistency Analysis}\label{Section Simulation 3}

Two 2D simulation datasets Simu8240 (with 50 local maps) and Simu35188 (with 700 local maps) which are available on OpenSLAM (project 2D-I-SLSJF) are used to check the
consistency of the proposed Linear SLAM algorithm for pose-feature map joining. Fig. \ref{SimuNEES_a} and Fig. \ref{SimuNEES_b} illustrate the simulation scenarios (the ground truth) and the results of pose-feature map joining using Linear SLAM.

All the features present in the final global map and the corresponding information matrix were used to compute the normalised estimation error squared (NEES) \cite{Bailey-consistency} for checking the consistency \cite{Shoudong-Convergence} of Linear SLAM algorithm. The results shown in Table \ref{Tab:NEES Check} indicate that the Linear SLAM results are consistent.

\begin{figure*}
\centering \subfigure[Feature uncertainty] {\label{fig:Un_VP1}
\includegraphics[width=0.3\textwidth,height=0.2\textheight]{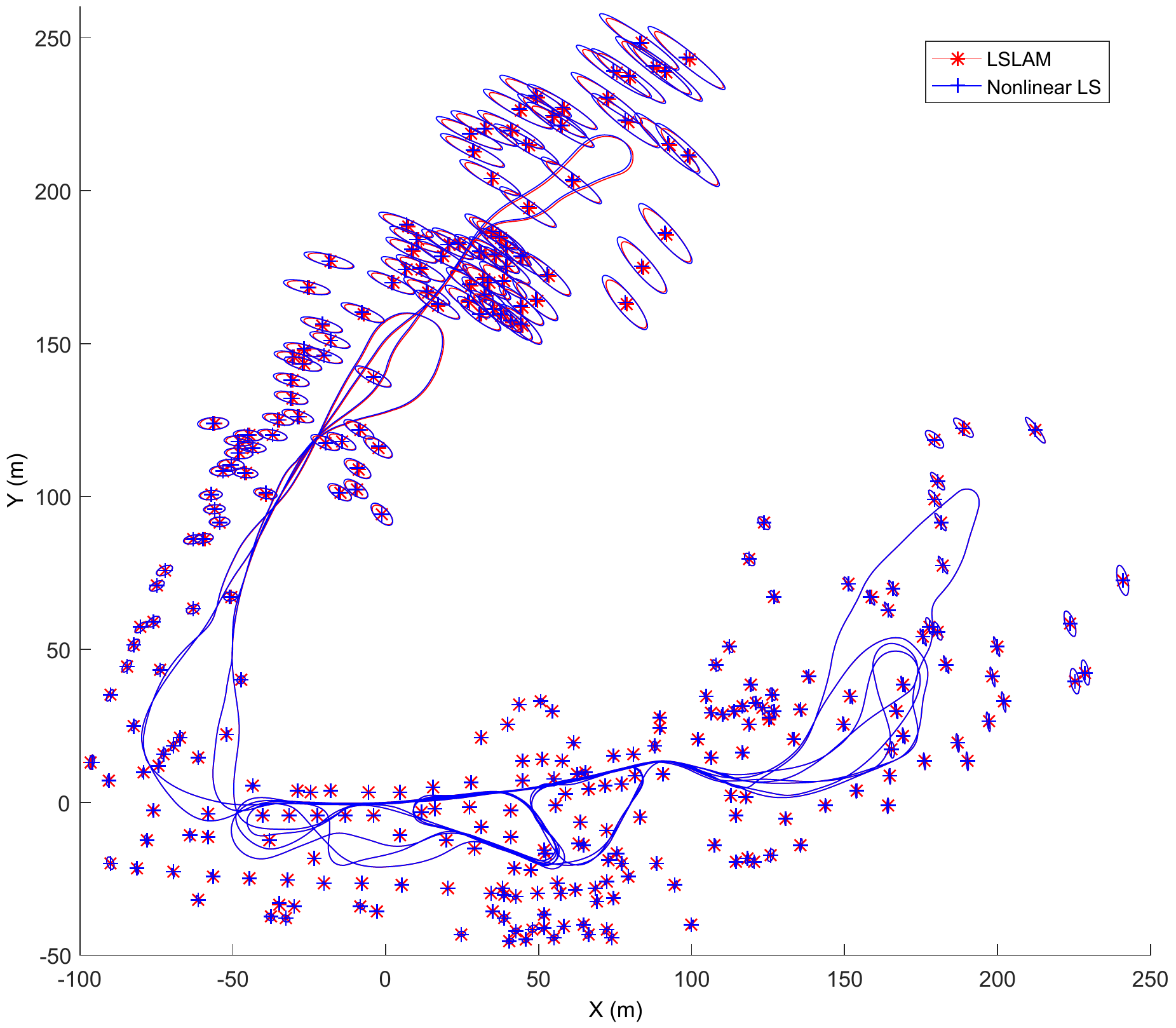}}
\centering \subfigure[Local view] {\label{fig:Un_VP2}
\includegraphics[width=0.3\textwidth,height=0.2\textheight]{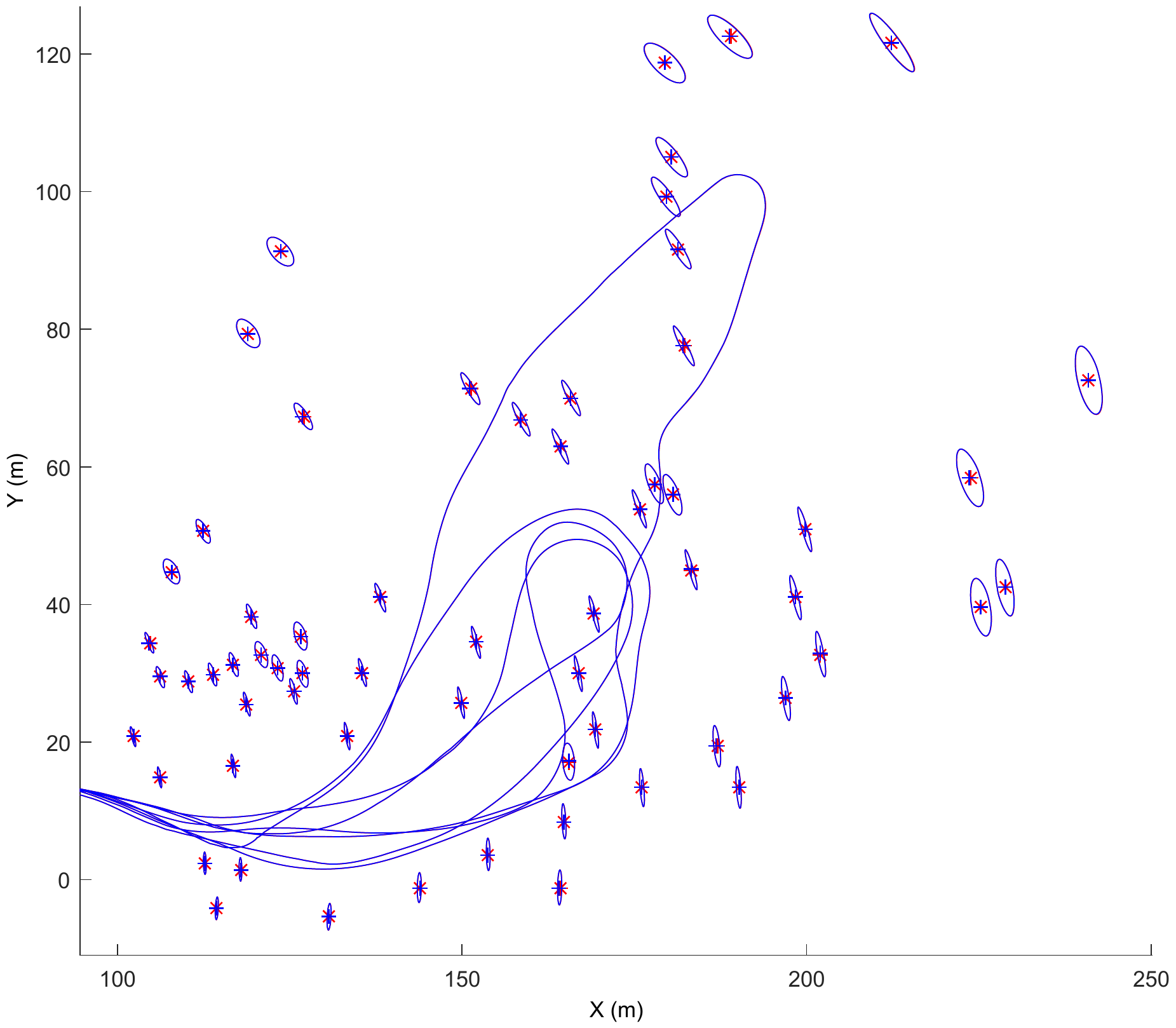}}
\centering \subfigure[Pose uncertainty] {\label{fig:Un_VP3}
\includegraphics[width=0.36\textwidth,height=0.2\textheight]{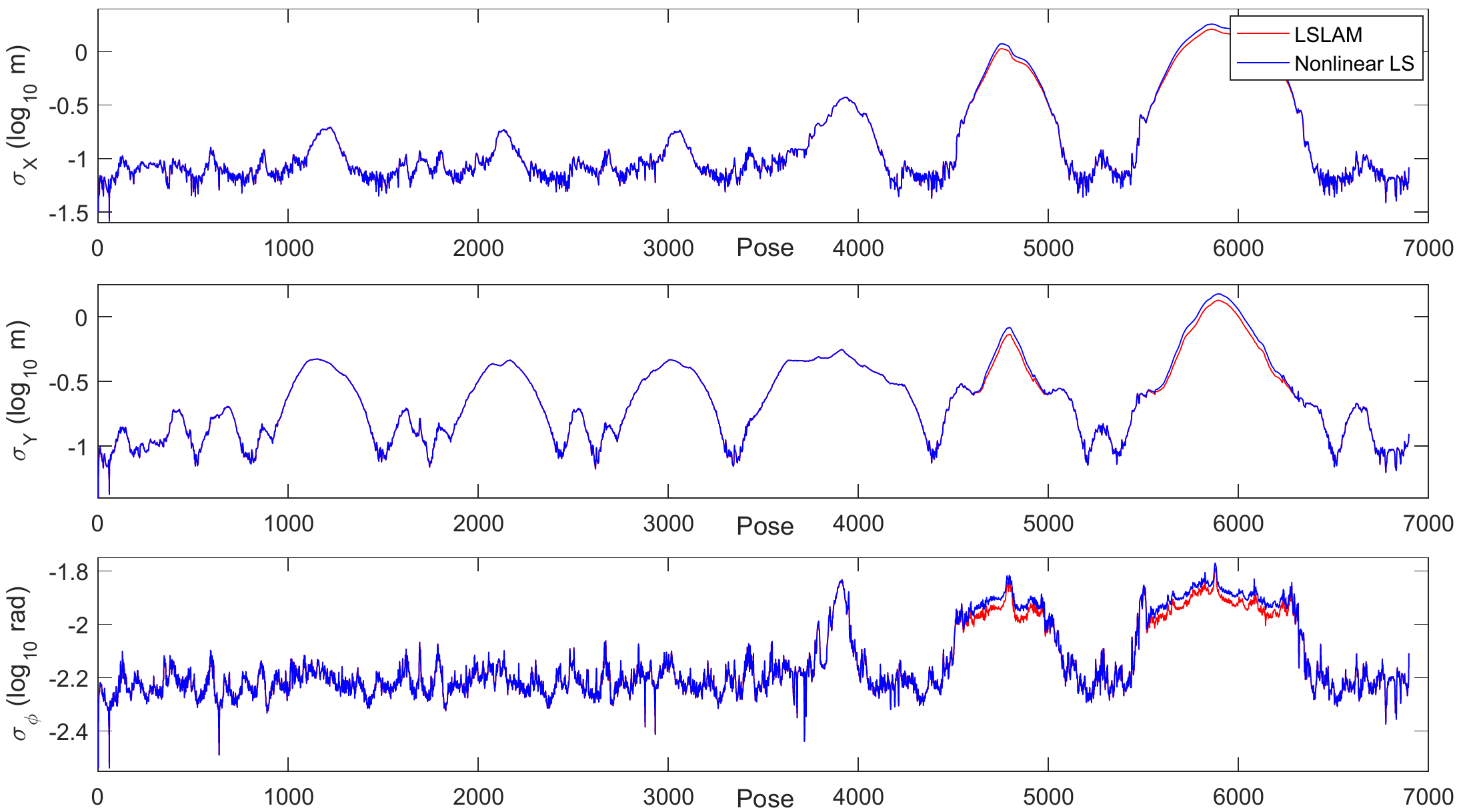}}
 \caption{Uncertainty estimations of both features (a and b) and poses (c) by Linear SLAM and Nonlinear LS for Victoria Park dataset. For better visualisation, the uncertainty shows in (a) and (b) is scaled by 2.} \label{fig:Un_VP}
\end{figure*}

\begin{figure*}
\centering \subfigure[Feature uncertainty] {\label{fig:Un_DLR1}
\includegraphics[width=0.3\textwidth,height=0.2\textheight]{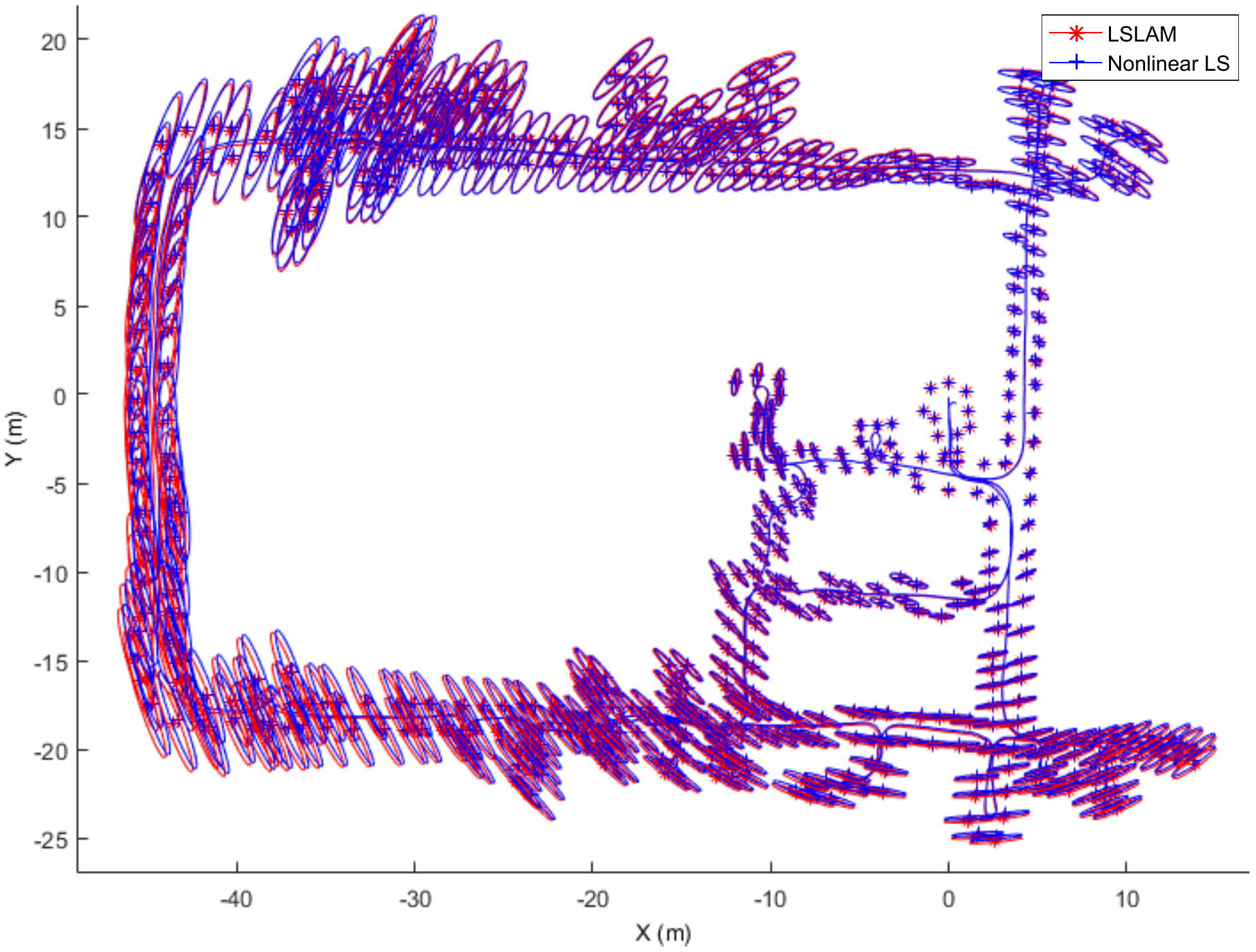}}
\centering \subfigure[Local view] {\label{fig:Un_DLR2}
\includegraphics[width=0.3\textwidth,height=0.2\textheight]{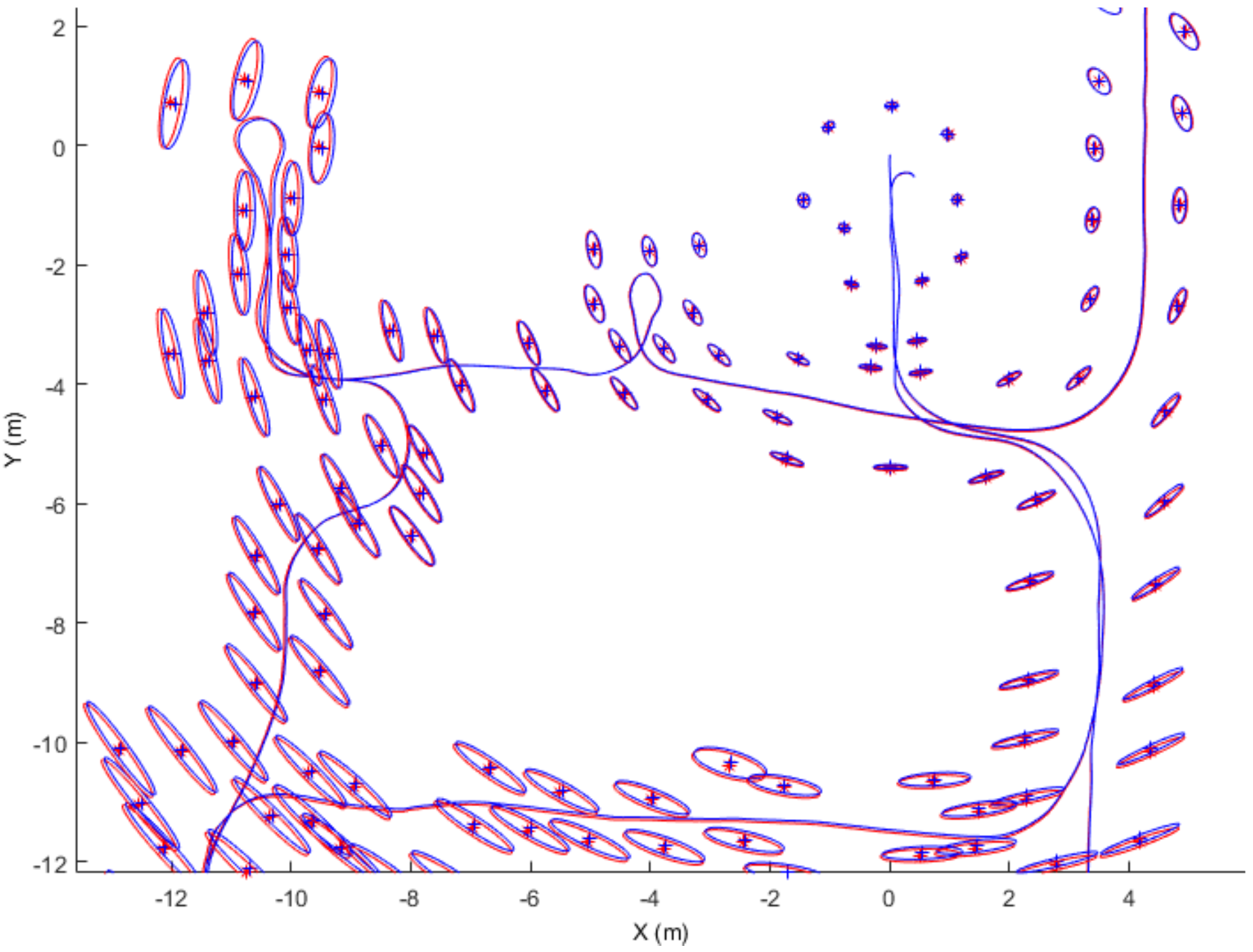}}
\centering \subfigure[Pose uncertainty] {\label{fig:Un_DLR3}
\includegraphics[width=0.36\textwidth,height=0.2\textheight]{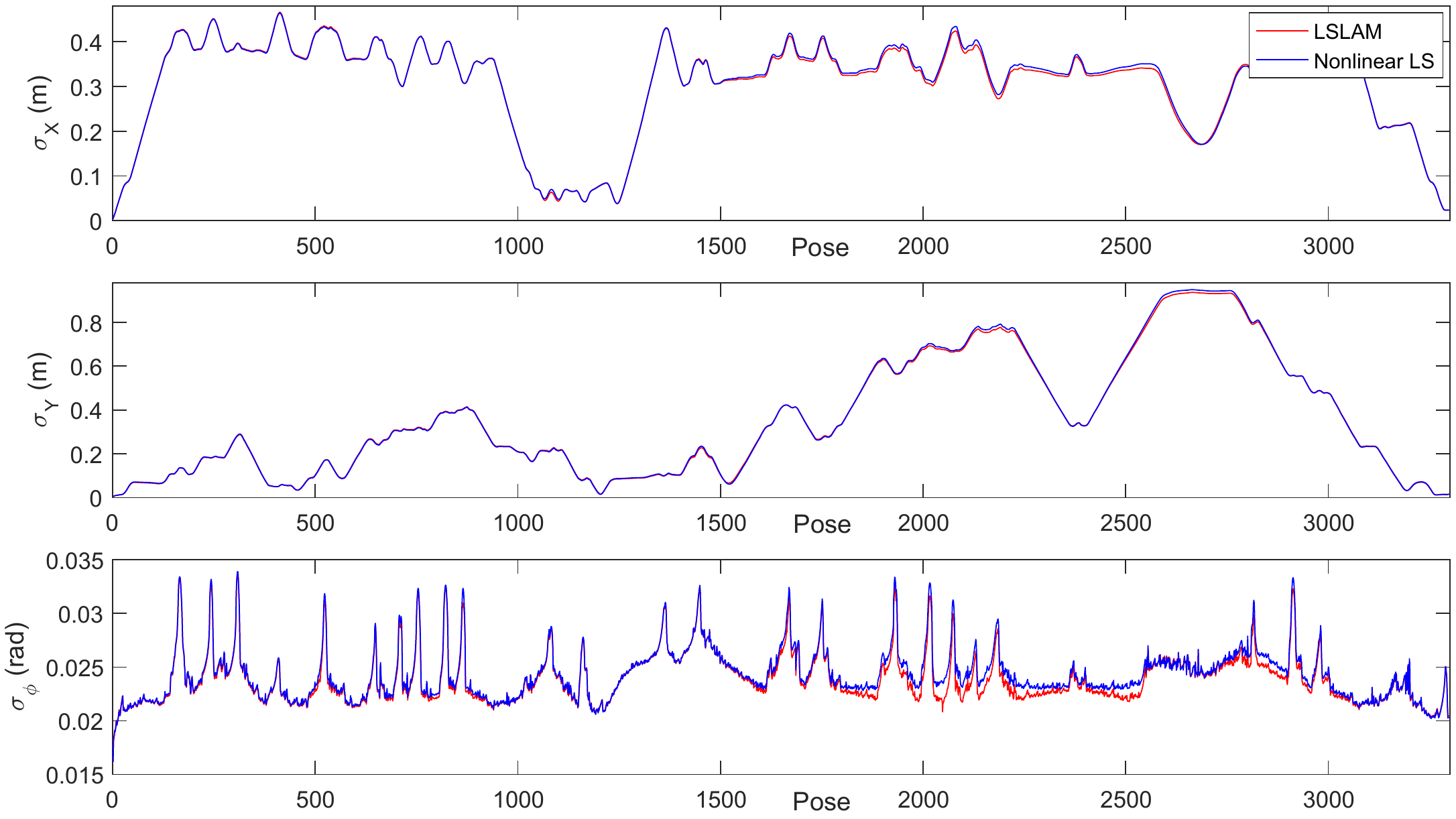}}
 \caption{Uncertainty estimations of both features (a and b) and poses (c) by Linear SLAM and Nonlinear LS for DLR dataset.} \label{fig:Un_DLR}
\end{figure*}

The uncertainties obtained from the proposed Linear SLAM algorithm are also compared to the uncertainties obtained by performing the Nonlinear LS algorithm. Both VicPark6898 and DLR3298 datasets are employed to this comparative evaluation. The uncertainties of the features estimated from Linear SLAM and Nonlinear LS are shown in Fig. \ref{fig:Un_VP}(a)(b) and Fig. \ref{fig:Un_DLR}(a)(b) in red and blue, respectively. And the uncertainties of the poses estimated from both two algorithms are shown in Fig. \ref{fig:Un_VP3} and Fig. \ref{fig:Un_DLR3}. As we can see from the figures, the uncertainties estimated from the proposed Linear SLAM algorithm is very close to those obtained from the Nonlinear LS algorithm, which indicates that the proposed Linear SLAM algorithm can obtain accurate results in term of both estimates and uncertainty.

\subsubsection{3D Pose-feature Map Joining}

For the 3D pose-feature map joining, a simulation is done with a trajectory of 871 poses and uniformly distributed features in the environment. The pose-feature map joining using Linear SLAM is applied with 870 local maps and the result is shown in Fig. \ref{fig:3D Simu870}. The RMSE of the position of poses and features by Linear SLAM as compared with the ground truth are 0.003266m and 0.003158m, respectively, with the computational cost of 1.00s.

\begin{figure*}
\centering \subfigure[Intel] {\label{fig:Intel}
\includegraphics[width=0.3\textwidth,height=0.2\textheight]{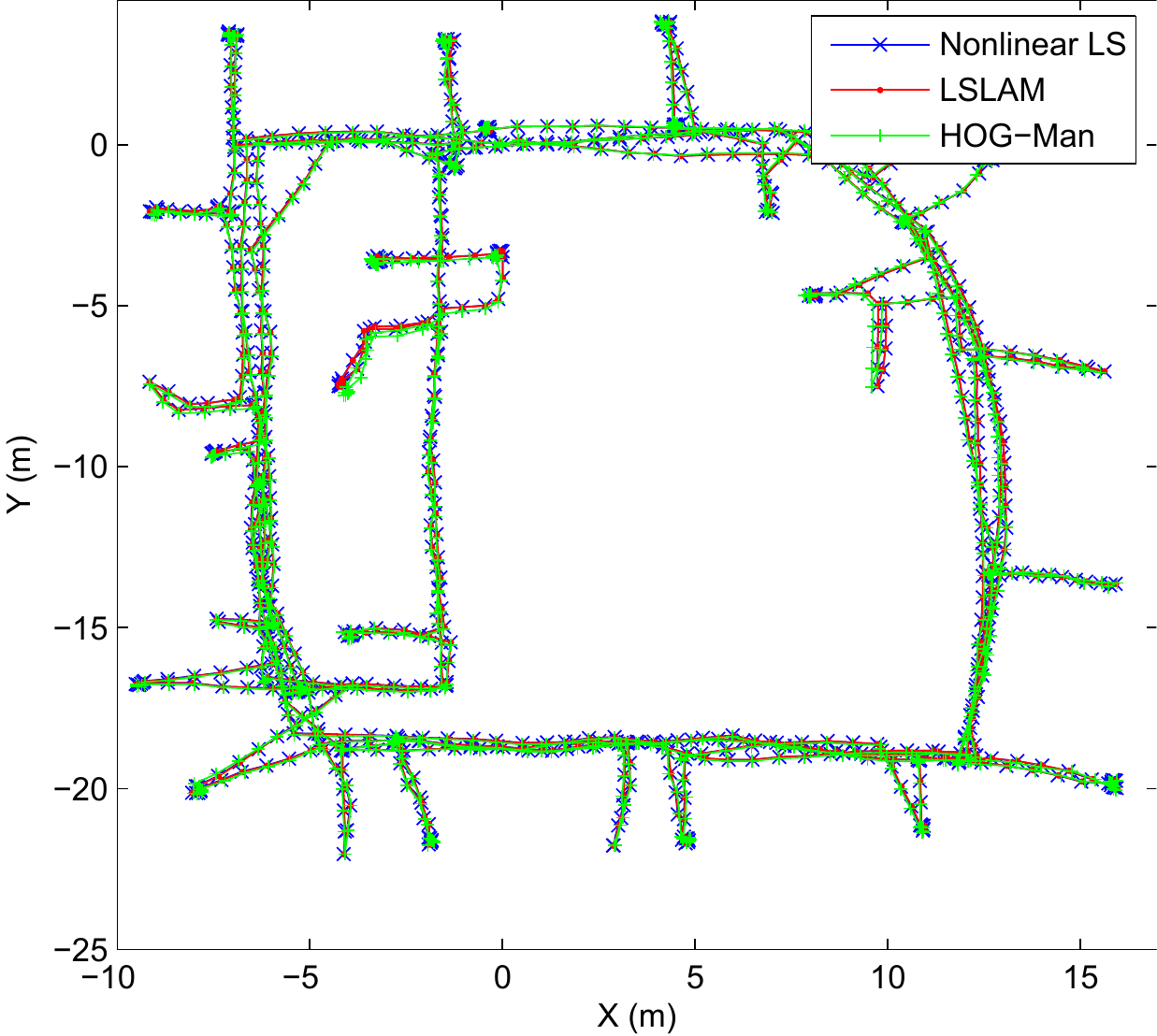}}
\centering \subfigure[Manhattan] {\label{fig:Manhattan}
\includegraphics[width=0.3\textwidth,height=0.2\textheight]{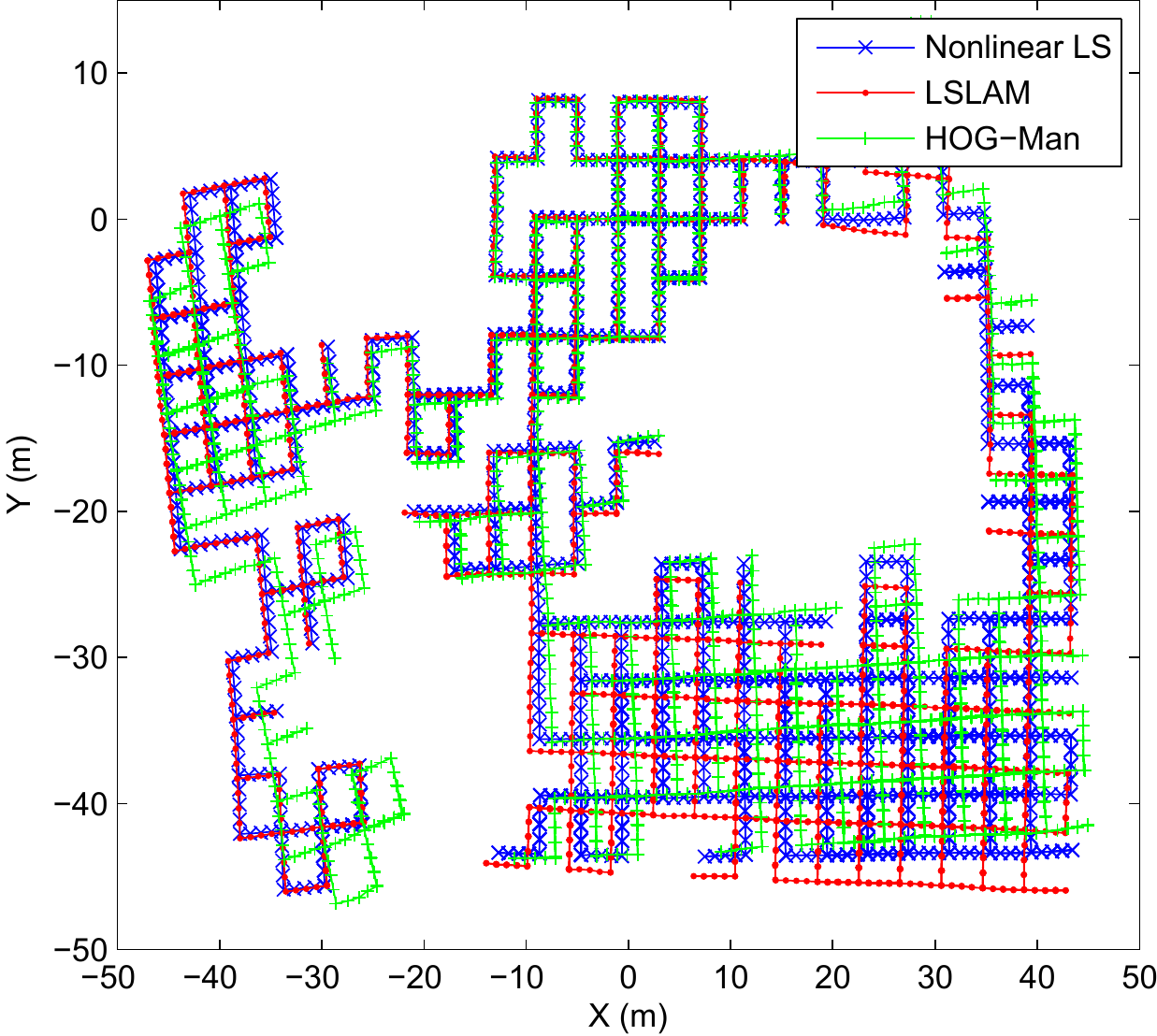}}
\centering \subfigure[City10000] {\label{fig:City10000}
\includegraphics[width=0.3\textwidth,height=0.2\textheight]{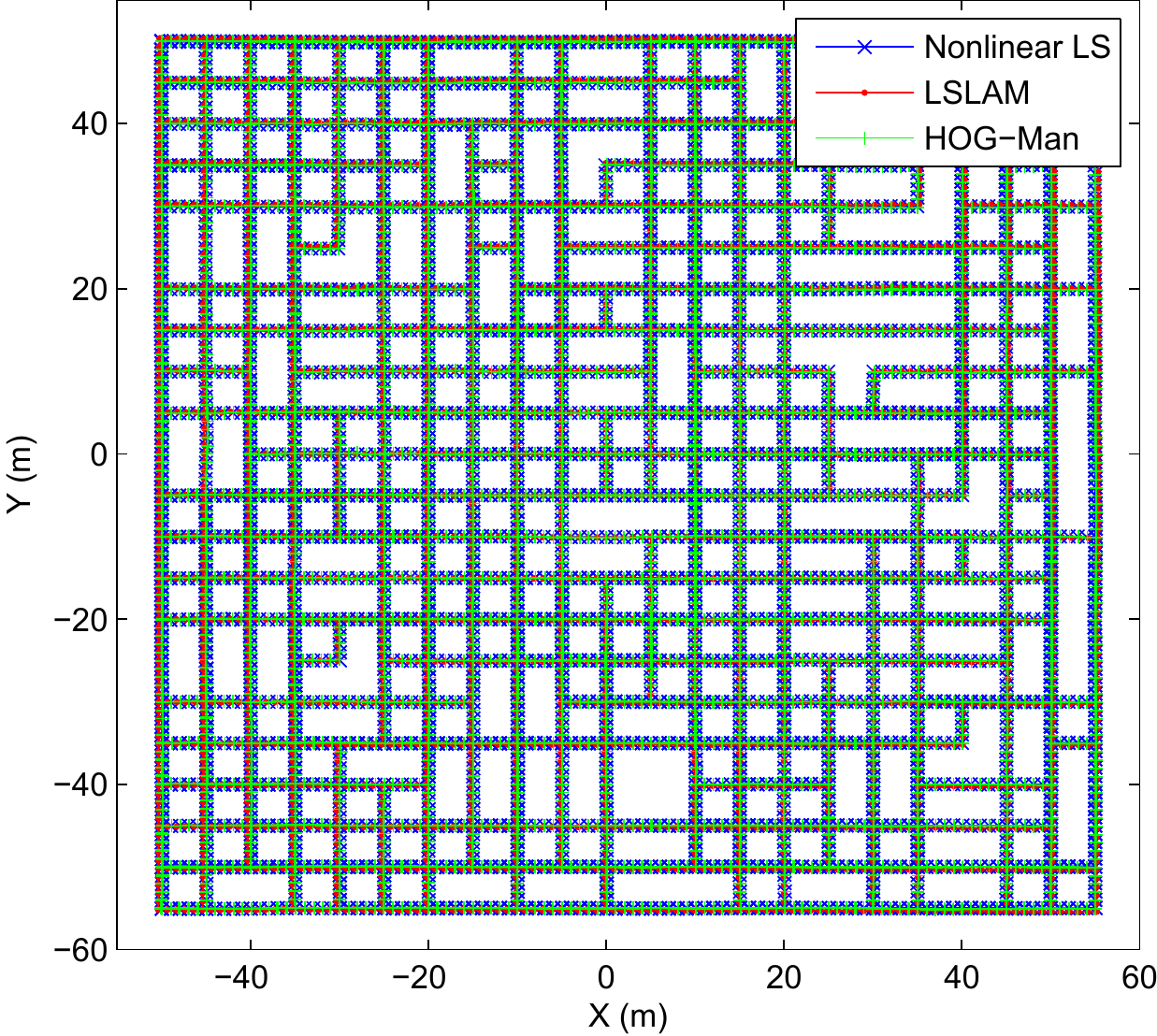}}
\centering \subfigure[Sphere] {\label{fig:Sphere}
\includegraphics[width=0.3\textwidth,height=0.2\textheight]{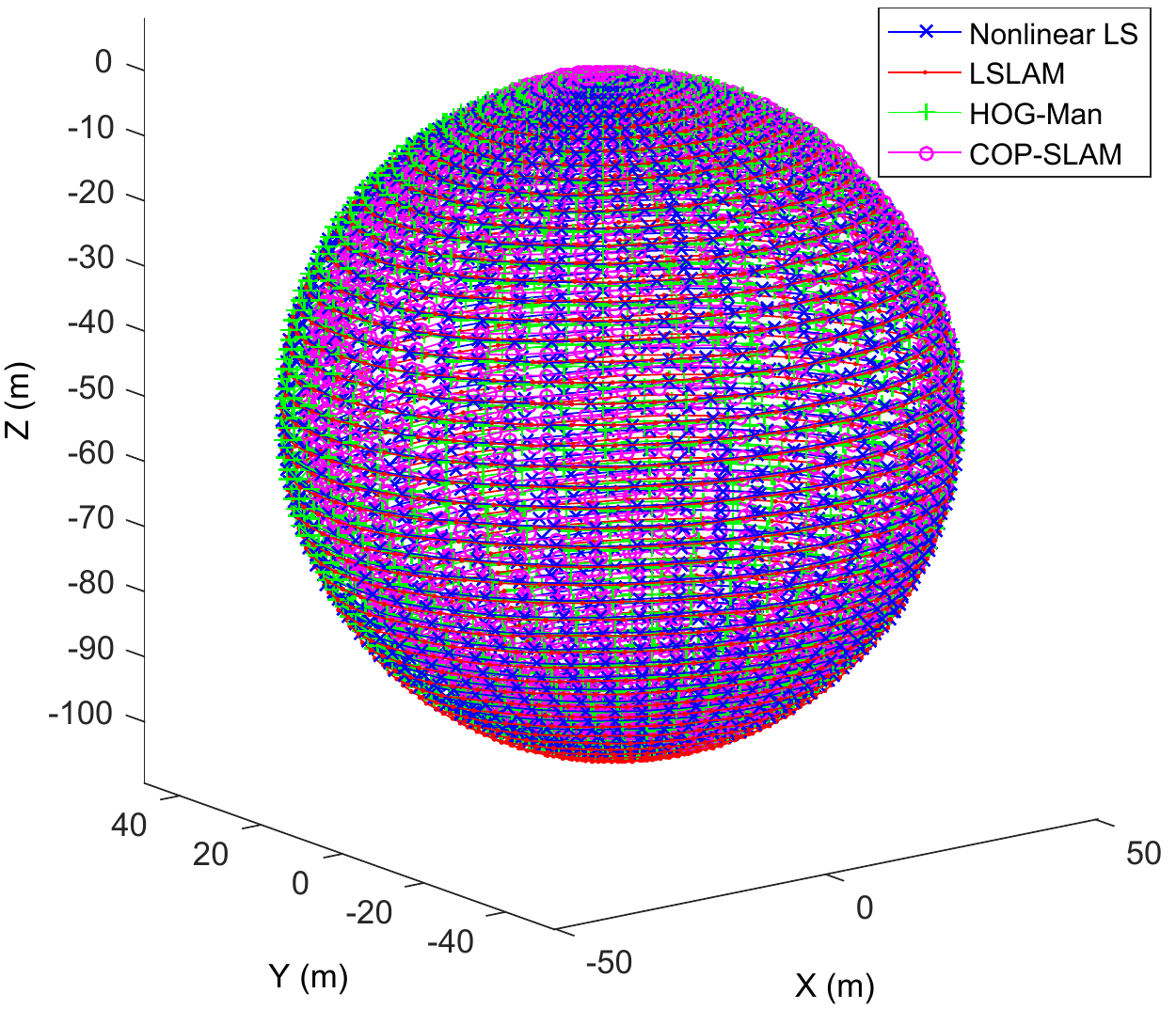}}
\centering \subfigure[Parking garage] {\label{fig:Parking garage}
\includegraphics[width=0.3\textwidth,height=0.2\textheight]{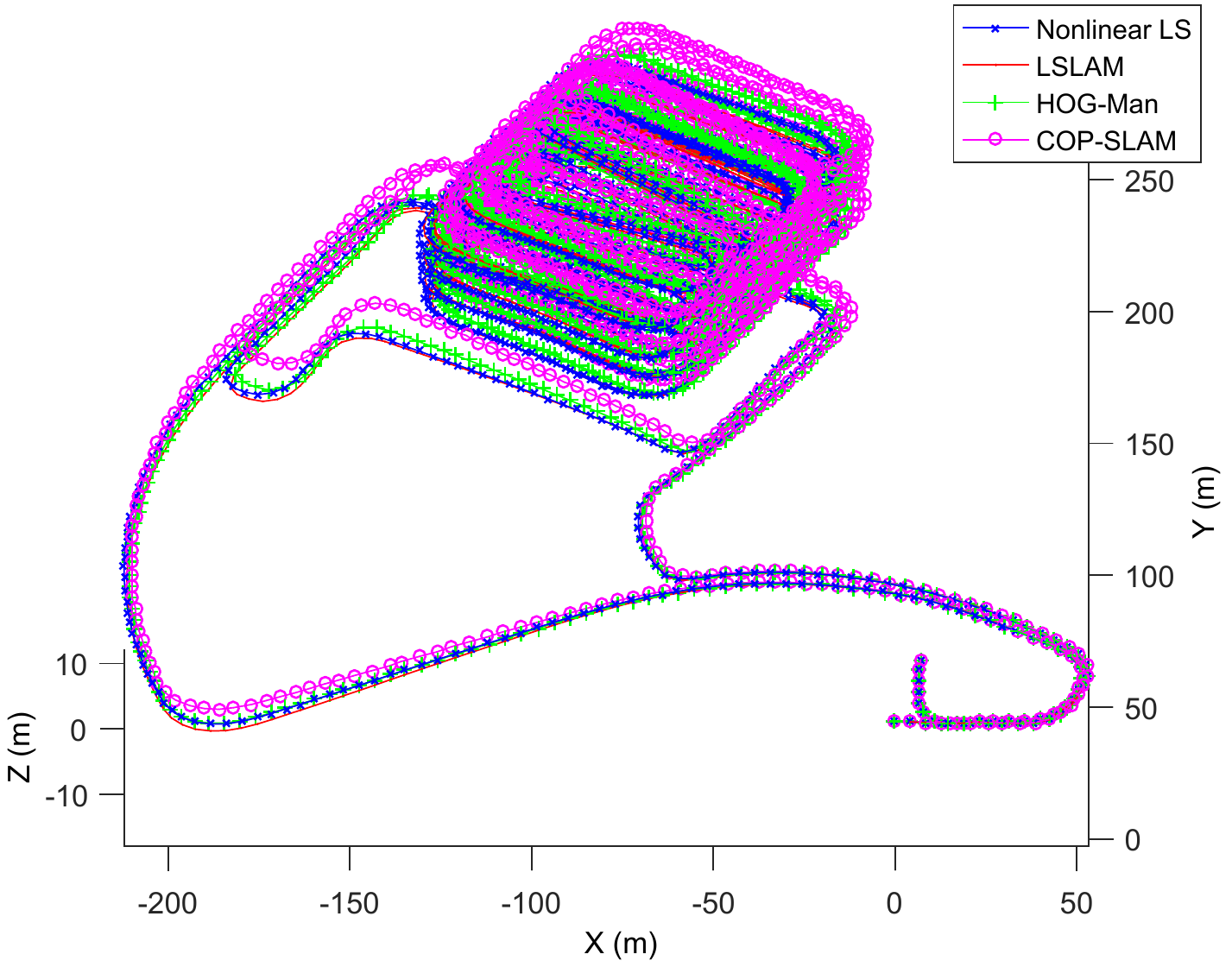}}
 \caption{2D and 3D pose-only map joining results of Linear SLAM (LSLAM). Fig. \ref{fig:Intel} to Fig. \ref{fig:City10000} are the results of Intel, Manhattan and City10000 2D pose graph datasets. Fig. \ref{fig:Sphere} and Fig. \ref{fig:Parking garage} are the results of Sphere and Parking garage 3D pose graph datasets. All results are compared with those of HOG-Man and nonlinear least squares. The COP-SLAM is also compared using 3D Sphere and Parking garage datasets. The quantitive comparison is given in Table \ref{Tab:RMSEPoseGraph}.} \label{fig:PoseGraph}
\end{figure*}

\subsection{Pose-only Map Joining}

For 2D pose-only map joining, three 2D pose graph SLAM datasets, including one experimental dataset Intel \cite{Hahnel-Intel} and two simulation datasets Manhattan \cite{Olson} and City10000 \cite{Kaess12ijrr}, are used. For 3D pose-only map joining, two 3D pose graph SLAM datasets, including one simulation dataset Sphere \cite{iSAM} and one experimental dataset Parking Garage \cite{G2O}, are used.

\begin{table*}
\tabcolsep 0pt \caption{The Absolute and Relative RMSE* and $\chi^2$ of Different Pose Graph SLAM Algorithms}
\label{Tab:RMSEPoseGraph}
\begin{center}
\begin{tabular*}{1.00\textwidth}{@{\extracolsep{\fill}} lcccccccccc}
\hline\hline
\multicolumn{1}{c}{} &\multicolumn{3}{c}{HOG-Man} &\multicolumn{3}{c}{COP-SLAM} &\multicolumn{3}{c}{Linear SLAM} & \multicolumn{1}{c}{Nonlinear LS}\quad\\
\quad Dataset & RMSE(Abs) & RMSE(Rel) & $\chi^2$ & RMSE(Abs) & RMSE(Rel) & $\chi^2$ & RMSE(Abs) & RMSE(Rel) & $\chi^2$ & $\chi^2$ \quad\\
\hline
\quad Intel & 0.054648 & 0.017277 & 972.00 &   &   &   & \textbf{0.006571} & \textbf{0.000216} & \textbf{546.51} & 546.46 \quad\\
\quad Manhattan & 1.180292 & 0.028890 & 548.01 &   &   &   & \textbf{1.114862} & \textbf{0.011576} & \textbf{214.12} & 137.91 \quad\\
\quad City10000 & \textbf{0.062695} & 0.019269 & 1276.5 &   &   &   & 0.191676 & \textbf{0.004678} & \textbf{601.38} & 511.99 \quad\\
\quad Sphere & 1.854379 & 0.090491 & 2051.6 & 2.877556 & 0.236187 & 5532.6 & \textbf{1.303615} & \textbf{0.050658} & \textbf{1363.8} & 1154.5 \quad\\
\quad ParGarage & 2.315629 & 0.008350 & 2.1998 & 4.847823 & 0.273661 & 1268.9 & \textbf{1.603590} & \textbf{0.004693} & \textbf{1.5978} & 1.2888 \quad\\
\hline\hline
\end{tabular*}
\end{center}
* Both absolute RMSE (RMSE(Abs)) and relative RMSE (RMSE(Rel)) (in meters) of the pose positions are w.r.t. the results of the Nonlinear LS.
\end{table*}

The results of 2D pose-only map joining using Linear SLAM are shown in Fig. \ref{fig:Intel} to Fig. \ref{fig:City10000} while the results of 3D pose-only map joining using Linear SLAM are shown in Fig. \ref{fig:Sphere} to Fig. \ref{fig:Parking garage}. For these five datasets, each local map is directly obtained using the relative poses with respect to one pose, thus there is no nonlinear optimization process for local map building involved in the Linear SLAM algorithm. For comparison, HOG-Man \cite{HOG-Man} and Nonlinear LS are applied to get the SLAM results for these five pose graph datasets. COP-SLAM \cite{COP-SLAM} is also performed to the 3D Sphere and Parking Garage datasets. The results are also shown in Fig. \ref{fig:PoseGraph}. The RMSE of both the absolute trajectory error \cite{Endres-ICRA2012} and the relative trajectory error \cite{Kummerle-AR09}, as well as the $\chi^2$ of Linear SLAM, HOG-Man, and COP-SLAM as compared with the Nonlinear LS SLAM results are shown in Table \ref{Tab:RMSEPoseGraph}. It can be seen that, almost all the RMSE and the $\chi^2$ of the results from Linear SLAM are smaller than those from HOG-Man and COP-SLAM (only the absolute RMSE of Linear SLAM using City10000 dataset is slightly larger than that of HOG-Man).

The computational cost of Linear SLAM for these five datasets are 0.22s, 0.85s, 5.01s, 2.52s and 1.50s, respectively. For the pose graph SLAM datasets, the computational cost of Linear SLAM is about 20 times cheaper than that of HOG-Man \cite{HOG-Man}. Both of the two algorithms solve the SLAM problem in a hierarchical manner. Our implementation of Linear SLAM is slower than g2o \cite{G2O} mainly due to the way our code is structured (g2o code is highly optimized and is one of the fastest implementations of Nonlinear LS SLAM). In the experiments, g2o is initialized by using the results from Linear SLAM. While, Linear SLAM does not require any initial value. The COP-SLAM \cite{COP-SLAM} is faster than Linear SLAM when performing the two 3D datasets, while the accuracy is the worst comparing to Linear SLAM and HOG-Man.

\subsection{Feature-only Map Joining}

For 2D feature-only map joining, the VicPark200 and DLR200 datasets are used. Here the local map datasets are the same as the ones used in the experiment of 2D pose-feature map joining in Section
\ref{Sec:PoseFeatureResult}. Similar to D-SLAM map joining (DMJ) \cite{Huang:DSLAM-AR}, a preprocessing is done to make sure there are at least 2 (for 2D) or 3 (not collinear, for 3D) common features between every two contiguous local maps such that the local maps can be combined together (If there are not enough common features between the two local maps, then the Linear SLAM algorithm for pose-feature maps is used to join them together to form a larger local map). Then the local maps are transformed into the coordinates defined by features and all the poses are marginalized out before the map joining.

\begin{figure*}
\centering \subfigure[VicPark200] {\label{fig:VicPark200_DSLAM}
\includegraphics[width=0.3\textwidth]{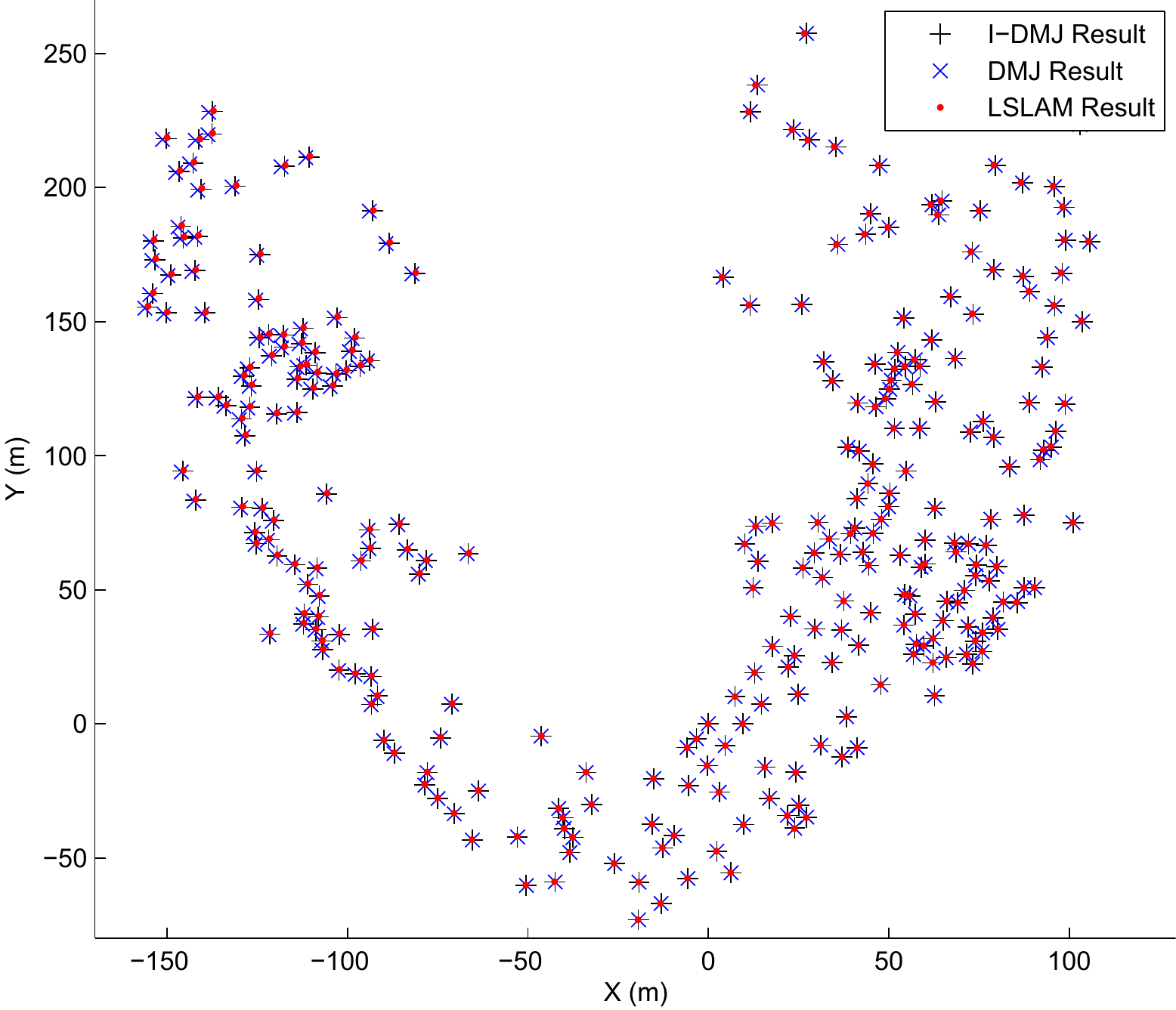}}
\centering \subfigure[DLR200] {\label{fig:DLR200_DSLAM}
\includegraphics[width=0.3\textwidth]{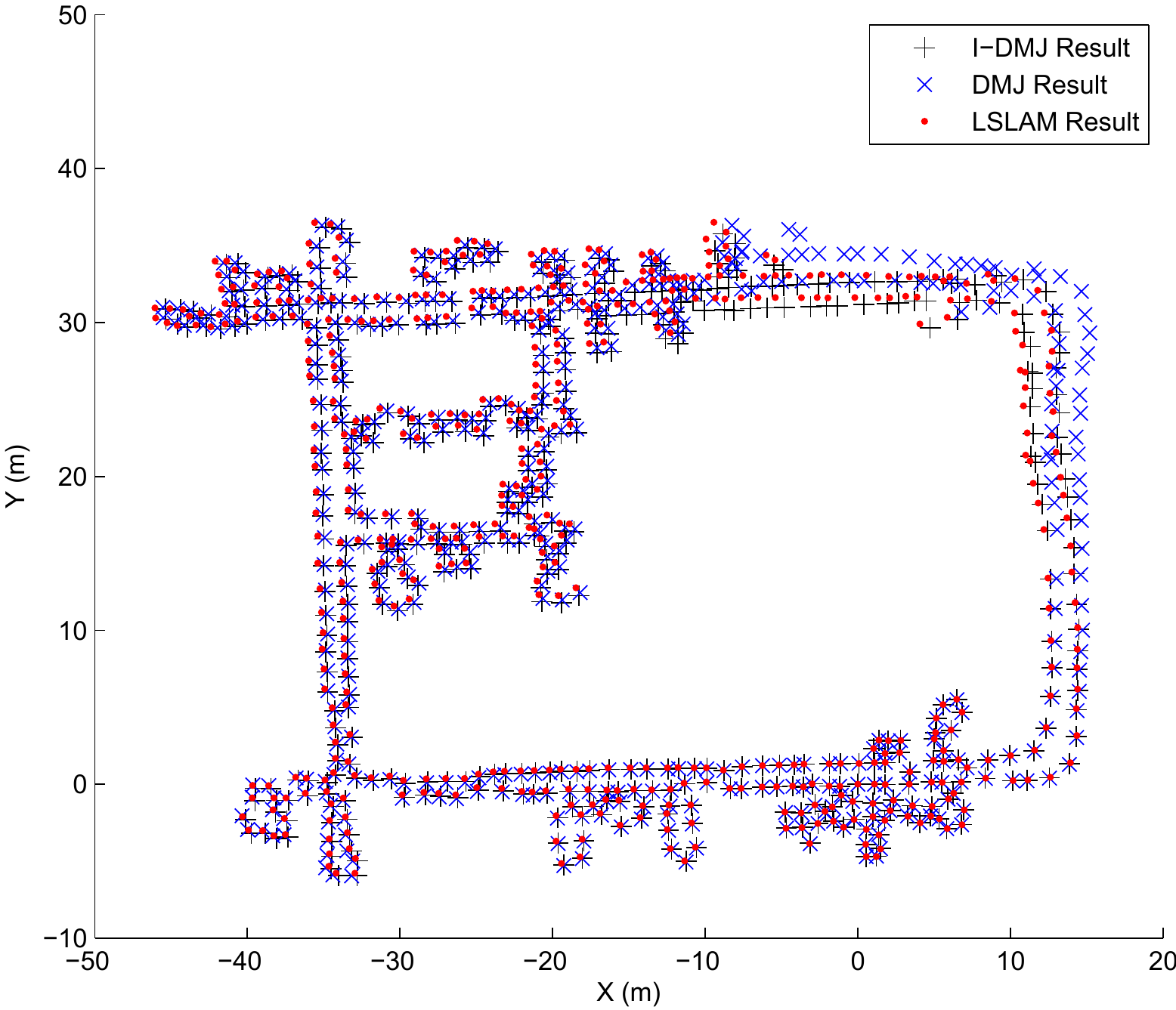}}
\centering \subfigure[3D Simu871] {\label{fig:3D Simu870_DSLAM}
\includegraphics[width=0.3\textwidth]{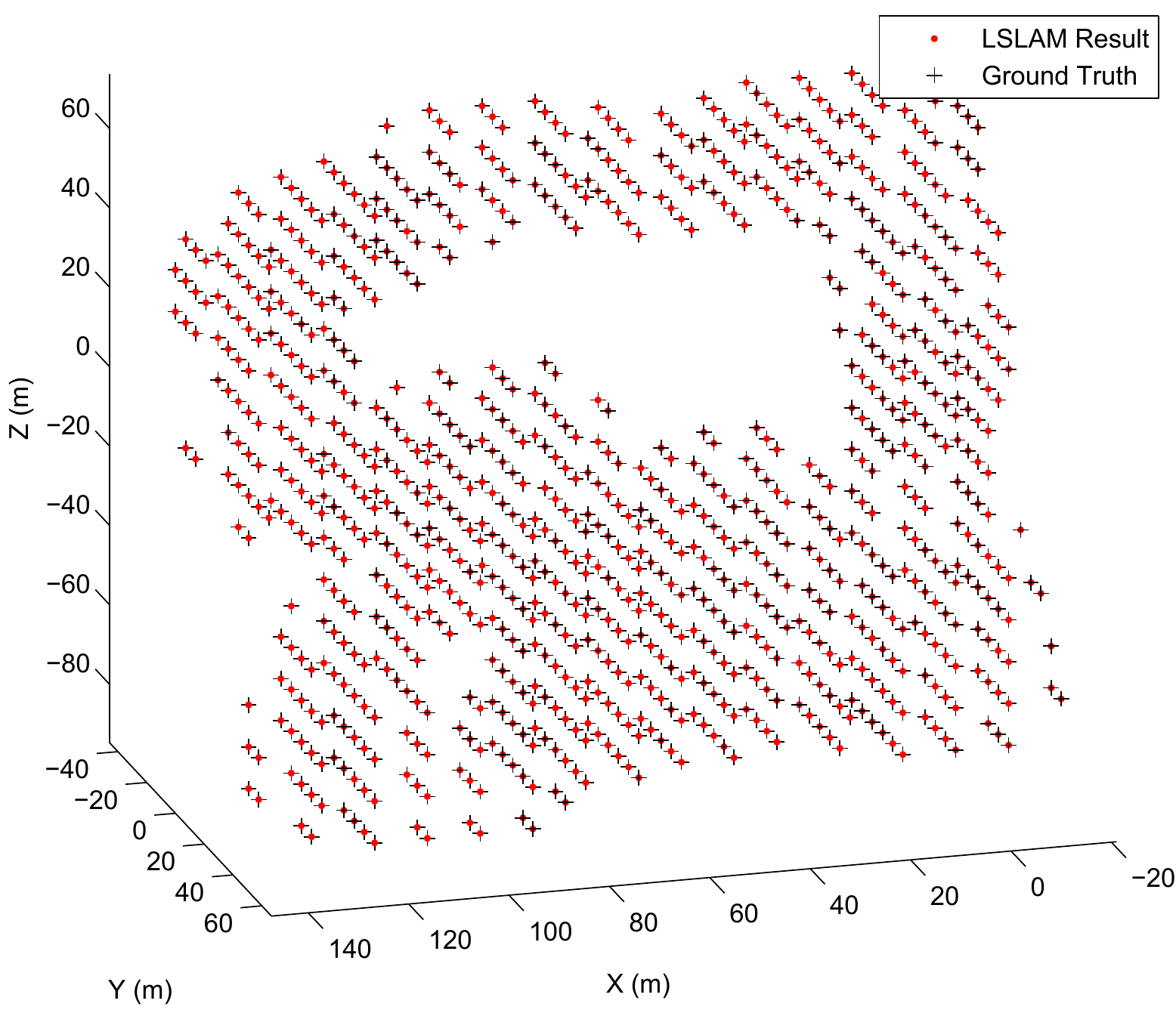}}
 \caption{2D and 3D feature-only map joining results using Linear SLAM (LSLAM). Fig.
 \ref{fig:VicPark200_DSLAM} is the result of 2D Victoria Park
 datasets in the coordinate frame defined by Feature 84 and 85. Fig.
 \ref{fig:DLR200_DSLAM} is the result of 2D DLR datasets in the coordinate frame defined by Feature 367 and 369. Both the 2D results are compared with DMJ and I-DMJ \cite{Huang:DSLAM-AR}. Fig. \ref{fig:3D Simu870_DSLAM} is the result of the 3D simulation with 870 local maps in the coordinate frame defined by Feature 1758, 1757 and 1359.} \label{fig:FeatureOnly}
\end{figure*}

The results of 2D feature-only map joining using Linear SLAM algorithm for the two datasets are
shown in Fig. \ref{fig:VicPark200_DSLAM} and Fig. \ref{fig:DLR200_DSLAM}. For comparison, the results from DMJ and Iterated DMJ (I-DMJ) \cite{Huang:DSLAM-AR} are also shown in the figures.
Similar to I-SLSJF, the result of I-DMJ is equivalent to that of
performing nonlinear least squares optimization using all the
feature-only local maps as observations \cite{Huang:DSLAM-AR}. As
shown in Fig. \ref{fig:VicPark200_DSLAM} and Fig.
\ref{fig:DLR200_DSLAM}, the results of Linear SLAM are better than
those of DMJ, while close to the optimal I-DMJ results.

The RMSE of Linear SLAM and DMJ as compared with the optimal I-DMJ results are shown in Table \ref{Tab:RMSEDSLAM}. The computational cost of Linear SLAM for VicPark200 and DLR200 datasets are 1.14s, 0.58s respectively, which is significantly cheaper than that of DMJ (I-DMJ) \cite{Huang:DSLAM-AR}.

For the 3D feature-only SLAM, the simulated 3D Simu871 dataset is used, and
the result of Linear SLAM algorithm as well as the ground truth are shown in Fig. \ref{fig:3D Simu870_DSLAM}. The RMSE of Linear SLAM as compared with the ground truth is shown in Table \ref{Tab:RMSEDSLAM}. The computational cost of Linear SLAM is 2.61s.

In summary, simulation and experimental results using publicly available datasets demonstrated that the proposed Linear SLAM is consistent and efficient and can produce results very close to the full nonlinear least squares based methods. For most of the datasets, its accuracy is better than other map joining based SLAM algorithms and some efficient SLAM algorithms such as  COP-SLAM and HOG-Man.

\begin{table}
\tabcolsep 0pt
\caption{RMSE* of Feature Locations for Different Feature-Only Map Joining Algorithms}
\label{Tab:RMSEDSLAM}
\begin{center}
\begin{tabular*}{0.50\textwidth}{@{\extracolsep{\fill}} lccc}
\hline\hline
\multicolumn{1}{c}{} &\multicolumn{1}{c}{DMJ} &\multicolumn{1}{c}{Linear SLAM} & \multicolumn{1}{c}{I-DMJ}\quad\\
\quad Dataset & Feature & Feature & Feature \quad\\
\hline
\quad VicPark200 & 0.268626 & \textbf{0.193141} & 0 \quad\\
\quad DLR200 & 0.536729 & \textbf{0.393460} & 0 \quad\\
\quad 3D Simu871 & N/A & 0.0014850 & N/A \quad\\
\hline\hline
\end{tabular*}
\end{center}
* For joining 2D feature-only maps, the result of I-DMJ is used as the benchmark. For the 3D Simu871 dataset used in feature-only map joining, the ground truth is the benchmark.
\end{table}

\section{Discussions}\label{sec:discussions}

The Linear SLAM algorithm proposed in this paper is based on map joining. Map joining has already been shown to be able to improve the efficiency for large-scale SLAM as well as reduce the linearization errors. Similar to recursive nonlinear estimation (e.g. \cite{iSAM}), map joining also solves a recursive version of the problem step by step, thus does not get stuck in local minima very often. Apart from the advantages of existing map joining based algorithms, the key advantage of Linear SLAM is that it only requires solving linear least squares instead of nonlinear least squares. Thus there is no need to compute initial value and no need to perform iterations. The key to make the map joining problem linear is that the two maps to be fused are always built in the same coordinate frame. Basically, when joining two local maps built at different coordinate frames, the estimation/optimization problem is nonlinear, which is the case in most of the existing map joining algorithms. But when joining two local maps with the same coordinate frame, the problem can be linear, which is the idea used in this paper. The sequential map joining \cite{Tardos-map-joining,Williams-PhD-thesis,Shoudong:SLSJF-TRO} and Divide and Conquer map joining \cite{Divide-and-Conquer:-EKF-SLAM}) join two maps at a time, thus both strategies can be applied in Linear SLAM.

Similar to SLSJF \cite{Shoudong:SLSJF-TRO}, CF-SLAM \cite{Combined-filter} and other optimization based submap joining approaches \cite{Huang:DSLAM-AR,HOG-Man}, the sparseness of the information matrix is maintained in the proposed linear map joining algorithm as the coefficient matrix $A$ in (\ref{Eq::A_Pose_Feature}) is always sparse. Also similar to SLSJF and CF-SLAM, there is no information loss or information reuse in the Linear SLAM algorithm. The simulation and experimental results demonstrated that Linear SLAM results are better than that of CF-SLAM and SLSJF in most of the cases. The reason is, when joining two maps, CF-SLAM and SLSJF use EIF while Linear SLAM performs the optimal linear least squares. The Linear SLAM result is not as good as that of I-SLSJF because an additional smoothing step through nonlinear optimization is applied in I-SLSJF. It is clear that fusing local maps built in different coordinate frames optimally using linear approach is impossible. Thus if we want to apply smoothing after joining all the local maps together by Linear SLAM, we have to use nonlinear optimization.

The difference between the result of Linear SLAM and the best solution to the full nonlinear least squares SLAM comes from two reasons: (i) Instead of using the original odometry and observation information, we summarize the local map information as the local map estimate together with its uncertainty (information matrix) and use this information in the map joining. This is the case for many map joining algorithms including CF-SLAM, SLSJF and I-SLSJF. (ii) Instead of fusing all the local maps together in one go using nonlinear optimization (as in I-SLSJF), we fuse two maps at a time (as in CF-SLAM and SLSJF), resulting in a suboptimal solution. It should be pointed out that the difference is due to the linearization performed (Linear SLAM, CF-SLAM, SLSJF, I-SLSJF are all equivalent to full least squares SLAM when the odometry and observation functions are linear).

In general, the more accurate the local maps are, the closer the Linear SLAM result to the best solution to the full nonlinear least squares SLAM. In some of the experimental results shown in Section \ref{sec:results} (e.g. all the Linear SLAM results using the pose graph SLAM datasets), one local map was directly obtained using the information related to one pose. In this case, Linear SLAM completely avoids the nonlinear optimization in the local map building process. However, its performance may be sacrificed because the quality of the local maps is not well controlled. To improve the performance of Linear SLAM, we recommend to use nonlinear optimization techniques to build high quality small local maps (accurate estimate and small covariance matrix) taking into account the connectivity of the local map graph \cite{Olson-Kaess-RSS-workshop}, and then apply our linear map joining algorithm. Furthermore, the solution from Linear SLAM (using least squares to build local maps without marginalization) can be used as an (excellent) initial guess for the iterative approaches to solve the full nonlinear least squares SLAM, similar to T-SAM2 \cite{TSAM2} or \cite{Grisetti-IROS2012}. However, this will add more nonlinear components in the algorithm. The tradeoff between improving the accuracy of the result and reducing the nonlinear components in  Linear SLAM algorithm requires further investigations.

It should also be pointed out that the three local maps in Fig. \ref{fig:localmap1}, Fig. \ref{fig:localmap2} and Fig. \ref{fig:localmap3} are all optimal for the given local map data; they are equivalent and can be transferred from one to another using coordinate transformations. However, when using these different versions of local maps in map joining, the final results can be slightly different due to the different linearization performed. For the actual nonlinear odometry and observation models, using which version of local maps gives the best solution depends on many factors (such as the size and number of local maps, the number of features observed at each step, the noise level of odometry and observation, etc.) and requires further investigations.


We would like to point to literature \cite{huangICRA2012,WangRSS2012} indicating that SLAM problem is close to linear and its dimensionality can be substantially reduced as possible reasons why the linear approximation proposed in this paper produces solutions that are very close to those obtained using optimal nonlinear least squares SLAM. Linear SLAM is a nice way to separate the linear components from the nonlinear components in the SLAM problem through submap joining.

Unlike traditional local submap joining, the final global map obtained from this Linear SLAM algorithm is not in the coordinate frame of the first local map. For Divide and Conquer map joining, the coordinate frame of the final global map is the same as the coordinate frame of the last two maps that were fused together. For sequential map joining, the coordinate frame of the final global map is the last robot pose in the last local map. This is somewhat similar to the robocentric mapping \cite{Robocentric-mapping} where the map is transformed into the coordinate frame of the current robot pose in order to reduce the linearization error in the EKF framework. As discussed in Section \ref{sec:coordinate-frame}, if a final global map in the coordinate frame of the first robot pose is required, e.g. in order to compare the result with traditional local submap joining, only a coordinate transformation described in Section \ref{testShoudong} is needed. The global map after the coordinate transformation is equivalent to the global map obtained from the Linear SLAM algorithm.

Using the proposed Linear SLAM framework, the pose graph SLAM and the D-SLAM mapping can be solved in the same way as feature-based SLAM. The differences are the treatments of orientation of poses and the coordinate transformations. The feature-only map joining requires the coordinate transformation to be done where the coordinate frames are defined by 2 (for 2D cases) or 3 features (for 3D cases). In pose graph SLAM, there can be a number of common poses in the two pose graphs to be fused. Thus the wraparound of robot orientation angles need to be considered. Because we only fuse two maps at a time, the wraparound issue can be dealt with very easily. This is different from the wraparound issue in the linear approximation approach proposed in \cite{Carlone2011,Martinez}. Some other limitations of the linear method in \cite{Carlone2011} are: (1) it can only be applied to 2D pose graph SLAM; (2) it requires special structure on the covariance matrices.

Data association is not considered in the experimental results presented in this paper. However, since our Linear SLAM approach is using linear least squares and the associated information matrices are always available, many of the data association methods suggested for EIF based SLAM algorithms (such as SLSJF \cite{Shoudong:SLSJF-TRO}, CF-SLAM \cite{Combined-filter} and ESEIF \cite{Walter-IJRR}), or optimization based SLAM algorithms (such as iSAM \cite{iSAM}), including the strategies for covariance submatrix recovery, can be applied together with the proposed Linear SLAM algorithm.

\section{Conclusion}\label{sec:conclusion}

This paper demonstrated that submap joining can be implemented in a way such that only solving linear least squares problems and performing coordinate transformations are needed. There is no assumption on the structure of local map covariance matrices and the approach can be applied to feature-based SLAM, pose graph SLAM, as well as D-SLAM, for both 2D and 3D scenarios. Since closed-form solutions exist for linear least squares problems, this new approach avoids the requirements of accurate initial value and iterations which are necessary in most of the existing nonlinear optimization based SLAM algorithms.

Simulation and experimental results using publicly available datasets demonstrated the consistency, efficiency and accuracy of the Linear SLAM algorithms, and that Linear SLAM outperforms some other map joining based SLAM algorithms and some recent developed efficient SLAM algorithms. Although the result of Linear SLAM looks very promising, it is not equivalent to the optimal solution to SLAM based on nonlinear least squares starting from an accurate initial value. Like some other submap joining algorithms, how different the Linear SLAM result is from the optimal solution depends on many factors such as the quality of the original sensor data as well as the size and quality of the local maps. Further research is necessary to analyze this in details and develop smart strategies for separating SLAM data for building local maps in order to optimize the performance of Linear SLAM.



\appendix

\section{Proof of Lemma 1}\label{Appendix::A1}

In this appendix, we provide a proof of Lemma 1 in Section \ref{testShoudong}. Here we use some notations in
Section \ref{sec:why-linear}.

Suppose $(\textbf{Z}, \; I_Z)$ are the information from the two
local maps defined in (\ref{Eq::ObsFuncMJ}) and (\ref{Eq::MJ
Information}). The unique optimal solution to minimize
(\ref{Eq::LLSMJ}) is given by (\ref{Eq::LLSMJSolveSt}) and the
corresponding information matrix is given by
(\ref{Eq::LLSMJSolveI}).

That is, for any $\bar{\textbf{X}}^{G_{12}}\neq
\hat{\bar{\textbf{X}}}^{G_{12}}$,
\begin{equation}\label{LS1} \| Z- A\bar{\textbf{X}}^{G_{12}} \|^2_{I_Z}>\| Z- A\hat{\bar{\textbf{X}}}^{G_{12}}
\|^2_{I_Z}.
\end{equation}

The one-to-one transformation function that transforming between
the state vector $\bar{\textbf{X}}^{G_{12}}$ and
${\textbf{X}}^{G_{12}}$ are given in (\ref{Eq::NewSV}) and
(\ref{Eq:transform-g}).

The nonlinear least squares problem for building the global map $M^{G_{12}}$ in the coordinate frame of the end pose of local map 2 is to minimize (\ref{Eq::ObjFunc_N12}), which can be reformulated as
\begin{equation}\label{LS2}
\| Z- A g(\textbf{X}^{G_{12}}) \|^2_{I_Z}.
\end{equation}

The unique optimal solution to minimize (\ref{LS2}) must be
\begin{equation}\label{XG12*}
\hat{\textbf{X}}^{G_{12}} = g^{-1}(\hat{\bar{\textbf{X}}}^{G_{12}}).
\end{equation}

Because for any $\textbf{X}^{G_{12}}\neq
g^{-1}(\hat{\bar{\textbf{X}}}^{G_{12}})$, $g(\textbf{X}^{G_{12}})\neq
\hat{\bar{\textbf{X}}}^{G_{12}}$, and thus
\begin{equation}
\| Z- A g(\textbf{X}^{G_{12}}) \|^2_{I_Z} \neq \| Z-
A \hat{\bar{\textbf{X}}}^{G_{12}} \|^2_{I_Z}.
\end{equation}

Since $\| Z- A \hat{\bar{\textbf{X}}}^{G_{12}} \|^2_{I_Z}$ is the
minimal objective function value, thus
\begin{equation}\label{LS22}
\| Z- A g(\textbf{X}^{G_{12}}) \|^2_{I_Z} > \| Z-
A \hat{\bar{\textbf{X}}}^{G_{12}} \|^2_{I_Z}=\| Z-
A g(\hat{\textbf{X}}^{G_{12}}) \|^2_{I_Z}.
\end{equation}

Now we have proved that the optimal solution to (\ref{LS2}) must
be (\ref{XG12*}). That is, applying the coordinate transformation
on the linear least squares solution.

Because the Jacobian $J$ of $A g(\textbf{X}^{G_{12}})$ with respect to $\textbf{X}^{G_{12}}$, evaluated at $\hat{\textbf{X}}^{G_{12}}$ can be computed by
\begin{equation}
J = A \nabla
\end{equation}
where $\nabla$ is the Jacobian of $g(\textbf{X}^{G_{12}})$ with respect to
$\textbf{X}^{G_{12}}$, evaluated at $\hat{\textbf{X}}^{G_{12}}$, as given in (\ref{Eq::nabla}), thus the corresponding information matrix can be computed as
\begin{equation}\label{LS2JTJ}
\begin{aligned}
I^{G_{12}} = J^T I_Z J = \nabla^T \, A^T I_Z A \, \nabla = \nabla^T
\,\bar{I}^{G_{12}} \, \nabla
\end{aligned}
\end{equation}
which is equivalent to computing the information matrix through
the information matrix of $\hat{\bar{\textbf{X}}}^{G_{12}}$ as in (\ref{Eq::InfoM}).

Thus we have proved Lemma 1.

\begin{figure}[t]
\vspace*{0.55cm}
\centering
\includegraphics[width=0.42\textwidth]{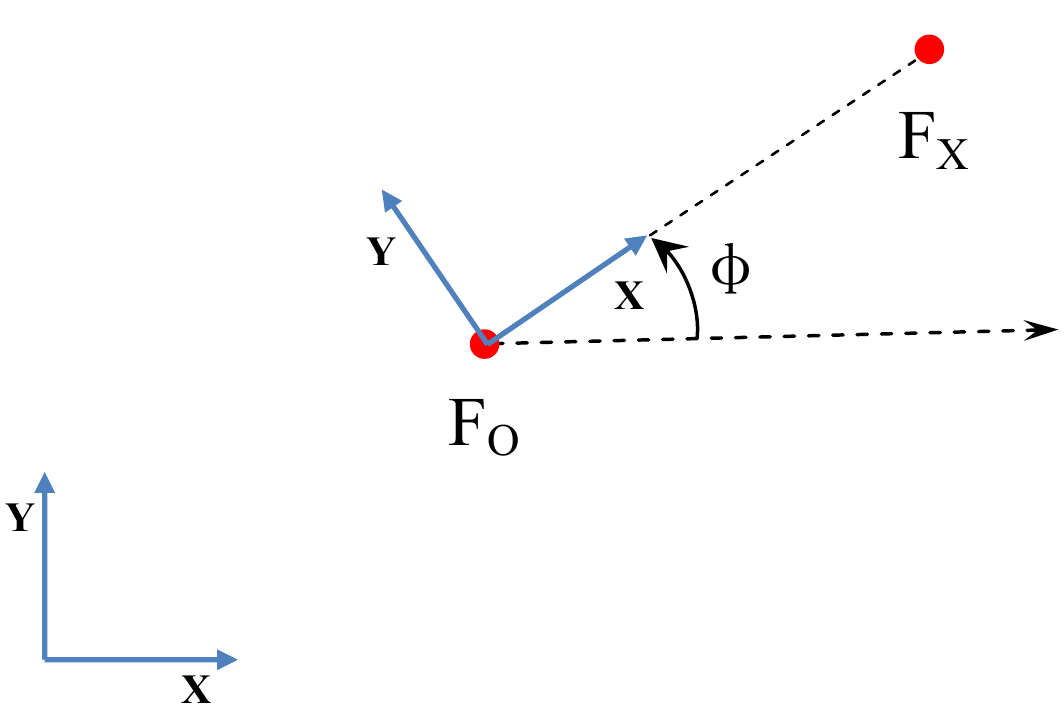}
\caption{Definition of coordinate frame in 2D feature-only map.}
\label{fig:DSLAM_2D}
\end{figure}

\section{Formula of Coordinate Transformation for Feature-only Maps}\label{Appendix::A2}

In this appendix, we provide the details about the definition of the coordinate frame and the coordinates transformation for feature-only map, in 2D and 3D scenarios.

\begin{figure}[t]
\centering
\includegraphics[width=0.42\textwidth]{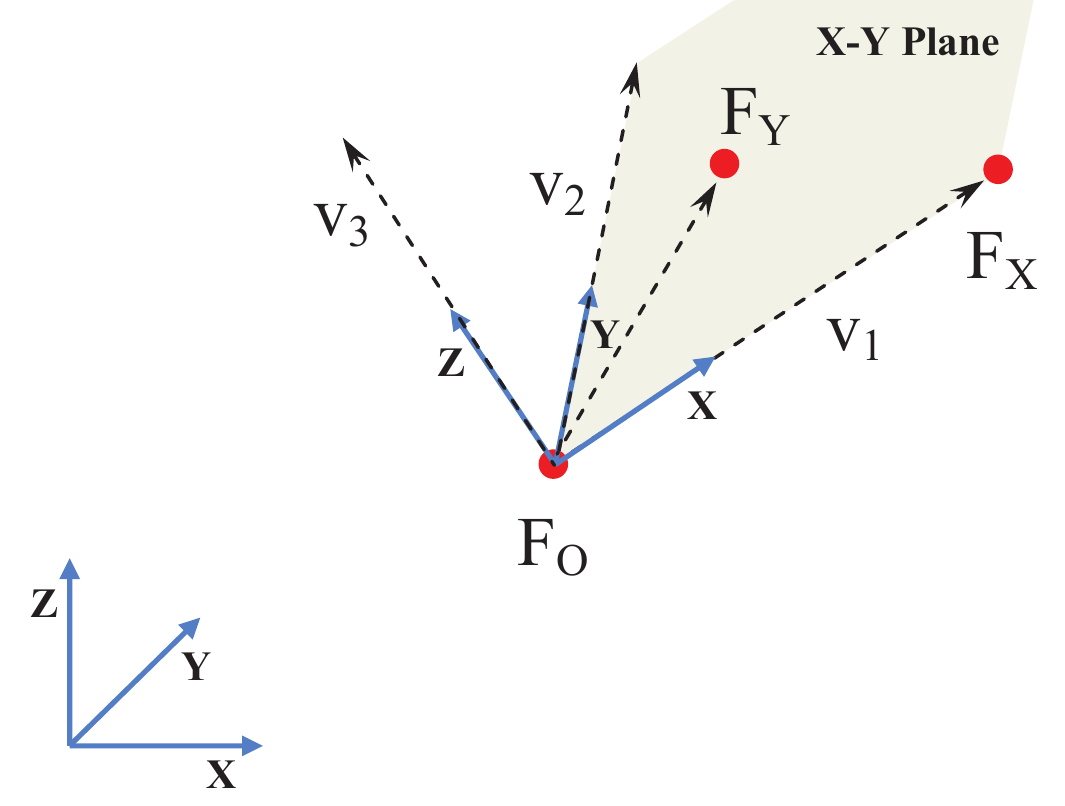}
\caption{Definition of coordinate frame in 3D feature-only map.}
\label{fig:DSLAM_3D}
\end{figure}

\subsection{Coordinate Frame in 2D Feature-only Map}\label{App::2D_Definition}

In a 2D feature-only map, two features are used to define the coordinate frame of the map.

Suppose the two features which are used to define
the coordinate frame are $\textbf{F}_{O}$ and $\textbf{F}_{X}$. We define $\textbf{F}_{O}$ as the origin of the map, and define the direction from feature $\textbf{F}_{O}$ to $\textbf{F}_{X}$ (the vector $\textbf{F}_{X}-\textbf{F}_{O}$) as the $x$ axis of the coordinate frame. The coordinate frame defined by feature $\textbf{F}_{O}$ and $\textbf{F}_{X}$ is shown in Fig. \ref{fig:DSLAM_2D}.

By definition of the coordinate frame using $\textbf{F}_{O}$ and $\textbf{F}_{X}$, $\textbf{F}_{O}=[0,0]^T$ and the $y$ coordinate of $\textbf{F}_{X}$ is $0$. These three parameters are not included in the state vector of the feature-only map.

\subsection{Coordinate Frame in 3D Feature-only Map}\label{App::3D_Definition}

In a 3D feature-only map, three features are used to define the coordinate frame of the map.

Suppose the three features which are used to define the coordinate frame are $\textbf{F}_{O}$, $\textbf{F}_{X}$ and $\textbf{F}_{Y}$. We first define $\textbf{F}_{O}$ as the origin of the map and define the direction from feature $\textbf{F}_{O}$ to $\textbf{F}_{X}$ (the vector $\textbf{F}_{X}-\textbf{F}_{O}$) as the $x$ axis of the coordinate frame. Then the $x$-$y$ plane is defined by the $x$ axis and the vector $\textbf{F}_{Y}-\textbf{F}_{O}$. Thus the $z$ axis can be defined by the cross product of the $x$ axis and $\textbf{F}_{Y}-\textbf{F}_{O}$, which is perpendicular to the $x$-$y$ plane. Then, the $y$ axis can be defined by the cross product of $z$ and $x$ axis. The coordinate frame defined by feature $\textbf{F}_{O}$, $\textbf{F}_{X}$ and $\textbf{F}_{Y}$ is shown in Fig. \ref{fig:DSLAM_3D}.

Because $\textbf{F}_{X}-\textbf{F}_{O}$ and $\textbf{F}_{Y}-\textbf{F}_{O}$ are used to define the $x$-$y$ plane, the three features which define the coordinate frame of the 3D feature-only map cannot be collinear.

By definition of the coordinate frame using $\textbf{F}_{O}$, $\textbf{F}_{X}$ and $\textbf{F}_{Y}$, $\textbf{F}_{O}$ is the origin thus $\textbf{F}_{O} = [0,0,0]^T$. $\textbf{F}_{X}$ is on the $x$ axis thus the $y$ and $z$ coordinates of $\textbf{F}_{X}$ are $0$. $\textbf{F}_{Y}$ is in the $x-y$ plane thus the $z$ coordinate of $\textbf{F}_{Y}$ is $0$. These six parameters defines the 6 DOF 3D coordinate frame, thus are not included in the state vector of the feature-only map.

\subsection{Coordinate Transformation for 2D Feature-only Map}

Suppose in the current coordinate frame, $\textbf{F}_{O}=[x_{F_{O}}, y_{F_{O}}]^T$, $\textbf{F}_{X}=[x_{F_{X}}, y_{F_{X}}]^T$ and a feature position is $\textbf{F}$. In the new coordinate frame defined by $\textbf{F}_{O}$ and $\textbf{F}_{X}$ as described in Appendix \ref{App::2D_Definition},  the new coordinates of the same feature, $\textbf{F}'$, can be obtained by a coordinate transformation
\begin{equation}\label{Eq::transform}
\textbf{F}' = R \; (\textbf{F}-\textbf{t}).
\end{equation}

From Fig. \ref{fig:DSLAM_2D}, it is easy to see that the translation $\textbf{t} = \textbf{F}_{O}$. And the rotation angle $\phi$ from the current coordinate frame to the new coordinate frame can be computed as
\begin{equation}
\phi = \textrm{atan}2 \left ( y_{F_{X}}-y_{F_{O}},\; x_{F_{X}}-x_{F_{O}} \right).
\end{equation}
Thus the rotation matrix $R = r(\phi)$ where $r(\cdot)$ is the angle-to-matrix function.

\subsection{Coordinate Transformation for 3D Feature-only Map}

Now we consider the 3D coordinate transformation from the current coordinate frame to the coordinate frame defined by the features $\textbf{F}_{O}$, $\textbf{F}_{X}$ and $\textbf{F}_{Y}$.

The coordinate transformation in 3D is also in the format of (\ref{Eq::transform}). The translation in the 3D coordinate transformation is also $\textbf{t} = \textbf{F}_{O}$.

For the rotations, suppose vectors $\textbf{v}_1$, $\textbf{v}_2$ and $\textbf{v}_3$ are computed as
\begin{equation}
\left \{
\begin{aligned}
\textbf{v}_1 &= \textbf{F}_X-\textbf{F}_O\\
\textbf{v}_3 &=  \textbf{v}_1 \times (\textbf{F}_Y-\textbf{F}_O)\\
\textbf{v}_2 &=  \textbf{v}_3 \times \textbf{v}_1\\
\end{aligned}
\right.
\end{equation}
As described in Appendix \ref{App::3D_Definition}, $\textbf{v}_1$, $\textbf{v}_2$ and $\textbf{v}_3$ are the vectors in the direction of the $x$, $y$ and $z$ axis of the coordinate frame defined by the features $\textbf{F}_{O}$, $\textbf{F}_{X}$ and $\textbf{F}_{Y}$, respectively. Thus we have
\begin{equation}
\left \{
\begin{aligned}
R \; \textbf{v}_1 &= [\| \textbf{v}_1 \|,0,0]^T\\
R \; \textbf{v}_2 &= [0,\| \textbf{v}_2 \|,0]^T\\
R \; \textbf{v}_3 &= [0,0,\| \textbf{v}_3 \|]^T.\\
\end{aligned}
\right.
\end{equation}
So the rotation matrix $R$ in (\ref{Eq::transform}) is given by
\begin{equation}
R = [\textbf{v}_1/\| \textbf{v}_1 \|,\textbf{v}_2/\| \textbf{v}_2 \|,\textbf{v}_3/\| \textbf{v}_3 \|]^T.
\end{equation}

\end{document}